\documentclass[10pt,a4paper,oneside]{article}
\usepackage[top=0.75in,bottom=0.75in,left=0.75in,right=0.75in]{geometry}
\usepackage{amsmath}
\usepackage{graphicx}
\usepackage{subfigure}
\usepackage{amssymb}
\numberwithin{equation}{section}
\usepackage{subfigure}
\usepackage{threeparttablex}
\usepackage{csquotes}
\usepackage{enumitem}
\usepackage{blindtext}
\usepackage{scrextend}
\addtokomafont{labelinglabel}{\sffamily}
\usepackage{epstopdf}
\usepackage{caption}
\usepackage{array}
\usepackage{lineno}
\usepackage{color,colortbl}
\usepackage{hyperref}
\hypersetup{
    colorlinks=true,
    linkcolor=blue,
    filecolor=magenta,      
    urlcolor=cyan,
}
\newcolumntype{Y}{>{\centering\arraybackslash}X}

\definecolor{Gray}{gray}{0.9}           
\definecolor{LightCyan}{rgb}{0.88,1,1}

\graphicspath{{Images/}}
\usepackage{cite}
\usepackage{epsfig}
\usepackage{amsmath,amssymb, amsthm,latexsym,graphics,graphicx,color}
\usepackage{booktabs}
\usepackage{algorithmic}
\usepackage{algorithm}

\let\tempcommand\description
\renewcommand{\description}{
  \tempcommand
  \setlength{\itemsep}{1pt}
  \setlength{\parskip}{0pt}
  \setlength{\parsep}{0pt}
}

\graphicspath{{images/}}

\begin{document}
\title{Automated Segmentation of Retinal Layers from Optical Coherent Tomography Images Using Geodesic Distance}
\author{Jinming Duan$^*$, Christopher Tench$^\dagger$, Irene Gottlob$^\ddagger$, Frank Proudlock$^\ddagger$, Li Bai$^*$\\
$^*$School of Computer Science, University of Nottingham, UK \\
$^\dagger$School of Medicine, University of Nottingham, UK \\
$^\ddagger$Ophthalmology Department, University of Leicester, UK}
\date{}
\maketitle

\begin{abstract}
\noindent Optical coherence tomography (OCT) is a non-invasive imaging technique that can produce images of the eye at the microscopic level. OCT image segmentation to localise retinal layer boundaries is a fundamental procedure for diagnosing and monitoring the progression of retinal and optical nerve disorders. In this paper, we introduce a novel and accurate geodesic distance method (GDM) for OCT segmentation of both healthy and pathological images in either two- or three-dimensional spaces. The method uses a weighted geodesic distance by an exponential function, taking into account both horizontal and vertical intensity variations. The weighted geodesic distance is efficiently calculated from an Eikonal equation via the fast sweeping method. The segmentation is then realised by solving an ordinary differential equation with the geodesic distance. The results of the GDM are compared with manually segmented retinal layer boundaries/surfaces. Extensive experiments demonstrate that the proposed GDM is robust to complex retinal structures with large curvatures and irregularities and it outperforms the parametric active contour algorithm as well as the graph theoretic based approaches for delineating the retinal layers in both healthy and pathological images. 
\end{abstract}

\section{\textbf{Introduction}}
\label{Introduction}
Optical coherence tomography (OCT) is a powerful imaging modality used to image biological tissues to obtain structural and molecular information \cite{huang1991optical}. By using the low coherence interferometry, OCT can provide high-resolution cross-sectional images from backscattering profiles of biological samples. Over the past two decades, OCT has become a well-established imaging modality and widely used by ophthalmologists for diagnosis of retinal and optical nerve diseases. One of the OCT imaging biomarkers for retinal and optical nerve disease diagnosis is the thickness of the retinal layers. Automated OCT image segmentation is therefore necessary to delineate the retinal boundaries. 

Since the intensity patterns in OCT images are the result of light absorption and scattering in retinal tissues, OCT images usually contain a significant amount of speckle noise and inhomogeneity, which reduces the image quality and poses challenges to automated segmentation to identify retinal layer boundaries and other specific retinal features. Retinal layer discontinuities due to shadows cast by the retinal blood vessels, irregular retinal structures caused by pathologies, motion artefacts and sub-optimal imaging conditions also complicate the OCT images and cause inaccuracy or failure of automated segmentation algorithms. 

\begin{figure}[h!] 
\centering  
{\includegraphics[width=0.65\textwidth]{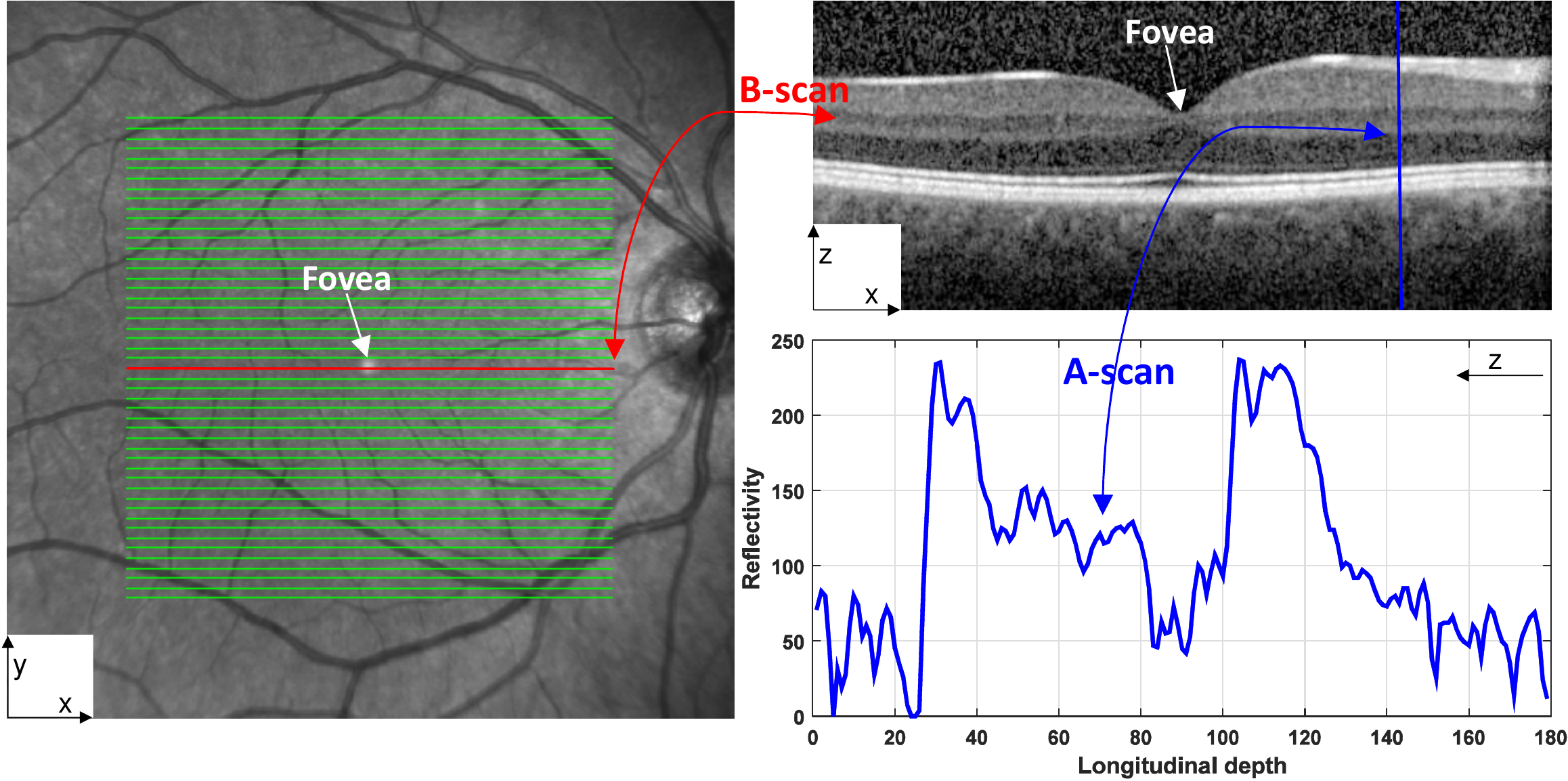}}
\vspace{-10pt}
\caption{A en-face fundus image (left) with lines overlaid representing the locations of each B-scan within a volumetric OCT data. The red line corresponds to the B-scan in the image (right top). One vertical A-scan of the B-scan is shown in the plot (right bottom). The fovea region is characterised by a depression in the centre of the retina surface.}
\label{fig:ABV}
\vspace{-10pt}
\end{figure}

Over the past two decades a number of automatic and semi-automatic OCT segmentation approaches have been proposed. These approaches can be roughly categorised into three families: A-scan based methods, B-scan based methods and volume based methods, as illustrated in Figure~\ref{fig:ABV}. A-scan based methods \cite{hee1995optical,koozekanani2001retinal,ishikawa2002detecting,ishikawa2005macular,shahidi2005quantitative,fernandez2005automated,mayer2008automatic} detect intensity peak or valley points on the boundaries in each A-scan profile and then form a smooth and continuous boundary by connecting the detected points using model fitting techniques. These methods can be inefficiency and lack of accuracy. Common approaches for segmenting two-dimensional (2D) B-scans include active contour methods \cite{fernandez2005delineating,mujat2005retinal,mishra2009intra,yazdanpanah2009intra,ghorbel2011automated,rossant2015parallel}, shortest-path based graph search \cite{chiu2010automatic,yang2010automated} and statistical shape models \cite{kajic2010robust,kajic2012automated,pilch2012automated} (i.e. active shape and appearance models \cite{cootes1995active, cootes2001active}). B-scans methods outperform A-scan methods in general. However, they are prone to the intrinsic speckle noise in OCT images and more likely to fail in detection of pathological retinal structures. Three-dimensional (3D) scan of the retina is now widely used in commercial OCT devices. Existing volume based segmentation methods mainly use 3D graph based methods \cite{haeker2006segmentation,garvin2008intraretinal,garvin2009automated,quellec2010three,antony2012incorporation,dufour2013graph,kafieh2013intra,tian2015real,tian2016performance} and pattern recognition \cite{vermeer2010automated,vermeer2011automated,fuller2007segmentation,szkulmowski2007analysis,lang2013retinal}. Benefiting from contextual information represented in the analysis graph, graph based methods provide optimal solutions and ideal for volumetric data processing. However, the computation can be very complex and slow. Pattern recognition methods normally require training data manually segmented by experts in order to learn a feasible model for classification. These approaches also suffer in accuracy and efficiency. Segmentation of retinal layers in OCT images thereby remains a challenging problem.

In this paper, we propose an algorithm for retinal layer segmentation based on a novel geodesic distance weighted by an exponential function. As opposed to using a single horizontal gradient as in other works \cite{chiu2010automatic,tian2015real,tian2016performance}, the exponential function employed in our method integrates both horizontal and vertical gradient information and can thus account for variations in the both directions. The function plays the role of enhancing the foveal depression regions and highlighting weak and low contrast boundaries. As a result, the proposed geodesic distance method (GDM) is able to segment complex retinal structures with large curvatures and other irregularities caused by pathologies. We compute the weighted geodesic distance via an Eikonal equation using the fast sweeping method \cite{zhao2005fast,tsai2003fast,duan2015surface}. A retinal layer boundary can then be detected based on the calculated geodesic distance by solving an ordinary differential equation via a time-dependent gradient descent equation. A local search region is identified based on the detected boundary to delineate all the nine retinal layer boundaries and overcome the local minima problem of the GDM. We evaluate the proposed GDM through extensive numerical experiments and compare it with state-of-the-art OCT segmentation approaches on both healthy and pathological images. 

In the following sections, we shall first review the state-of-the-art methods that are to be compared with the proposed GDM, such as parallel double snakes \cite{rossant2015parallel}, Chiu's graph search \cite{chiu2010automatic}, Dufour's method \cite{dufour2013graph}, and OCTRIMA3D \cite{tian2015real,tian2016performance}. This will be followed by the details of the proposed GDM, ground-truth validation, numerical experimental results, and comparison of the GDM with the state-of-the-art methods.

\section{\textbf{Literature Review}}
\label{reviewedMethod}
\begin{figure}[h!] 
\centering  
{\includegraphics[height=0.23\textwidth]{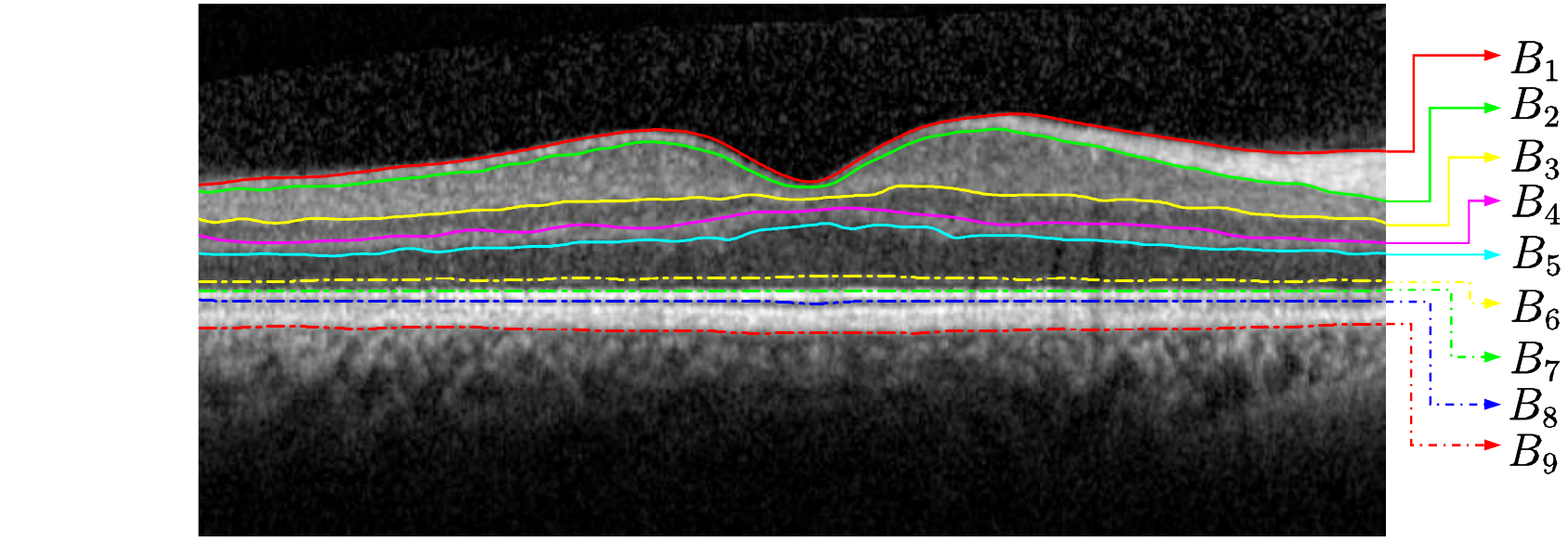}}
\vspace{-5pt}
\caption{An example cross-sectional B-Scan OCT image centred at the macula, showing nine target intra-retinal layer boundaries detected by the proposed method. The names of these boundaries labelled as notations $B_1$,$B_2$...$B_9$ are summarised in Table~\ref{tb:OCTBoundary}. Knowledge of these layer boundaries allows us to calculate the retinal layer thickness, which is imperative for detecting and monitoring ocular diseases.}
\label{fig:OCTBoundary}
\end{figure}

In this section, we will provide an overview of the state-of-the-art methods (i.e. parallel double snakes \cite{rossant2015parallel}, Chiu's method \cite{chiu2010automatic}, OCTRIMA3D \cite{tian2015real,tian2016performance}, Dufour's method \cite{dufour2013graph}) that will be compared with our proposed GDM in Section \ref{GDM}. For a complete review on the subject, we refer the reader to \cite{debuc2011review}. Among the four methods reviewed, the first two can only segment B-scans, while the latter two are able to extract retinal surfaces from volumetric OCT data. We note that the term `surface' refers to a set of voxels that fall on the interface between two adjacent retinal layer structures. The retinal layer boundaries to be delineated are shown in Figure~\ref{fig:OCTBoundary}.

\textbf{Parallel double snakes (PDS)}: Rossant et al. \cite{rossant2015parallel} detected the pathological (retinitis pigmentosa) cellular boundaries in B-scan images by minimising an energy functional that includes two parallel active parametric contours. Their proposed PDS model consists of a centreline $C(s)=(x(s),y(s))$ parametrised by $s$ and two parallel curves $C_1(s)=C(s)+b(s)n(s)$ and $C_2(s)=C(s)-b(s)n(s)$ with $b(s)$ being a spatially varying half-thickness and $n(s)=(n_x(s),n_y(s))$ the normal vector to the the centreline $C(s)$. Specifically, their PDS model is defined as
\begin{equation}
E(C,{C_1},{C_2},b) = {E_{Image}}({C_1}) + {E_{Image}}({C_2}) + {E_{Int}}(C) + R\left( {{C_1},{C_2},b} \right)
\label{eq:PDS}
\end{equation}
where the image energy ${E_{Image}}({C_1}) =- \int_0^1 {{{\left| {\nabla I({C_1})} \right|}^2}ds}$ ($\nabla$ is the image gradient operator) attracts the parametric curve $C_1$ towards one of retinal borders of the input B-scan $I$, whilst ${E_{Image}}({C_2})$ handles curve $C_2$ which is parallel to $C_1$. The internal energy ${E_{Int}}(C)=\frac{\alpha }{2}\int_0^1 {{{\left| {{C_s}\left( s \right)} \right|}^2}ds}  + \frac{\beta }{2}\int_0^1 {{{\left| {{C_{ss}}\left( s \right)} \right|}^2}ds}$ imposes both first and second order smooth regularities on the central curve $C$, with $\alpha$ and $\beta$ respectively controlling the tension and rigidity of this curve. $R\left( {{C_1},{C_2},b} \right)=\frac{\varphi }{2}\int_0^1 {{{\left| {b'\left( C \right)} \right|}^2}ds}$ is a parallelism constraint imposed on $C_1$ and $C_2$. Nine retinal borders have been delineated by the method, i.e., ILM, RNFL$_o$, IPL-INL, INL-OPL, OPL-ONL, ONL-IS, IS-OS, OS-RPE and RPE-CH.

\newcolumntype{C}{>{\centering\arraybackslash}p{7em}}
\begin{table}[ht]
\centering  
\caption{Notations for nine retinal boundaries/surfaces, their corresponding names and abbreviations}
\vspace{-10pt}
\begin{tabular}{lcc}
\toprule
Notation  & Name of retinal boundary/surface & Abbreviation\\
\midrule
\rowcolor{Gray}
$B_1$&  internal limiting membrane & ILM \\
\rowcolor[RGB]{195, 195, 195}
$B_2$&  outer boundary of the retinal nerve fibre layer & RNFL$_o$\\
\rowcolor{Gray}
$B_3$&  inner plexiform layer-inner nuclear layer & IPL-INL  \\
\rowcolor[RGB]{195, 195, 195}
$B_4$&  inner nuclear layer-outer plexiform layer & INL-OPL \\
\rowcolor{Gray}
$B_5$&  outer plexiform layer-outer nuclear layer & OPL-ONL \\
\rowcolor[RGB]{195, 195, 195}
$B_6$& 	outer nuclear layer-inner segments of photoreceptors & ONL-IS\\
\rowcolor{Gray}
$B_7$&  inner segments of photoreceptors-outer segments of photoreceptors &IS-OS \\
\rowcolor[RGB]{195, 195, 195}
$B_8$&  outer segments of of photoreceptors-retinal pigment epithelium & OS-RPE\\
\rowcolor{Gray}
$B_9$&  retinal pigment epithelium-choroid & RPE-CH\\
\bottomrule
\end{tabular}
\label{tb:OCTBoundary}
\end{table} 

\textbf{Chiu's method}: Chiu et al. \cite{chiu2010automatic} modelled the boundary detection problem in OCT retinal B-scan as determining the shortest-path that connects two endpoints in a graph $G=(V,E)$, where $V$ is a set of nodes and $E$ is a set of undirected weights assigned to each pair of two nodes in the graph. Node $V$ corresponds to each pixel in the B-scan image, whilst weight $E$ is calculated from the intensity gradient of the image in its vertical direction. Each node is connected with its eight nearest neighbours and all other node pairs are disconnected, resulting in a sparse adjacency matrix of graph weights of vertical intensity variation. For example, an $M \times N$ sized image has an $MN \times MN$ sized adjacency matrix with $8MN$ non-zero filled entries. Mathematically, the weights between two nodes used in their method are calculated based on the pure vertical gradient value, defined as
\begin{equation}
w\left( {a,b} \right) = \left\{ \begin{array}{ll}
2 - \left( {g_a + g_b} \right) + {w_{\min }}&\text{if}\;\left| {a - b} \right| \le \sqrt 2 \\
0&{\text{otherwise}}
\end{array} \right.
\label{eq:chiu}
\end{equation}
where $g$ is the vertical gradient of a B-scan image; $a$ and $b$ denote receptively two separate nodes in $V$ and $w_{min}$ is a small positive value added to stabilise the system. The most prominent boundary is then detected as the minimum weighed path from the first to the last vertex in $V$ using Dijkstra's Algorithm. A similar region refinement technique to Section \ref{Detect9border} was used to detect seven retinal boundaries, i.e., ILM, RNFL$_o$, IPL-INL, INL-OPL, OPL-ONL, IS-OS and RPE-CH.

\textbf{Dufour's method}: Dufour et al. \cite{dufour2013graph} proposed a modification of optimal graph search approach \cite{song2010simultaneous} to segment retinal surfaces in OCT volume data. By using soft constraints and adding prior knowledge learned from a model, they improve the accuracy and robustness of the original framework. Specifically, their Markov random field based model is given by 
\begin{equation} \nonumber
E\left( S \right) = \sum\limits_{i = 1}^n {\left( {{E_{boundary}}\left( {{S_i}} \right) + {E_{smooth}}\left( {{S_i}} \right)} \right)}  + \sum\limits_{i = 1}^{n - 1} {\sum\limits_{j = i + 1}^n {{E_{{\mathop{ int}} er}}\left( {{S_i},{S_j}} \right)} } 
\end{equation}
where $S$ is a set of surfaces $S_1$ to $S_n$. The external boundary energy ${{E_{boundary}}\left( {{S_i}} \right)}$ is computed from the input 3D image data. The surface smoothness energy ${{E_{smooth}}\left( {{S_i}} \right)}$ guarantees the connectivity of a surface in 3D and regularises the surface. The interaction energy ${{E_{{\mathop{\rm int}} er}}\left( {{S_i},{S_j}} \right)}$ integrates soft constraints that can regularise the distances between two simultaneously segmented surfaces. This model is then built from training datasets consisting of fovea-centered OCT slice stacks. Their algorithm is capable to segment six retinal surfaces ($n=6$ in above formulation) in both healthy and macular edema subjects, i.e., ILM, RNFL$_o$, IPL-INL, OPL-ONL, IS-OS and RPE-CH. 

\textbf{OCTRIMA3D}: Tian et al. \cite{tian2015real,tian2016performance} proposed a real-time automatic segmentation of OCT volume data. The segmentation was done frame-by-frame in each 2D B-Scan by considering the spatial dependency between each two adjacent frames. Their work is based on Chiu's graph search framework \cite{chiu2010automatic} for B-Scan OCT images. However, in addition to Chiu's work they introduce the inter-frame flattening to reduce the curvature in the fovea region and thus the accuracy of their algorithm has been improved. Moreover, they apply inter-frame or intra-frame information to limit the search region in current or adjacent frame so that the computational speed of their algorithm can be increased. Furthermore, the biasing and masking techniques are developed so as to better attain retinal boundaries within the same search region. A totally eight retinal surfaces, i.e., ILM, RNFL$_o$, IPL-INL, INL-OPL, OPL-ONL, IS-OS, OS-RPE and RPE-CH, can be delineated by the method. To sum up, Table~\ref{tb:Checkmark} reports the retinal boundaries/surfaces segmented by the four methods as well as our GDM proposed in the next section. 

\newcolumntype{C}{>{\centering\arraybackslash}p{7em}}
\newcolumntype{g}{>{\columncolor[RGB]{255, 255, 255}}C}
\newcolumntype{k}{>{\columncolor[RGB]{255, 255, 255}}C}
\begin{table*}[htbp] 
\caption{Target boundaries/surfaces of the five methods compared in this paper (check mark means the boundary/surface can be segmented, while cross mark means the boundary/surface cannot be segmented).}
\vspace{-10pt}
\centering \setlength{\tabcolsep}{1pt}
\resizebox{\columnwidth}{!}{
\begin{tabular}{lgkCgkCgkC} 
\toprule
Method            &ILM ($B_1$)      &RNFL$_o$ ($B_2$)       &IPL-INL ($B_3$)  
				  &INL-OPL ($B_4$)  &OPL-ONL ($B_5$)        &ONL-IS ($B_6$) 
                  &IS-OS ($B_7$)    &OS-RPE ($B_8$)         &RPE-CH ($B_9$)  \\
\midrule
\rowcolor{Gray}
PDS \cite{rossant2015parallel}            &\checkmark     &\checkmark    &\checkmark     &\checkmark  &\checkmark  &\checkmark    &\checkmark     &\checkmark  &\checkmark      \\
\rowcolor[RGB]{195, 195, 195}				
Chiu's method \cite{chiu2010automatic}    &\checkmark     &\checkmark    &\checkmark     &\checkmark  &\checkmark  &$\times$      &\checkmark     &$\times$    &\checkmark      \\
\rowcolor{Gray}
Dufour's method  \cite{dufour2013graph}  &\checkmark     &\checkmark    &\checkmark     &$\times$    &\checkmark  &$\times$      &\checkmark     &$\times$    &\checkmark      \\
\rowcolor[RGB]{195, 195, 195} 
OCTRIMA3D   \cite{tian2015real,tian2016performance}      &\checkmark     &\checkmark    &\checkmark     &\checkmark  &\checkmark  &$\times$      &\checkmark     &\checkmark  &\checkmark      \\
\rowcolor{Gray}				  
GDM     &\checkmark     &\checkmark    &\checkmark     &\checkmark  &\checkmark  &\checkmark    &\checkmark     &\checkmark  &\checkmark      \\
			      
\bottomrule
\end{tabular}
}
\label{tb:Checkmark}
\end{table*}

\section{\textbf{The Proposed Geodesic Distance Method (GDM)}}
\label{GDM}
In this section, we propose a novel framework using the geodesic distance to detect from OCT images nine retinal layer boundaries defined in Figure~\ref{fig:OCTBoundary} and Table~\ref{tb:OCTBoundary}. As the proposed methodology applies equally to both 2D and 3D segmentation, we will illustrate the approach for 2D segmentation here, as the steps would be the same for 3D segmentation. Numerical implementation of the approach is given in Appendix. 

\subsection{Geodesic distance}
We use geodesic distance to identify the pixels on the boundaries of retinal layers in OCT images. The geodesic distance $d$ is the smallest integral of a weight function $W$ over all possible paths from two endpoints (i.e. $s_1$ and $s_2$). The weight function determines how the path goes from $s_1$ to $s_2$. Small weight at one point indicates that the path has high possibility of passing that point. Specifically, the weighted geodesic distance between two pixels/endpoints $s_1$ and $s_2$ is given as 
\begin{equation} \label{eq:GeodesicEq}
D\left( {{s_1},{s_2}} \right) = {\min _C}\int_0^1 {W^{-1}\left( {C\left( s \right)} \right)ds} 
\end{equation}
where $C\left( s \right)$ is the set of all the paths that link $s_1$ to $s_2$, and the path length is normalised and the start and end locations are $C(0)=s_1$ and $C(1)=s_2$, respectively. The infinitesimal contour length $ds$ is weighted by a non-negative function $W\left( {C\left( s \right)} \right)$. This minimisation problem can be interpreted as finding a geodesic curve (i.e. a path with the smallest weighted length) in a Riemannian space. In geometrical optics, it has proven that the solution of (\ref{eq:GeodesicEq}) satisfies the Eikonal equation (\ref{eq:eikonalEq}). 

\begin{figure}[h!] 
\centering  
{\includegraphics[width=1\textwidth]{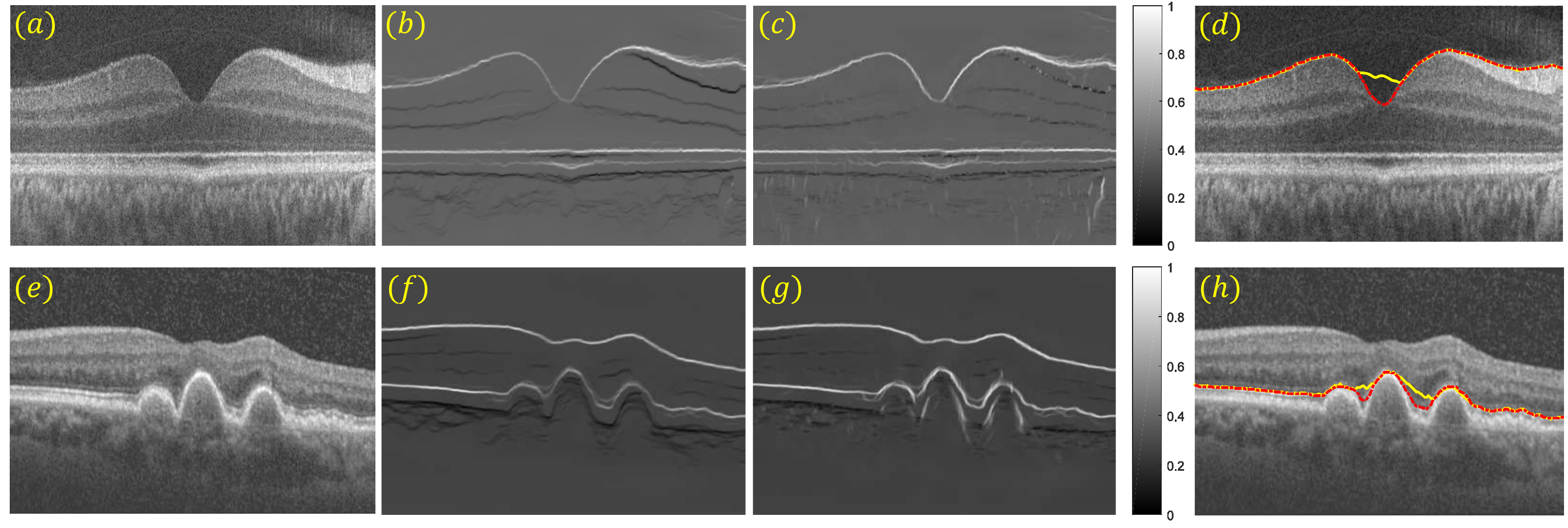}}\\
\vspace{-5pt}
\caption{Illustrating the effectiveness of the weight $W$ defined in (\ref{eq:weights}). (a) and (e): normal B-scan OCT data and pathological B-scan from an eye with dry age-related macular degeneration (drye-AMD); (b) and (f): vertical dark-to-bright gradient maps of (a) and (e), respectively; (c) and (g): dark-to-bright gradient maps calculated using equation (\ref{eq:weights}) with $\lambda=1$. Note the gradient values of pixels have been enhanced in the region with strong curvature and big bumps; (d) and (h): boundary detection results via the method described in Section \ref{EikonalEq} using different gradient maps. Yellow lines are computed using (b) and (f), whilst red lines using (c) and (g). }
\label{fig:gradinetMaps}
\end{figure}

The retinal layers of OCT images are normally near horizontal. The gradient in the vertical direction thus can be considered as a good candidate for computing weight $W$ in (\ref{eq:GeodesicEq}). For instance, each of the two prominent boundaries, e.g. ILM ($B_1$) and IS-OS ($B_7$) in Figure~\ref{fig:gradinetMaps} (a) and (e), is at the border of a dark layer above a bright layer. As a result, pixels in the region around the two boundaries will have high gradient values, as shown in Figure~\ref{fig:gradinetMaps} (b) and (f). As the retinal layers at each side of the boundary are either transiting from dark to bright or bright to dark, the non-negative weight function $W$ in this paper is defined based on intensity variation as follows
\begin{equation} \label{eq:weights}
W\left( x \right) = \left\{ \begin{array}{ll}
1 - exp\left(-\lambda \left( {1 - n\left( {{\nabla _x}I} \right)} \right)n\left( {\left| {{\nabla _y}I} \right|} \right)\right)&\text{dark-to-bright}\\
exp\left(-\lambda\left( {1 - n\left( {{\nabla _x}I} \right)} \right)n\left( {\left| {{\nabla _y}I} \right|} \right)\right)& \text{bright-to-dark}
\end{array} \right.
\end{equation}
where $I$ is an input OCT image; $n\left( \cdot  \right)$ is a linear stretch operator used to normalise values to between 0 and 1; $exp$ is the exponential function and $\lambda$ is a user-define parameter, together they enhance the foveal depression regions and highlight the weak retinal boundaries \cite{duan2016edgeweighted}; and $\nabla_x$ and $\nabla_y$ are the first-order gradient operator along x (vertical) and y (horizontal) direction, respectively. The two gradient operators are discretised using a central finite difference scheme under the Neumann boundary condition. (\ref{eq:weights}) also includes the positive horizontal gradient information $n\left( {\left| {{\nabla _y}I} \right|} \right)$, without which only vertical direction is accounted for and it is thus only applicable to flat retinal boundaries. Consequently, our proposed method is robust against curved features (e.g. the central region of the fovea) as well as other irregularities (e.g. bumps or large variations of boundary locations) caused by pathologies. In other words, the proposed method with the weight $W$ defined in (\ref{eq:weights}) can deal with both normal and pathological images, as illustrated in Figure~\ref{fig:gradinetMaps} as well as in the experimental section.

\subsection{Selection of endpoints $s_1$ and $s_2$}
\begin{figure}[h!] 
\centering  
{\includegraphics[width=0.8\textwidth]{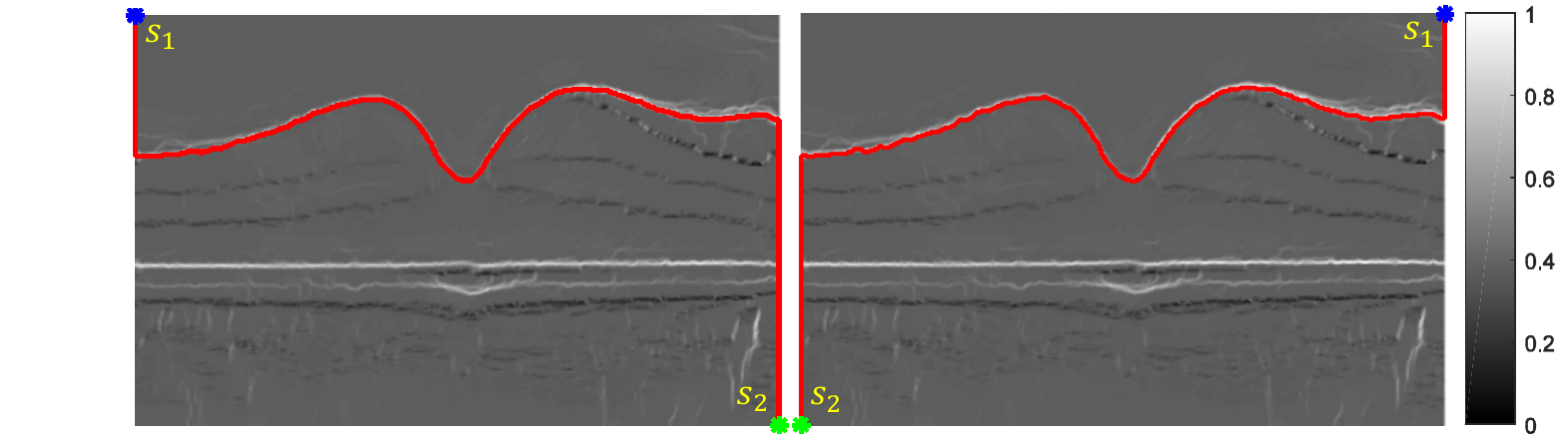}}\\
\vspace{-5pt}
\caption{Two segmentation examples using different automatic endpoints initialisations on a drak-to-bright gradient map. $s_1$ and $s_2$ are start and end points, respectively.}
\label{fig:endPoints}
\end{figure}

For fully automated segmentation, it is essential to find a way to initialise the two endpoints $s_1$ and $s_2$ automatically. Since the retinal boundaries in the OCT images used in this paper run across the entire width of the image, we add an additional column on each side to the gradient map computed from (\ref{eq:weights}). As the the minimal weighted path is sought after, a weight $W_{max}$ larger than any of the non-negative weights calculated from (\ref{eq:weights}) is therefore assigned to each of the newly added vertical columns (note that we use $W^{-1}$ for the geodesic distance \ref{eq:GeodesicEq}, the minimal weighted path thereby prefers large weights). This forces the path traversal in the same direction as the newly added vertical columns with maximal weights, and also allows the start and end points to be arbitrarily assigned in the two columns. Once the retinal layer boundary is detected, the two additional columns can be removed. Figure~\ref{fig:endPoints} shows two examples of endpoint initialisation.

\subsection{\textbf{Eikonal equation and minimal weighted path}}
\label{EikonalEq}
The solution of (\ref{eq:GeodesicEq}) can be obtained by solving the Eikonal equation after the endpoints are determined. Specifically, over a continuous domain, the distance map $D(x)$ to the seed start point $s_1$ is the unique solution of the following Eikonal equation in the viscosity sense 
\begin{equation}\label{eq:eikonalEq}
\left| {\nabla D\left( x \right)} \right| = W^{-1}\left( x \right),\;\;\forall x \notin s_1
\end{equation}
with $D\left( s_1 \right) = 0$. The equation is a first order partial differential equation and its solution can be found via the classical fast marching algorithm \cite{sethian1996fast,sethian1999level} using an upwind finite difference approximation with the computational complexity $O(MNlog(MN))$ (MN is the total number of grid points). Recently, the fast sweeping algorithm \cite{zhao2005fast,tsai2003fast} has been proposed. This technique is based on a pre-defined sweep strategy, replacing the heap priority queue to find the next point to process, and thereby has the linear complexity of $O(MN)$. In this paper, we apply fast sweep for (\ref{eq:eikonalEq}) and its detailed 3D implementation has been given in Appendix. Figure~\ref{fig:distMap} shows two distance maps calculated using the dark-to-bright weight defined in (\ref{eq:weights}) and two different start points as shown in the examples in Figure~\ref{fig:endPoints}.

Once the geodesic distance map to the start point $s_1$ has been computed, the minimal weighted path (geodesic curve) between point $s_1$ and $s_2$ can be extracted from the following ordinary differential equation through the time-dependent gradient descent
\begin{equation} 
\gamma '\left( t \right) =  - {\eta _t}\nabla D\left( {\gamma \left( t \right)} \right),\;\; \gamma \left( 0 \right) = {s_2}
\end{equation}
where $\eta _t>0$ controls the parametrisation speed of the resulting curve. To obtain unit speed parametrisation, we use ${\eta _t} = \left| {\nabla D\left( {\gamma \left( t \right)} \right)} \right|_\varepsilon ^{ - 1}$. Since distance map $D$ is nonsmooth at point $s_1$, a small positive constant $\varepsilon$ is added to avoid dividing by zero. Note the point $s_1$ is guaranteed to be found from this ordinary differential equation because the distance field is monotonically increasing from $s_1$ to $s_2$, which can be observed in Figure~\ref{fig:distMap}. This technique can achieve sub-pixel accuracy for the geodesic path even if the grid is discrete. 
\begin{figure}[h!] 
\centering  
{\includegraphics[width=0.8\textwidth]{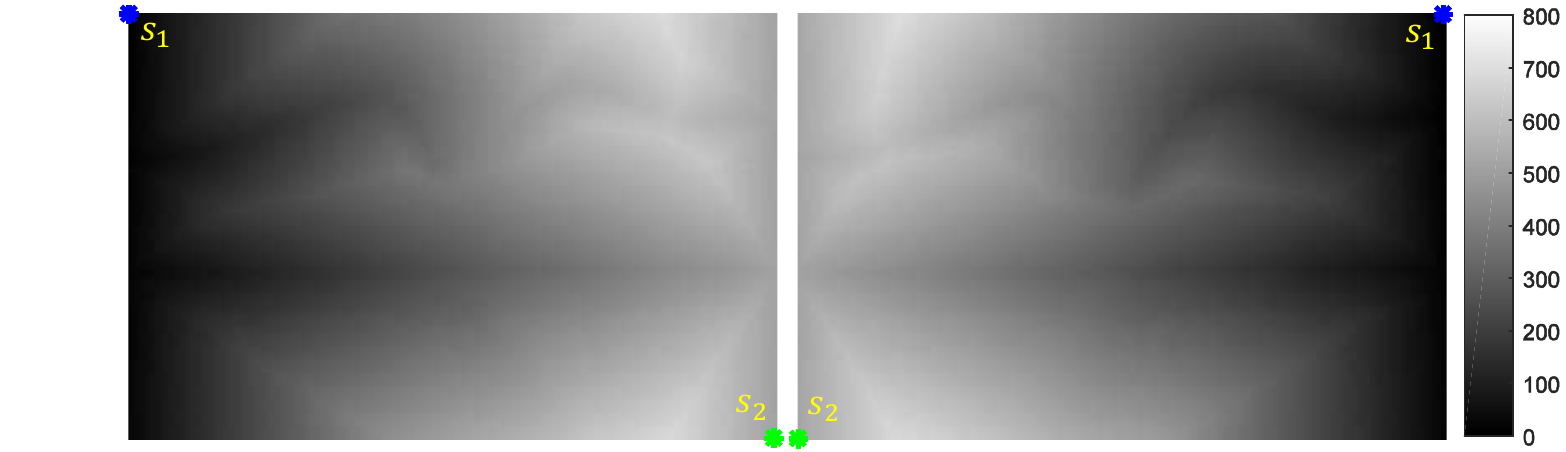}}\\
\vspace{-5pt}
\caption{Two distance maps calculated using the dark-to-bright weight $W^{-1}$ and two different start points in the two exmaples in Figure~\ref{fig:endPoints}, respectively. The distance values are expanded to $[0,800]$ for better visualisation.}
\label{fig:distMap}
\end{figure}

The geodesic curve is then numerically computed using a discretised gradient descent, which defines a discrete curve $\gamma^{k}$ using
\begin{equation} \label{eq:GradientDescentFlow}
{\gamma ^{k + 1}} = {\gamma ^k} - \tau G\left( {{\gamma ^k}} \right)
\end{equation}
where $\gamma^{k}$ is a discrete approximation of $\gamma(t)$ at time $t=k\tau$, and the time step size $\tau>0$ should be small enough. $G\left( x \right)$ is the normalised gradient $ {{\nabla D\left( {\gamma \left( t \right)} \right)} \mathord{\left/
 {\vphantom {{\nabla D\left( {\gamma \left( t \right)} \right)} {\left| {\nabla D\left( {\gamma \left( t \right)} \right)} \right|_\varepsilon ^{ - 1}}}} \right.
 \kern-\nulldelimiterspace} {\left| {\nabla D\left( {\gamma \left( t \right)} \right)} \right|_\varepsilon ^{ - 1}}}$ parametrised by the arc length. Once $\gamma^{k+1}$ reaches $s_1$, one of the retinal boundaries can be found. The following \hyperlink{alogrithm1}{\textbf{Algorithm 1}} concludes the proposed geodesic distance algorithm for extracting one retinal boarder in OCT images. 
\hypertarget{alogrithm1}{}
\begin{table}[h!] 
\centering
\begin{tabular}{p{11cm}}
\toprule
{\textbf{Algorithm 1}}: the proposed GDM for one retinal boundary detection\\
\midrule
\rowcolor{Gray}
1: Input OCT data $I$ (i.e. B-scan or volume)\\
\rowcolor[RGB]{195, 195, 195}
2: calculate dark-to-bright or bright-to-dark weight $W$ using (\ref{eq:weights})\\
\rowcolor{Gray}
3: pad two new columns to the weight and assign large values to them\\
\rowcolor[RGB]{195, 195, 195}
4: select two endpoints $s_1$ and $s_2$ on the two newly padded columns\\
\rowcolor{Gray}
5: calculate distance map $D$ in (\ref{eq:eikonalEq}) using fast sweeping algorithm\\
\rowcolor[RGB]{195, 195, 195}
6: find one retinal layer boundary $\gamma$ using the gradient descent flow (\ref{eq:GradientDescentFlow})\\
\rowcolor{Gray}
7: remove the additional columns in the edge detection result\\
\bottomrule
\end{tabular} 
\end{table} 

\subsection{\textbf{Detection of nine retinal layer boundaries}}
\label{Detect9border}
We have introduced how the proposed geodesic distance algorithm (\ref{eq:GeodesicEq}) can find the minimal weighted path across the whole width of the OCT image for one retinal layer boundary. In this section, we shall describe the implementation details of the proposed approach to delineate nine retinal layer boundaries as shown in Figure~\ref{fig:OCTBoundary} and Table~\ref{tb:OCTBoundary}. Since the proposed model (\ref{eq:GeodesicEq}) is not convex, its solution can easily get stuck in local optima. For example, Figure~\ref{fig:gradinetMaps} (c) and (g) have high gradient values in the region around both the ILM and IS-OS boundaries. However, in Figure~\ref{fig:gradinetMaps} (d) the algorithm detected the ILM boundary while in Figure~\ref{fig:gradinetMaps} (h) it detected the IS-OS. In order to eliminate such uncertainty, we dynamically define the search region based on the detected boundaries. The following section describes the proposed method in detail.

\subsubsection{\textbf{Detection of the IS-OS boundary}}
\begin{figure}[h!] 
\centering  
{\includegraphics[height=0.16\textwidth]{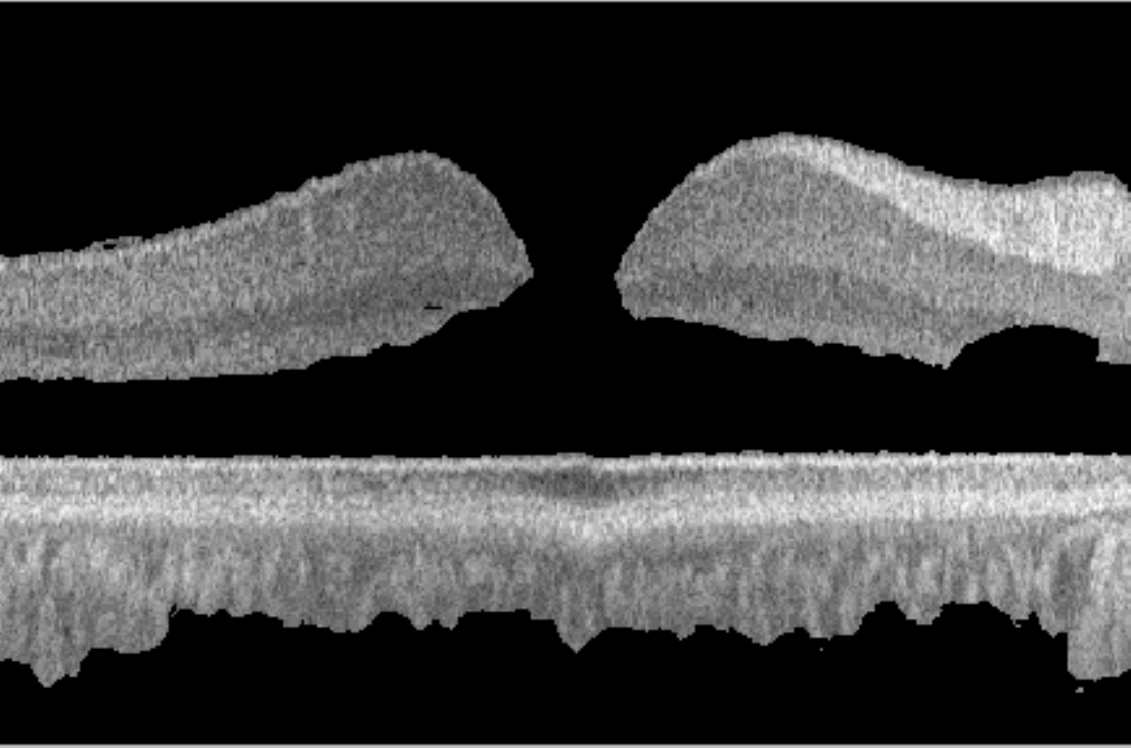}}
{\includegraphics[height=0.16\textwidth]{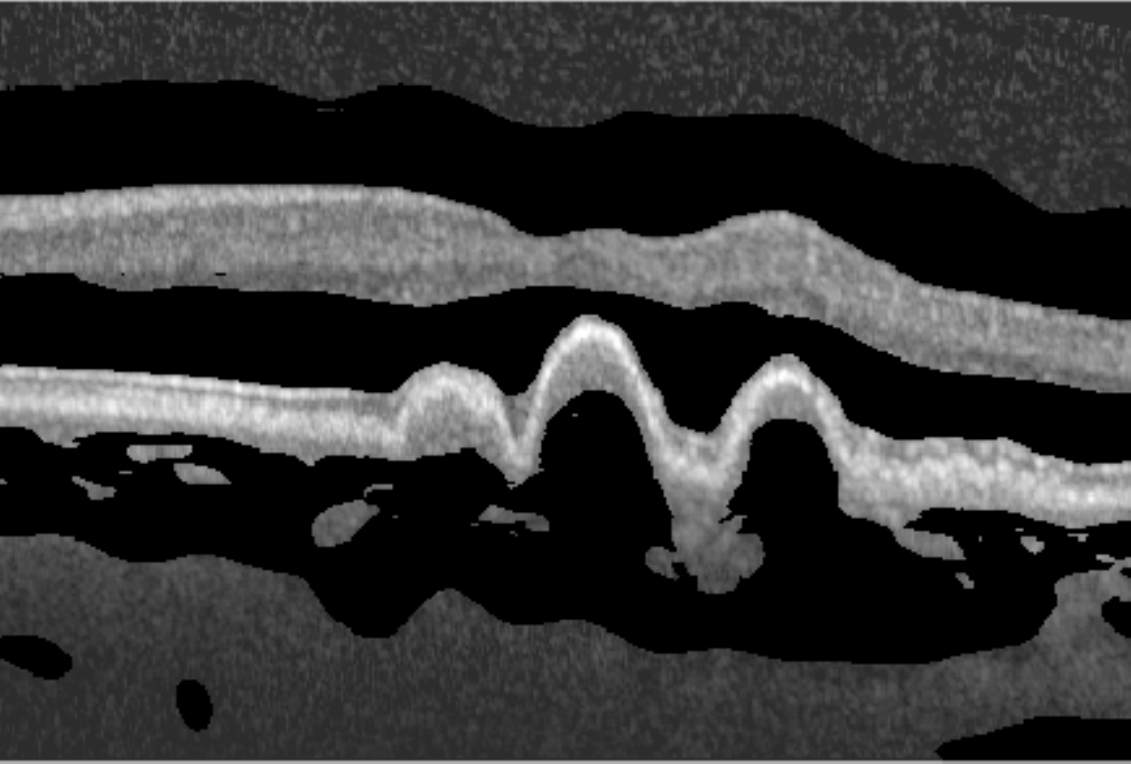}}
{\includegraphics[height=0.16\textwidth]{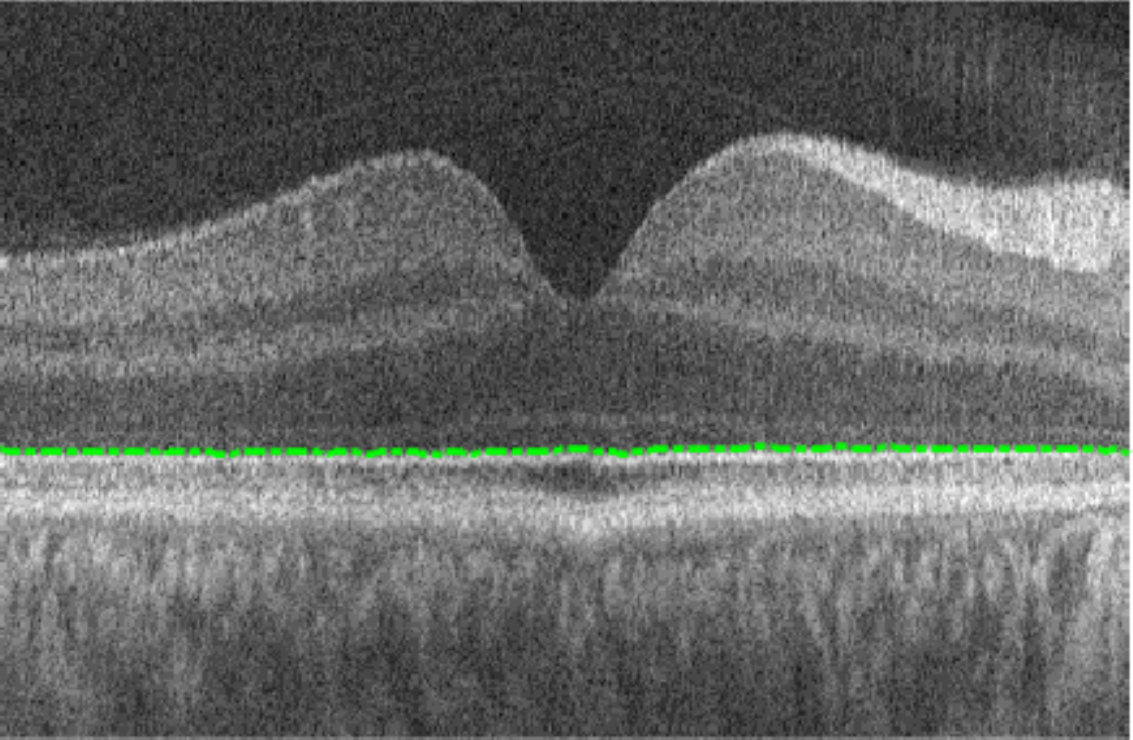}}
{\includegraphics[height=0.16\textwidth]{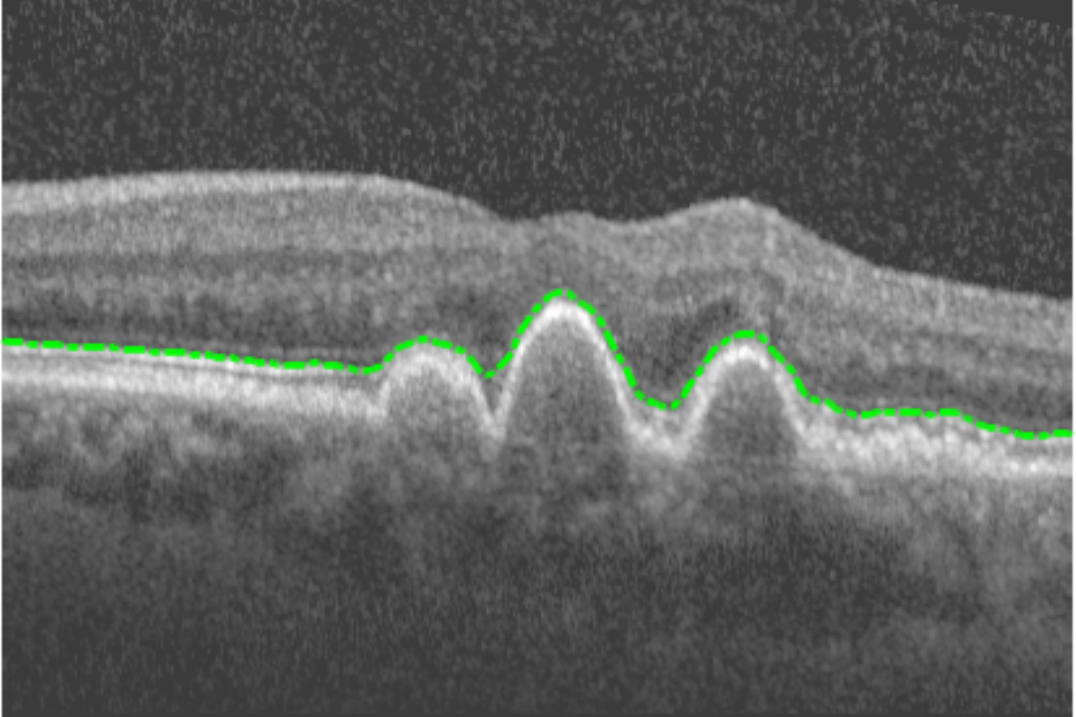}}
\vspace{-5pt}
\caption{Detecting the IS-OS boarders in the normal and pathological subjects after image enhancement via a local adaptive thresholding method (\ref{eq:threshold}).}
\label{fig:preSegImg}
\end{figure}

The intensity variation between two layers divided by the IS-OS ($B_7$) border are normally the most prominent in OCT B-scans. However, due to the fact that OCT images are always corrupted by speckle noise as a result of light absorption and scattering in the retinal tissue, it is not always the case. For example, the intensity variation around the IML ($B_1$) border sometimes can be more obvious than that around the IS-OS, as shown in the gradient image Figure~\ref{fig:gradinetMaps} (c). To make sure the segmentation of the IS-OS boundary we first enhance the IS-OS via a simple local adaptive thresholding approach\footnote{{http://homepages.inf.ed.ac.uk/rbf/HIPR2/adpthrsh.htm}}, which is given as follows
\begin{equation} \label{eq:threshold}
p = \left\{ \begin{array}{ll}
0 &ls\left( {I,ws} \right)-I > C\\
1 &\text{otherwise}
\end{array} \right.
\end{equation}
where $I$ is the input OCT image, and $ls\left( {p,ws} \right)$ means that $I$ is convolved with a suitable operator, i.e. the mean or median filter. $ws$ is the window size of the filter and $C$ a user-defined threshold value. In the paper, we use the mean filter with the window size $ws=100$ and set $C=0.01$. The enhanced image can be then obtained by multiplying the original image $I$ with $p$. The first two images in Figure~\ref{fig:preSegImg} illustrates that the contrast of the IS-OS boarder has been enhanced and the most obvious intensity variation now takes place around the IS-OS boundary. The IS-OS boundary then is detected on the dark-to-bright gradient image. Consequently, the delineated line is guaranteed to pass the IS-OS in both cases, as shown in the last two images in Figure~\ref{fig:preSegImg}.

\subsubsection{\textbf{Detection of the RPE-CH, OS-RPE and ONL-IS boundaries}}
Once the IS-OS ($B_7$) is segmented, it can be used as a reference to limit the search region for segmenting the RPE-CH ($B_9$), OS-RPE ($B_8$) and ONL-IS ($B_6$) boundaries. RPE-CH and OS-RPE are below the IS-OS and they are delineated in the following way: the RPE-CH can be extracted by applying the geodesic distance algorithm with the bright-to-dark gradient weights (\ref{eq:weights}) obtained from the region pixels below the detected IS-OS (i.e. the bright-to-dark gradient weights are set to zeros above the IS-OS); the OS-RPE is then delineated on the bright-to-dark gradient map in the region between the detected IS-OS and RPE-CH (i.e. the bright-to-dark gradient weights are set to zeros outside of the region between the IS-OS and RPE-CH). The dark-to-bright ONL-IS is above the IS-OS. The search region can be constructed between the IS-OS boundary and a parallel line above it with a diameter of 15 pixels. The dark-to-bright gradient weights outside of the region are then set to zeros. Hence, the only boundary in the search region of the dark-to-bright gradient image is the ONL-IS which can be extracted. 

\subsubsection{\textbf{Detection of the ILM and INL-OPL boundaries}}
Both the ILM ($B_1$) and INL-OPL ($B_4$) are at the border of a darker layer above a bright layer. The intensity variation around the IML boundary is much more prominent and thus the IML is segmented first. The detected ONL-IS ($B_6$) edge is taken as a reference and the dark-to-bright gradient weights below the ONL-IS is set to zeros. The ILM can then be obtained via the proposed method. The INL-OPL can then be easily detected on the dark-to-bright gradient map by simply limiting the search region between the ILM and ONL-IS (i.e. the dark-to-bright gradient values are set to zeros outside of the region between the ILM and ONL-IS). 

\subsubsection{\textbf{Detection of the OPL-ONL, IPL-INL and RNFL$_o$ boundaries}}
\begin{figure}[h!] 
\centering  
{\includegraphics[width=1\textwidth]{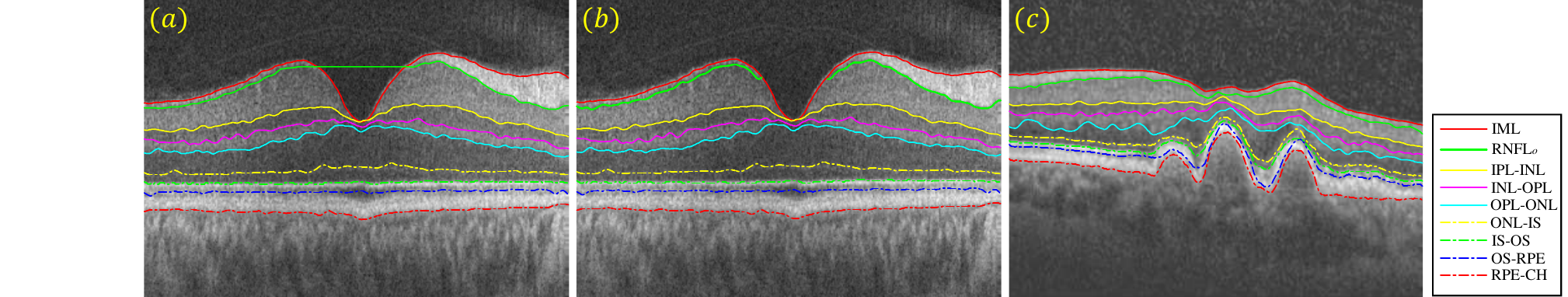}}\\
\vspace{-5pt}
\caption{The segmentation results of the nine retinal layer boundaries on both normal and dye-AMD pathological B-scans, as shown in (a) and (c). The detection of the RNFL$_o$ boundary however shows errors due to the absence of a search region for this boundary in (a). (b) shows that these errors have been corrected. }
\label{fig:SegAllLayers}
\end{figure}

\begin{figure}[h!] 
\centering  
{\includegraphics[width=0.8\textwidth, height=0.7\textwidth]{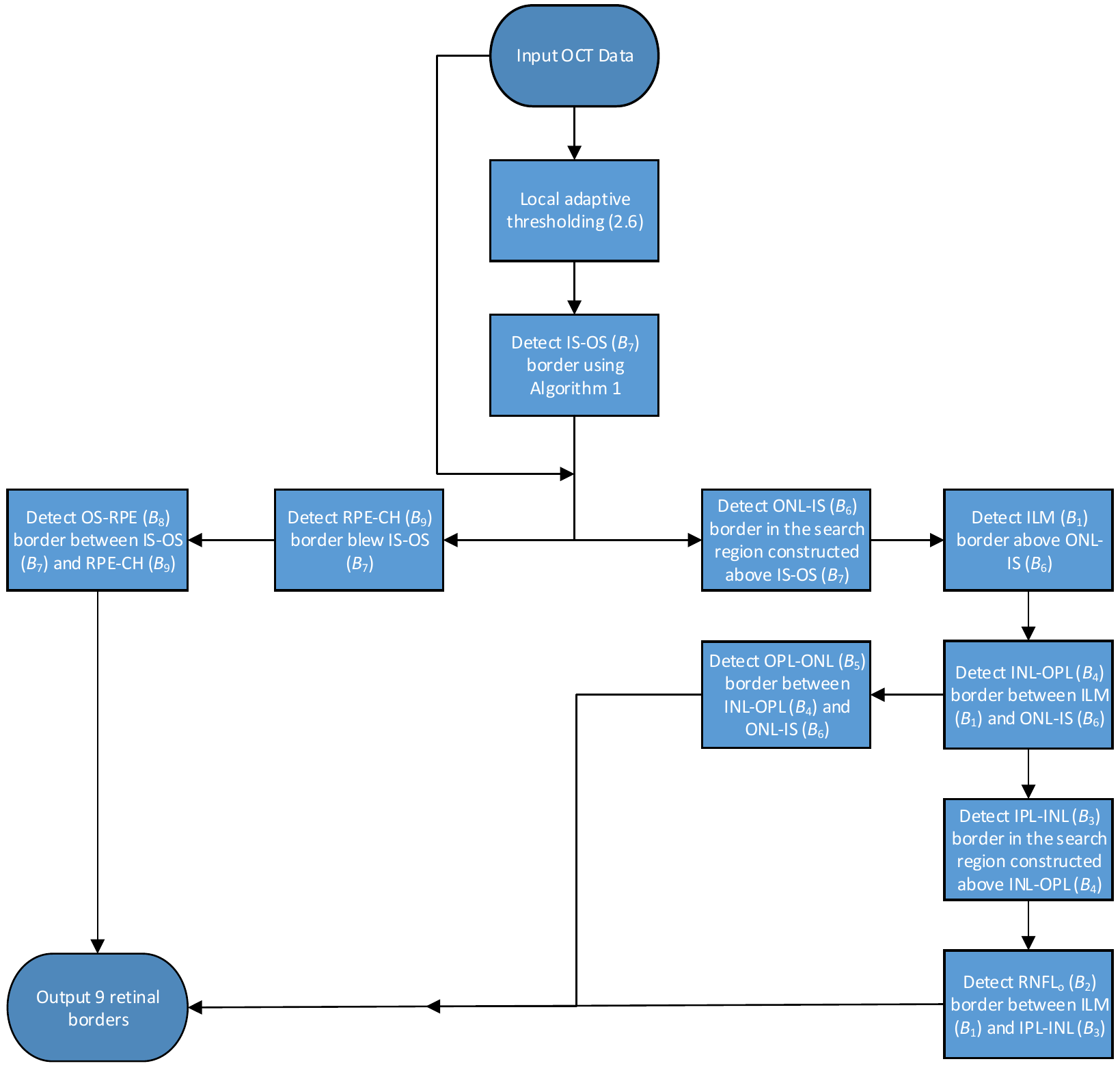}}\\
\vspace{-5pt}
\caption{The overview of the proposed framework for dynamically delineating nine retinal layer boundaries defined in Figure~\ref{fig:OCTBoundary} and Table~\ref{tb:OCTBoundary}. Section \ref{Detect9border} describes this flow chart in detail.}
\label{fig:flowChart}
\end{figure}

The OPL-ONL ($B_5$), IPL-INL ($B_3$) and RNFL$_o$ ($B_2$) demonstrate a bright layer above a darker layer and thus can be detected on the bight-to-dark gradient map defined in (\ref{eq:weights}). The segmented INL-OPL ($B_4$) and ONL-IS ($B_6$) are taken as two reference boundaries, and the OPL-ONL edge can be found by limiting the search region between the INL-OPL and ONL-IS. The search region for the IPL-INL can be then constructed between the INL-OPL boundary and a parallel line above it with a diameter of 20 pixels. The IPL-INL can be located on a bright-to-dark gradient map which is set to zeros outside of the search region constructed. Finally, the RNFL$_o$ ($B_2$) can be found in the search region between the two reference boundaries IPL-INL and IML ($B_1$). However, because the IPL-INL and IML boundaries are very close to each other in the central region of the fovea, the search region for the RNFL$_o$ are sometimes missing around the fovea region. This leads to segmentation errors of the RNFL$_o$, as shown in Figure~\ref{fig:SegAllLayers} (a). These errors however can be avoided by simply removing the spurious points detected on the RNFL$_o$ in the region above the IML, as shown in Figure~\ref{fig:SegAllLayers} (b). The proposed methods for segmenting nine retinal layer boundaries can be summarised in the flow chart as shown in Figure~\ref{fig:flowChart}. 

\section{\textbf{Experiment Setup}}
To evaluate the performance of the proposed GDM qualitatively and quantitatively, numerical experiments are conducted to compare it with the state-of-the-art approaches reviewed in Section \ref{reviewedMethod} on both healthy and pathological OCT retinal images. As the GDM is able to segment both 2D and 3D OCT images, we perform numerical experiments on both B-scan and volumetric OCT data. An anisotropic total variation \cite{goldstein2009split} is used to reduce noise prior to determining the layers boundaries/surfaces for all segmentation methods. In the following, we introduce the detailed procedure of OCT data acquisition, the evaluation metrics used to quantify the segmentation results, the final numerical results, and the computational complexity of different methods. 

\subsection{Clinical Data}
30 Spectralis SDOCT (ENVISU C class 2300, Bioptigen, axial resolution = 3.3µm, scan depth = 3.4mm, 32, 000 A-scans per second) B-scans from 15 healthy adults (mean age = 39.8 years, SD = 8.6 years; 7 male, 8 female) were used for the research. All the data was collected after informed consent was obtained and the study adhered to the tenets of the Declaration of Helsinki and Ethics Committee approval was granted. 

\textbf{2D B-scan data}: The normal vivo B-scan OCT data was imaged from the left and right eye of 15 healthy adults using a spectral domain OCT device with a chin rest to stabilise the head. The B-scan located at the foveal centre was identified from the lowest point in the foveal pit where the cone outer segments were elongated (indicating cone specialisation). To reduce the speckle noise and enhance the image contrast, every B-scan was the average of aligned images scanned at the same position. In addition to the 30 OCT images from the healthy subjects, another 20 B-scans from subjects with pathologies are also used to compare the proposed GDM with other approaches in pathological cases. These B-scans are from an eye with dry age-related macular degeneration (drye-AMD), which is available from Dufour's software package's website\footnote{{http://pascaldufour.net/Research/software$\_$data.html}}. The accuracy of segmentation results obtained by the three automated 2D methods (i.e. PDS, Chiu's method and GDM) over these healthy and pathological B-scans is evaluated using the ground truth datasets, which were manually delineated with extreme carefulness by one observer. 

\textbf{3D Volume data}: 10 Spectralis SD-OCT (Heidelberg Engineering GmbH, Heidelberg, Germany) volume data sets from 10 healthy adult subjects are used in this study. Each volume contains 10 B-scans, and the OCT A-scans outside the 6mm $\times$ 6mm (lateral $\times$ azimuth) area and centred at the fovea were cropped to remove low signal regions. All volumetric data can be downloaded from \cite{tian2015real}, where also contains the results of the \textbf{OCTRMA3D}, and the manual labellings from two graders. In this study we choose the manual labelling of grader 1 as the 3D ground truth.

\subsection{Evaluation Metrics}
Performance metrics are defined to demonstrate the effectiveness of the proposed method and compare it with the existing methods. Three commonly used measures of success for OCT boundary detection are signed error (SE), absolute error (AE) and Hausdorff distance (HD). Among them, SE indicates the bias and variability of the detection results. AE is the absolute difference between the automatic detection results and ground truth, while HD measures the distance between the farthest point of a set to the nearest point of the other and vice versa. Specifically, these metrics are denoted as 
\[\begin{array}{c}
{\rm{SE}}\left( {{B_i},{{\tilde B}_i}} \right) = \frac{1}{n}\sum\limits_{j = 1}^n {\left( {{B_{ij}} - {{\tilde B}_{ij}}} \right)} \\
{\rm{AE}}\left( {{B_i},{{\tilde B}_i}} \right) = \frac{1}{n}\sum\limits_{j = 1}^n {\left( {\left| {{B_{ij}} - {{\tilde B}_{ij}}} \right|} \right)} \\
{\rm{HD}}\left( {{B_i},{{\tilde B}_i}} \right) = \max \left( {\mathop {\max }\limits_{x \in {B_i}} \left\{ {\mathop {\min }\limits_{y \in {{\tilde B}_i}} \left\| {x - y} \right\|} \right\},\mathop {\max }\limits_{x \in {{\tilde B}_i}} \left\{ {\mathop {\min }\limits_{y \in {B_i}} \left\| {x - y} \right\|} \right\}} \right)
\end{array}\]
where $B_i$ and ${\tilde B}_i$ are respectively the detected boundaries and ground truth boundaries (i.e. manual labellings). $n$ is the number of pixels/volexs that fall on the retinal boundary/surface. Statistically, when the SE value is close to zero, the difference between $B_i$ and ${\tilde B}_i$ is small. In this case, the detection result is less bias. The measurements of AE and HD (varies from 0 to $\infty $ theoretically) signify the difference between two boundaries, e.g., 0 indicates that both retinal structures share exactly the same boundaries, and larger AE and HD values mean larger distances between the measured boundaries. We also monitor the overall SE (OSE), AE (OAE) and HD (OHD) during all the experiments. They are defined as 
\begin{align*}
\rm{OSE} &= \frac{1}{s}\sum\limits_{i = 1}^s {{\rm{SE}}\left( {{B_i},{{\tilde B}_i}} \right)} \\
\rm{OAE} &= \frac{1}{s}\sum\limits_{i = 1}^s {{\rm{AE}}\left( {{B_i},{{\tilde B}_i}} \right)} \\
\rm{OHD} &= \frac{1}{s}\sum\limits_{i = 1}^s {{\rm{HD}}\left( {{B_i},{{\tilde B}_i}} \right)} 
\end{align*}
where $s$ is the total number of retina boundaries one method can delineate. 

\subsection{Parameter Selection}
There are five parameters in the PDS model: three smooth parameters $\alpha$, $\beta$, $\varphi$ and two time step sizes $\gamma_C$ and $\gamma_b$ used within the gradient descent equations to minimise the functional (\ref{eq:PDS}) with respect to $C$ and $b$. In this study we use $\alpha=10$, $\beta=0$, $\varphi=700$, $\gamma_C=10$ and $\gamma_b \ge 2$ suggested in \cite{rossant2015parallel}. In addition, as the PDS is a nonconvex model and its segmentation results depend on initialisation. We initialise the parallel curves very closely to the true retinal boundaries for fair comparison with other methods. A maximal number of iterations number 500 is used to ensure convergence of the PDS model. The graph theoretic based methods, i.e., Chiu's method, OCTRIMA3D and Dufour's method, require no parameter input. Finally, our GDM has two build-in parameters: $\lambda$ in (\ref{eq:weights}) and $\tau$ in (\ref{eq:GradientDescentFlow}). We set $\lambda=10$ and $\tau=0.8$ to detect the retinal layers in the OCT images. 

\subsection{Numerical Results}
\label{NumericalRes}
\begin{figure}[h!] 
\centering  
{\includegraphics[height=0.9\textwidth]{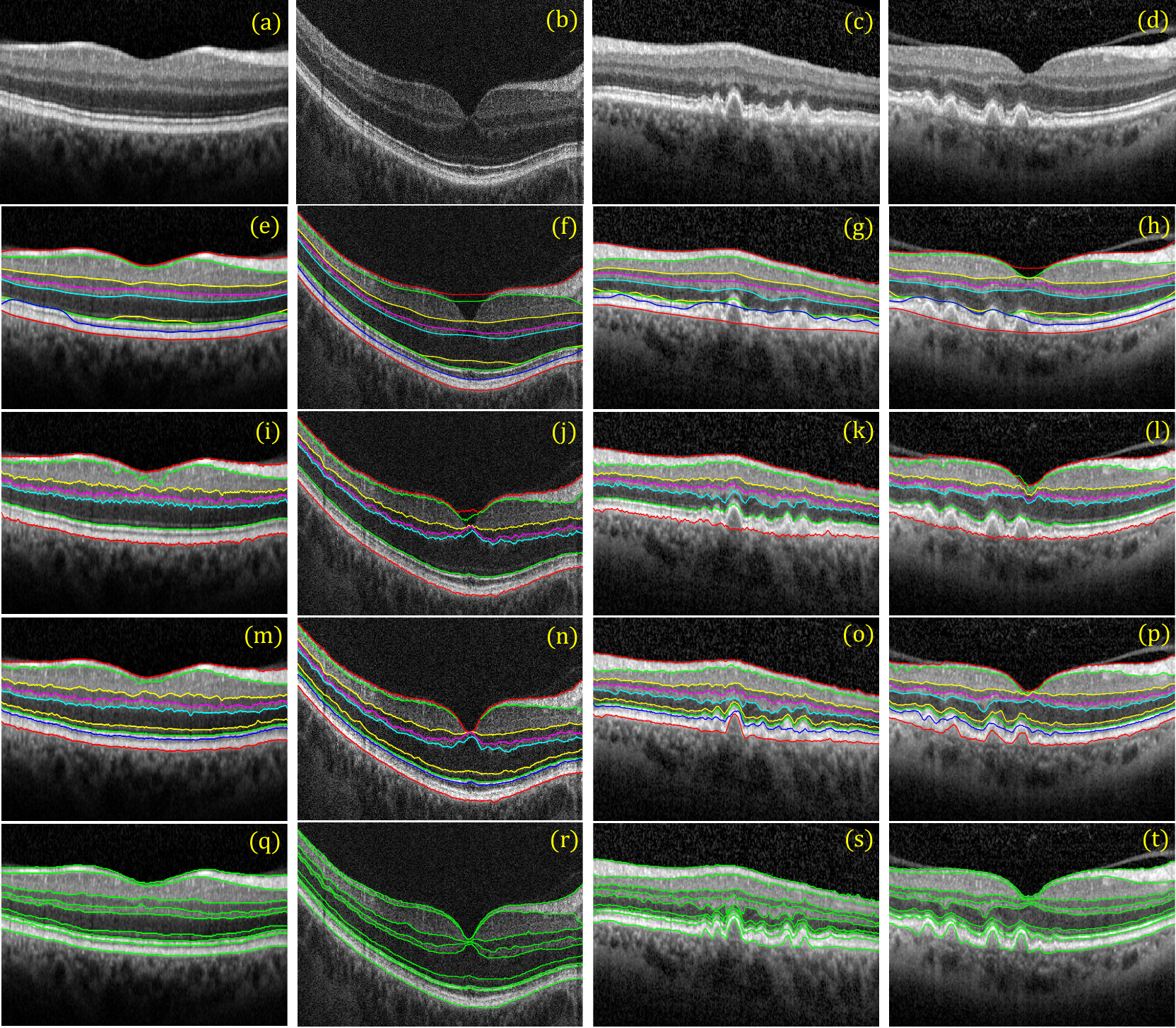}}
\vspace{-5pt}
\caption{Comparison of different segmentation methods on healthy and pathological 2D OCT B-scans. 1st row: healthy (i.e. first two) and pathological (i.e. last two) B-scans; 2nd row: results by the PDS model (\ref{eq:PDS}); 3rd row: results by Chiu's method; 4th row: results by the proposed GDM; 5th row: ground truth.}
\label{fig:SegComparison}
\end{figure}

We first visually compare the segmentation results of the proposed GDM method, the PDS (\ref{eq:PDS}) and Chiu's graph search method on both the healthy and pathological B-scans, which are shown in Figure~\ref{fig:SegComparison} (a)-(d). The PDS results as shown in (e)-(h) have some errors on some of the boundaries detected. For instance, the $B_1$ and $B_2$ cannot converge to the true retinal boundaries around the central fovea region, as shown in (f) and (h). This is because the PDS is the classical snake-driven model which has difficulty handling boundary concavity problem. Moreover, due to the fact that the $B_7$ has a much stronger image gradient than the $B_6$ and $B_8$, some parts of these two boundaries have been mistakenly attracted to the $B_7$. As Chiu's graph search method only considers the intensity changes in the pure vertical direction (\ref{eq:chiu}), it also fails segment the fovea region layers with strong curvature, as shown in (i)-(l). Moreover, the algorithm cannot handle irregular bumps caused by pathologies very well, which can be observed from the bottom $B_9$ line delineated in (k) and (l). In general, Chiu's method works very nicely when the retinal structures are flat or smooth without big changes on boundary locations. The results by the proposed GDM method, as shown in (m)-(p), performs better than the PDS and Chiu's methods when compared with the ground truth in the last row. As analysed in Section \ref{GDM}, the gradient weights defined in (\ref{eq:weights}) account for both vertical and horizontal variations, making it very suitable for both flat and nonflat retinal structures. Hence, the GDM is a better clinical tool for detecting retinal boundaries from both normal or pathological subjects.  

\clearpage

\begin{figure}[h!] 
\centering  
{\includegraphics[width=0.32\textwidth]{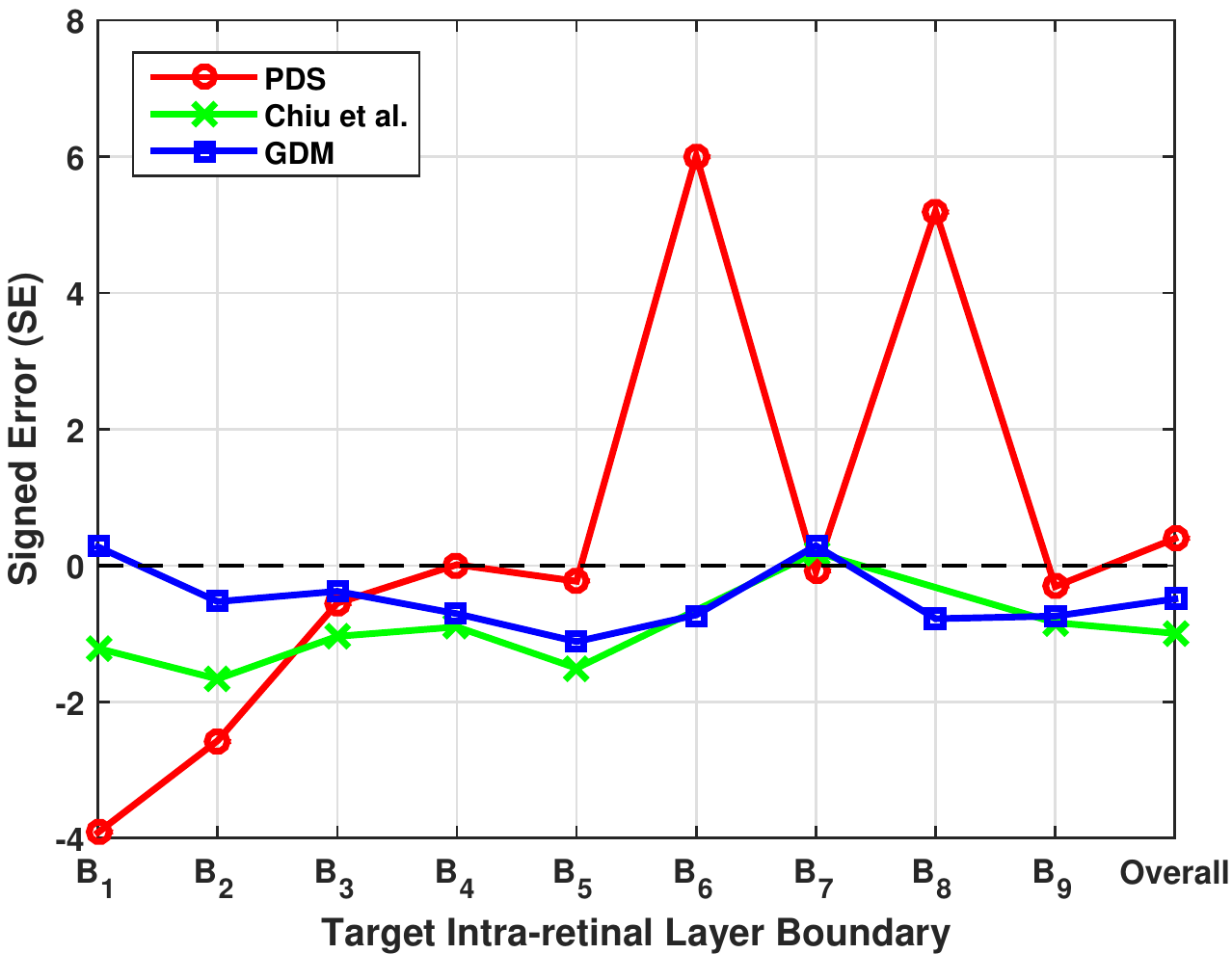}}
{\includegraphics[width=0.32\textwidth]{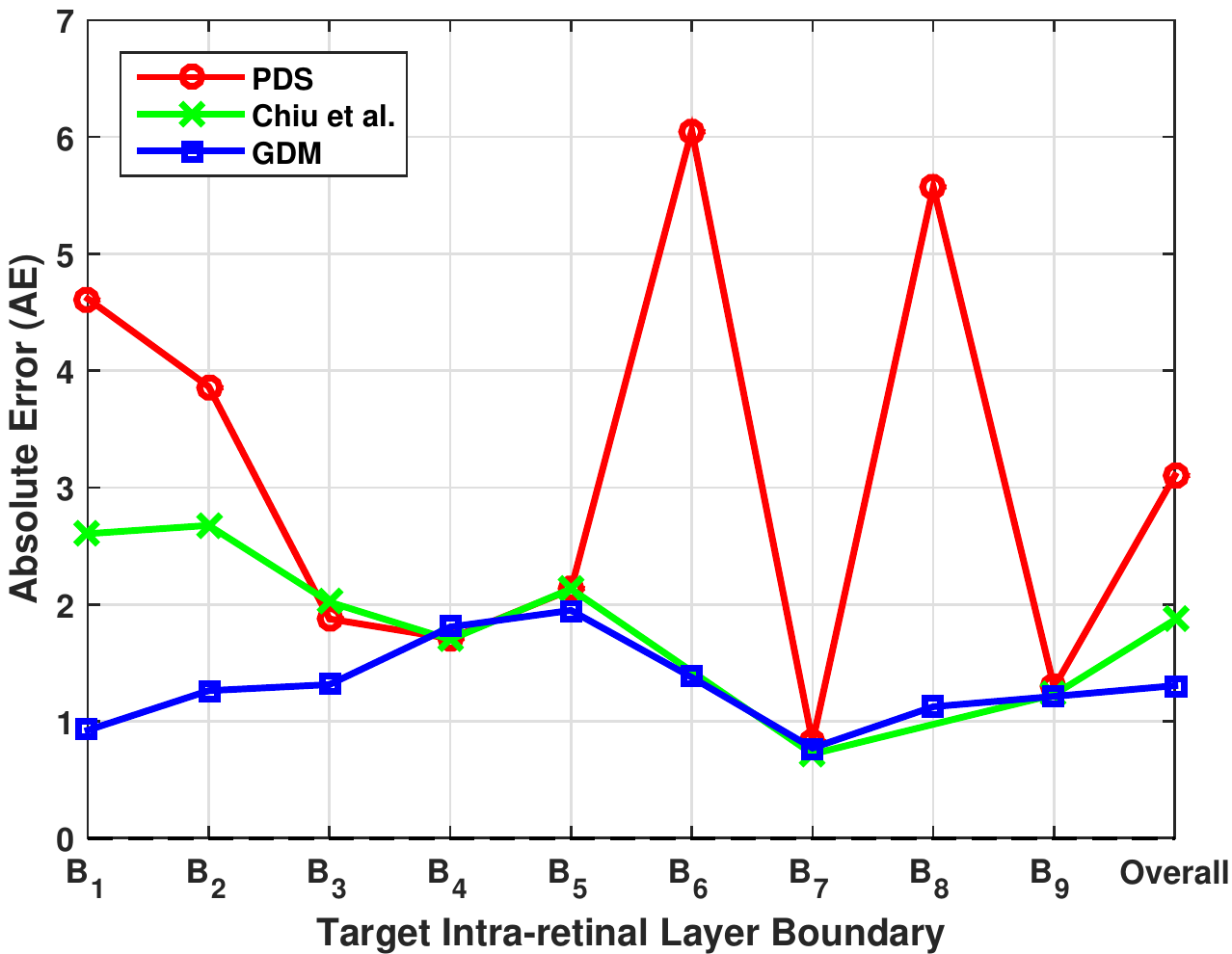}}
{\includegraphics[width=0.32\textwidth]{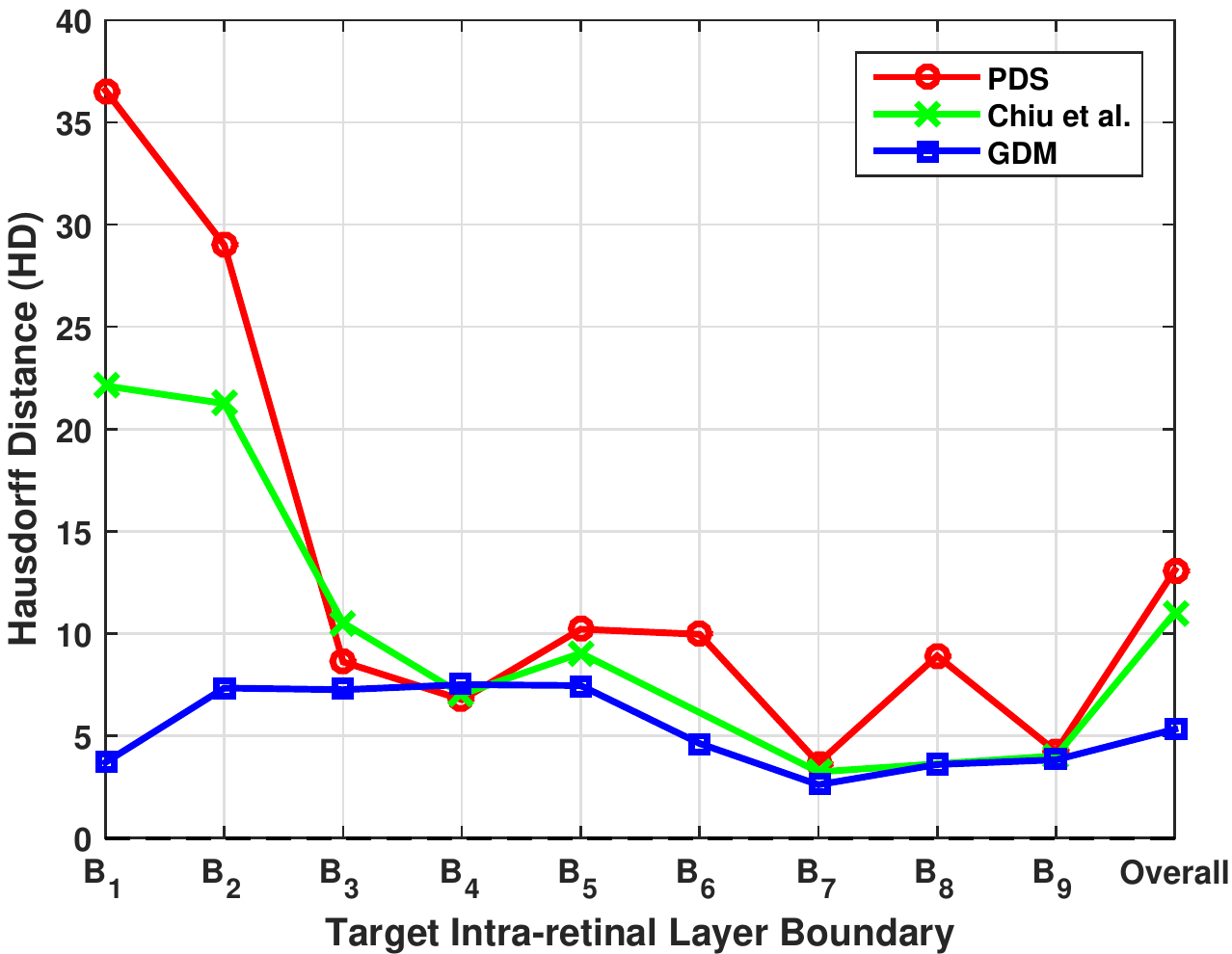}}\\
Mean value\\
\vspace{5pt}
{\includegraphics[width=0.32\textwidth]{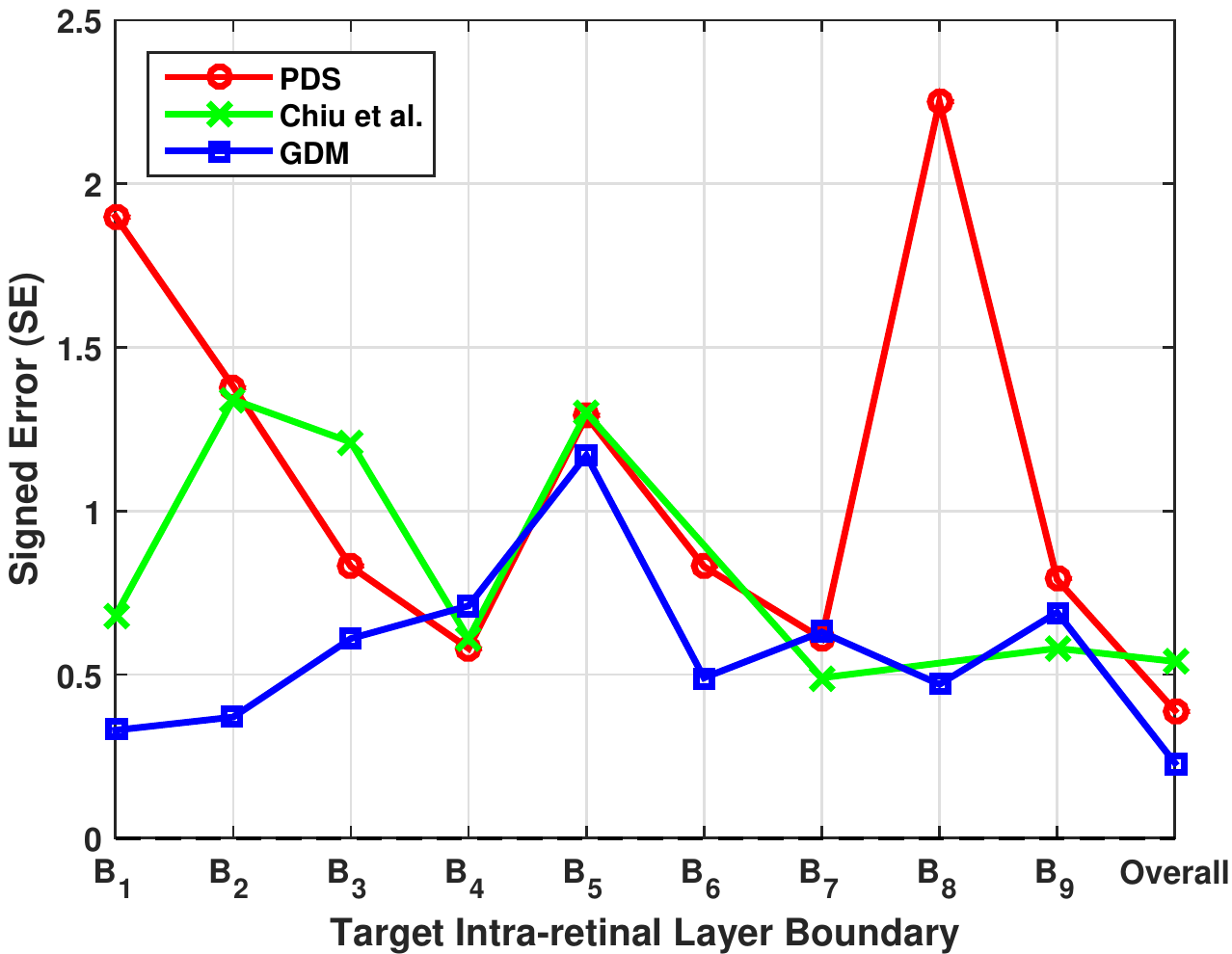}}
{\includegraphics[width=0.32\textwidth]{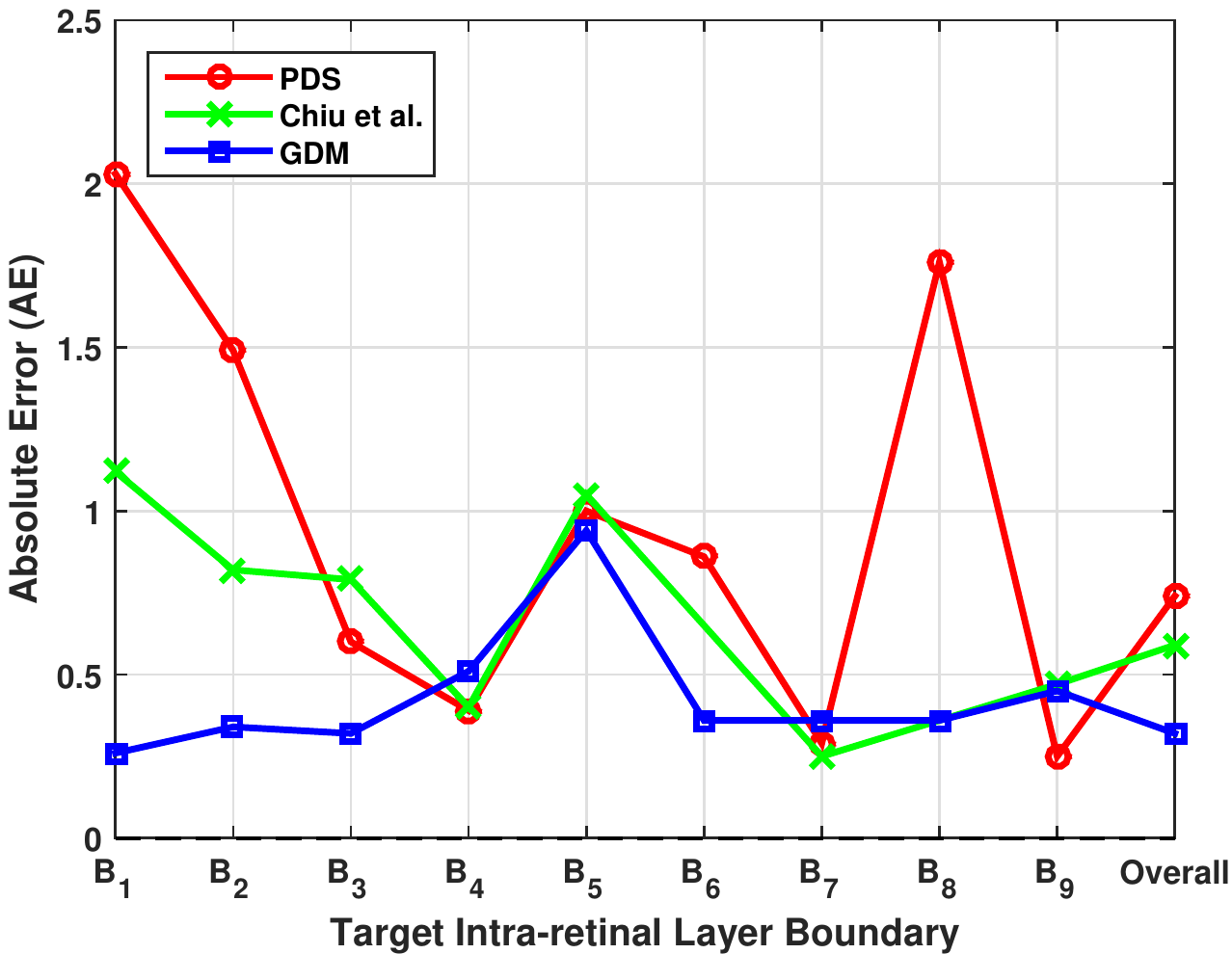}}
{\includegraphics[width=0.32\textwidth]{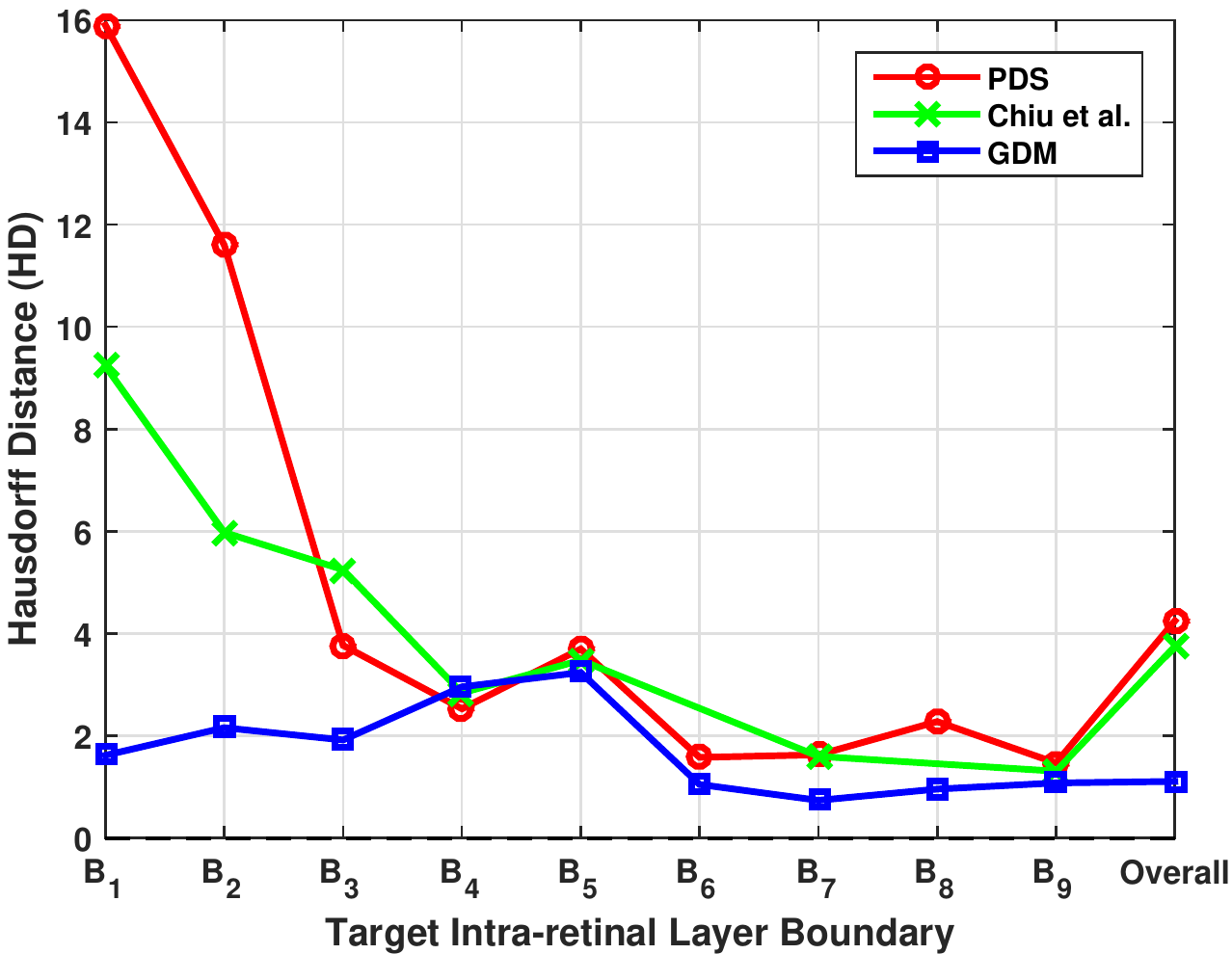}}\\
Standard deviation\\
\vspace{-5pt}
\caption{Plots of mean and standard derivation obtained by different methods in Table~\ref{tb:healthyResults} for healthy B-scans. The 1st and 2nd rows respectively denote the mean and standard derivation of the SE ($\mu m$), AE ($\mu m$) and HD ($\mu m$) for segmentation of boundary $B_1$ to $B_9$ using the PDS, Chiu's mehtod and GDM. The overall value is the average result over all boundaries. }
\label{fig:plotHealthyResults}
\end{figure}

\begin{figure}[h!] 
\centering  
{\includegraphics[width=0.32\textwidth]{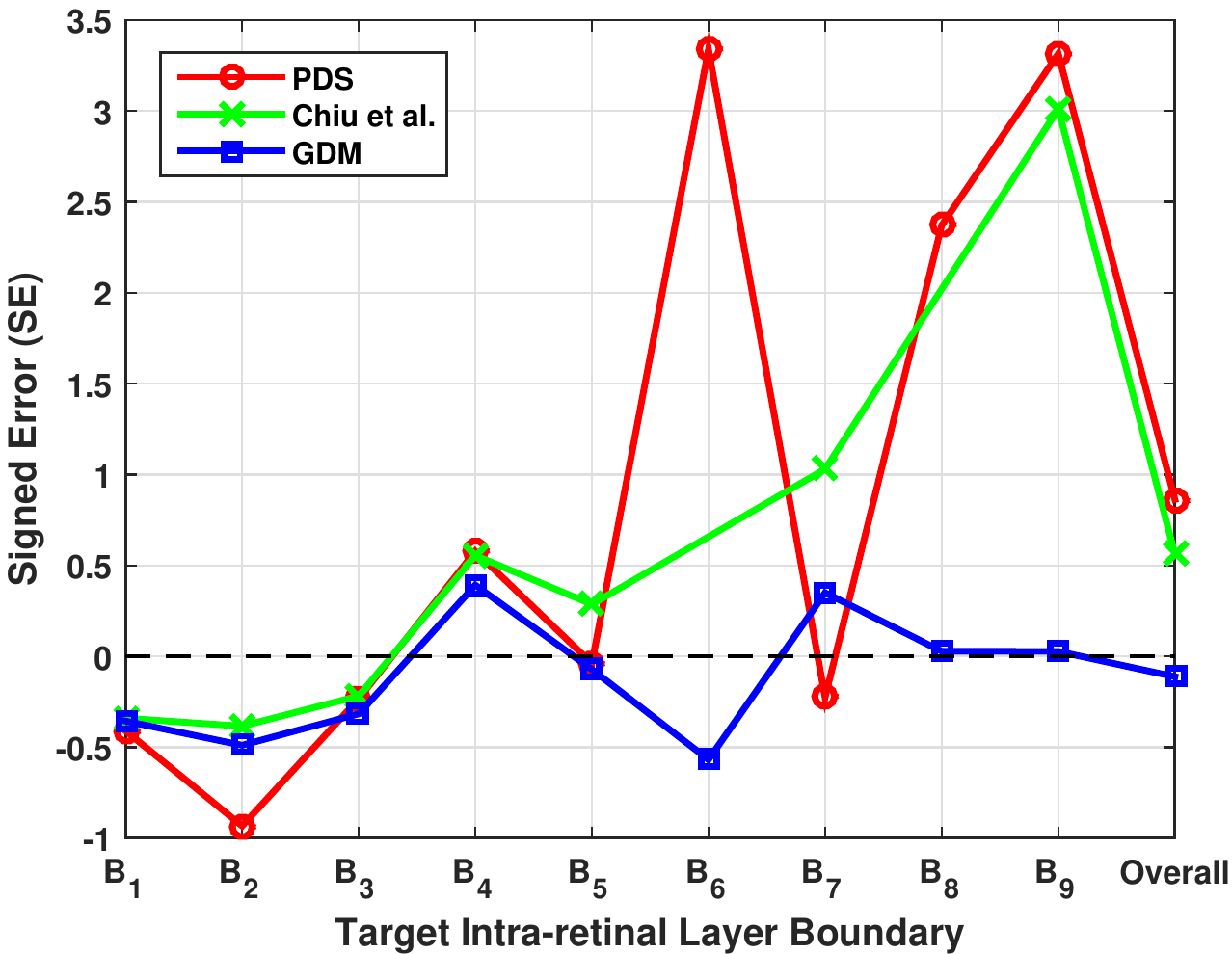}}
{\includegraphics[width=0.32\textwidth]{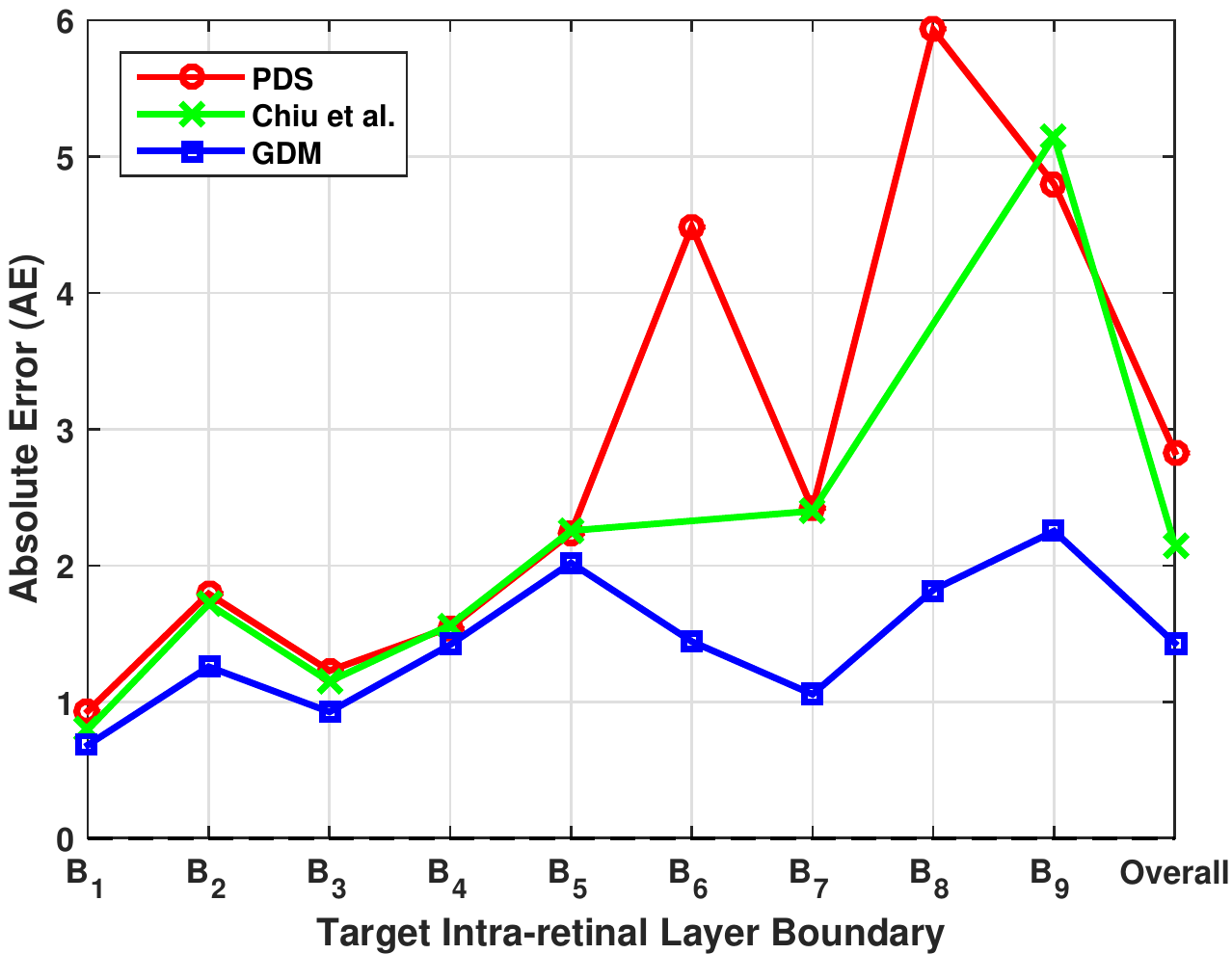}}
{\includegraphics[width=0.32\textwidth]{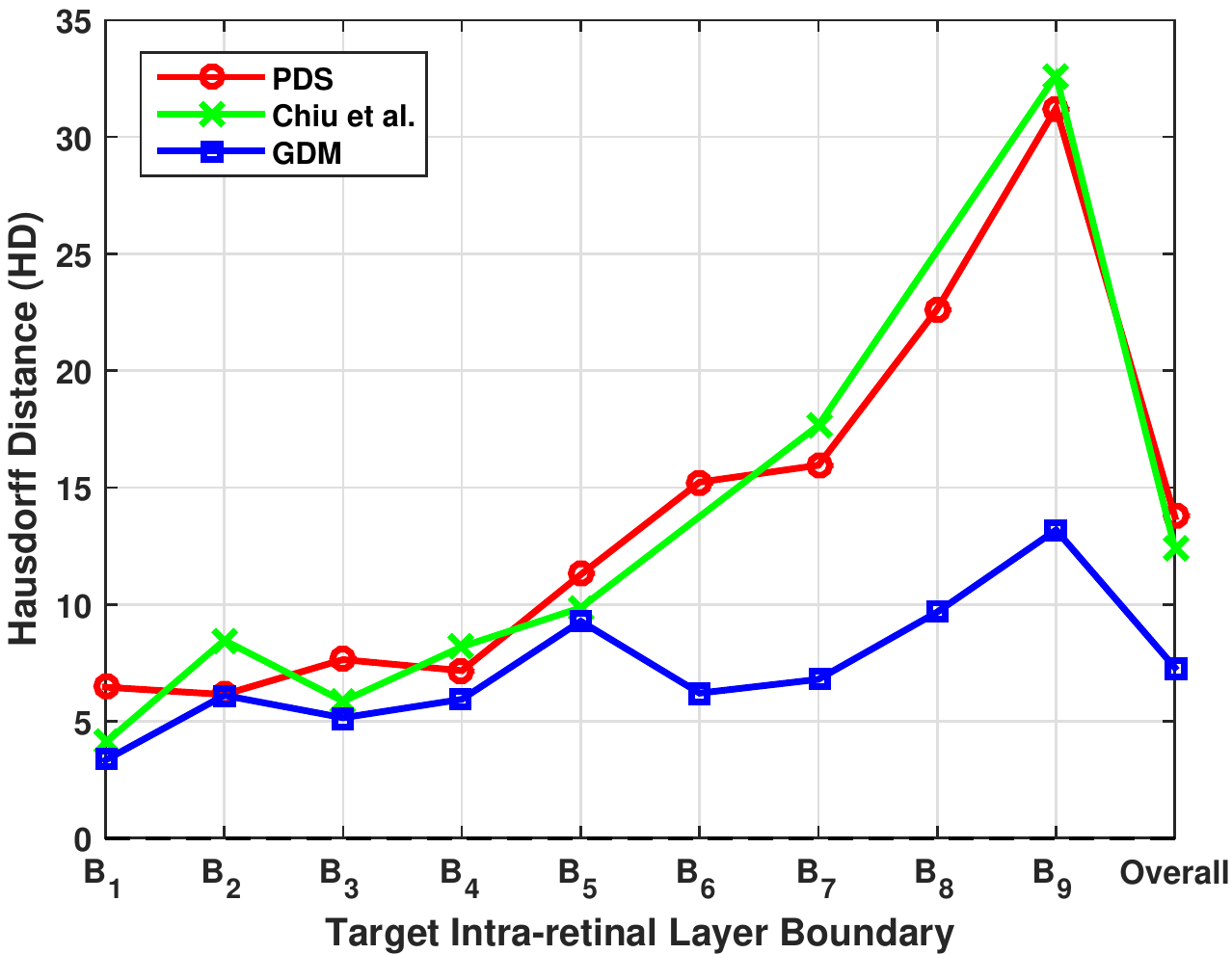}}\\
Mean value\\
\vspace{5pt}
{\includegraphics[width=0.32\textwidth]{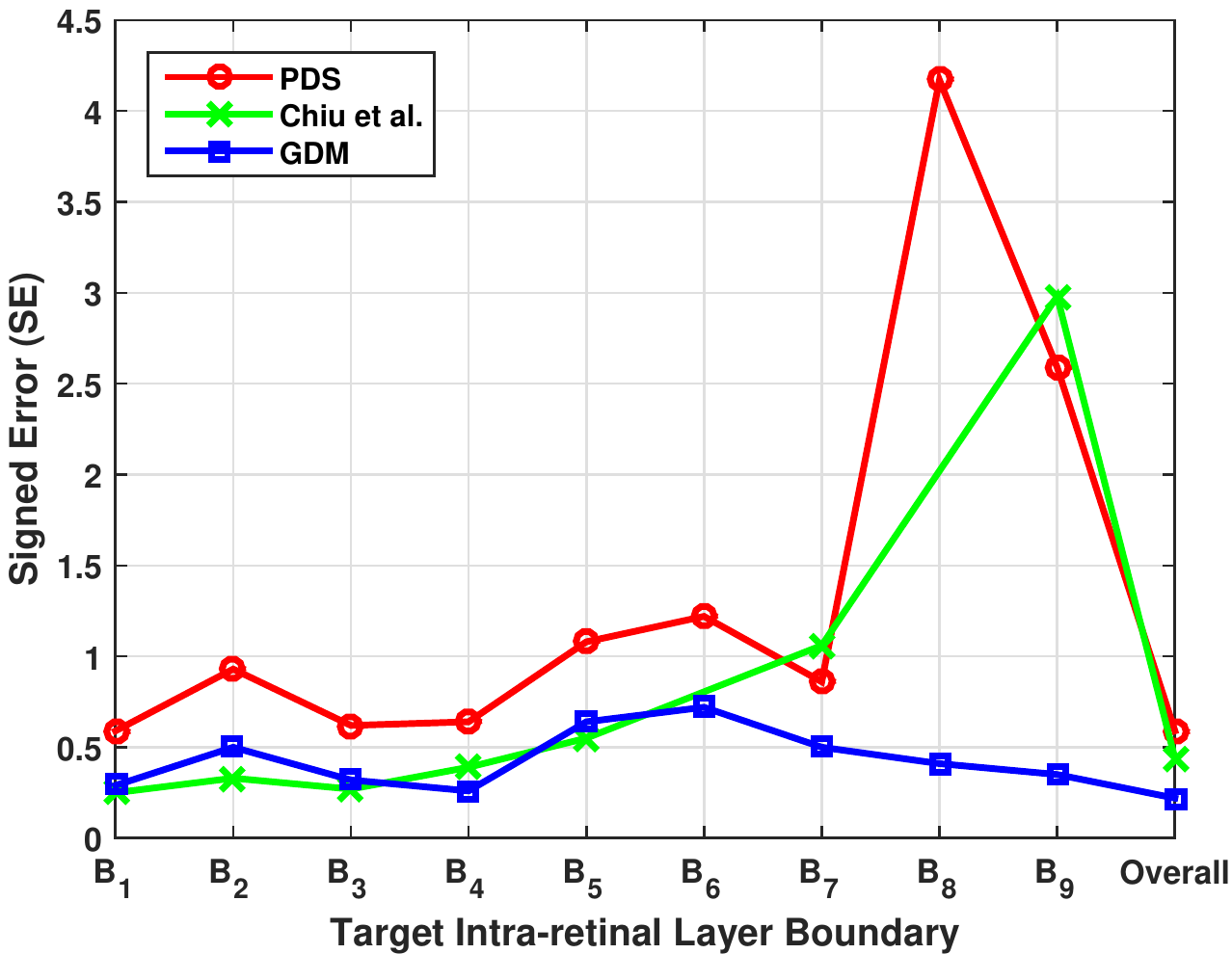}}
{\includegraphics[width=0.32\textwidth]{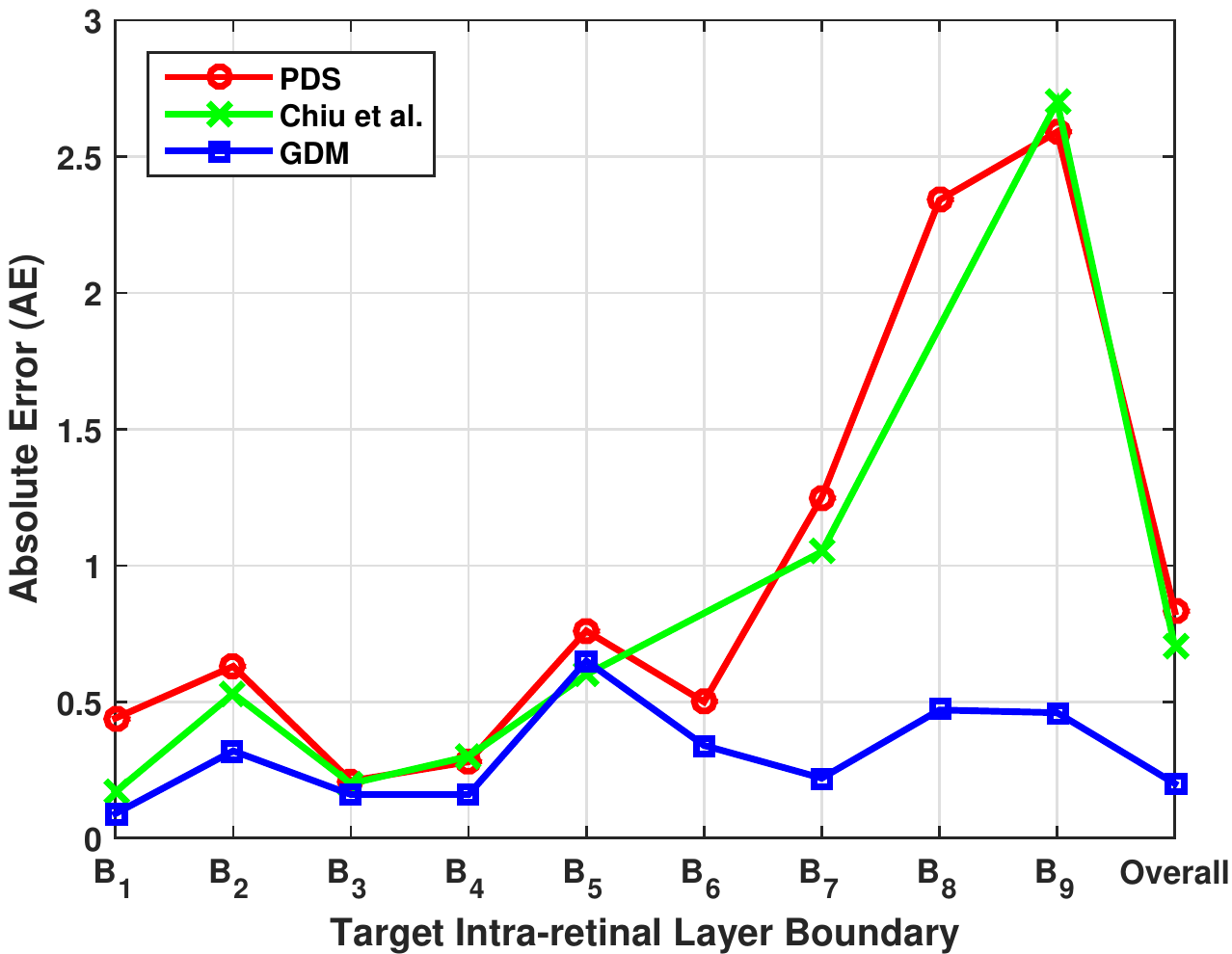}}
{\includegraphics[width=0.32\textwidth]{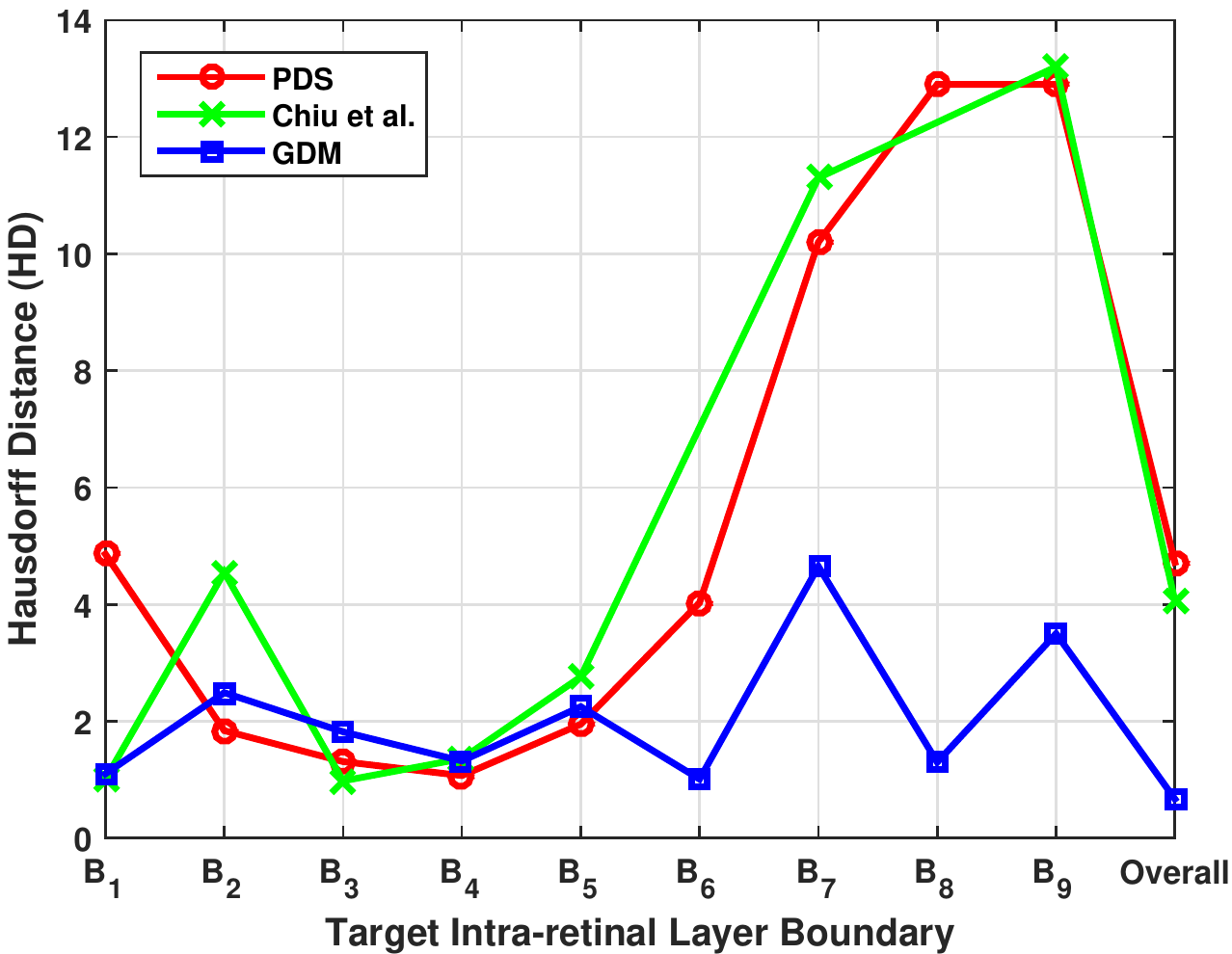}}\\
Standard deviation\\
\vspace{-5pt}
\caption{Plots of mean and standard derivation obtained by different methods in Table~\ref{tb:pathologicalResults} for pathological B-scans. The 1st and 2nd rows respectively denote the mean and standard derivation of the SE ($\mu m$), AE ($\mu m$) and HD ($\mu m$) for segmentation of boundary $B_1$ to $B_9$ using the PDS, Chiu's mehtod and GDM. The overall value is the average result over all boundaries. }
\label{fig:plotPathologicalResults}
\end{figure}

The accuracy of the segmentation results by different methods against ground truth over 30 healthy and 20 pathological B-scans is indicated in Table~\ref{tb:healthyResults} and Table~\ref{tb:pathologicalResults}, respectively. In order to make the comparison clearer, we plot the data in the two tables in Figure~\ref{fig:plotHealthyResults} and Figure~\ref{fig:plotPathologicalResults}, respectively. 

In Table~\ref{tb:healthyResults} and Figure~\ref{fig:plotHealthyResults}, the SE shows that the PDS leads to very large segmentation bias with the largest error being 6.01$\mu m$, whilst the bias of the GDM is less than 1.22$\mu m$ for all the retinal layer boundaries. Moreover, the mean SE plot of the GDM is close to zero, which means the GDM are less biased than the other two methods. Large errors of the PDS normally take place at the $B_1$, $B_2$, $B_6$ and $B_8$, which is consistent with visual inspection on the healthy scans in Figure~\ref{fig:SegComparison}. Furthermore, the mean AE quantities and plots show that the GDM performs the best for all the boundaries. Particularly at the $B_1$ and $B_2$ where the curved fovea region is located, the HD values of the GDM (3.702$\pm$1.62$\mu m$, 7.340$\pm$2.16$\mu m$) are significantly lower than those of the PDS (36.56$\pm$15.9$\mu m$, 29.00$\pm$11.6$\mu m$) and Chiu's method (22.12$\pm$9.23$\mu m$, 21.25$\pm$5.98$\mu m$). However, the accuracy of different methods are comparable at flat or smooth retinal boundaries such as $B_4$, $B_7$ and $B_9$. Finally, as the manual segmentation traces the small bumps of the true boundaries but the segmentation results by the PDS are however very smooth, the overall accuracy of the method is the lowest among all the approaches compared. 

\newcolumntype{g}{>{\columncolor[RGB]{195, 195, 195}}c}
\newcolumntype{k}{>{\columncolor{Gray}}c}
\begin{table*}[htbp] 
\caption{Mean and standard deviation of SE ($\mu m$), AE ($\mu m$) and HD ($\mu m$) calculated using the results of different methods (the PDS, Chiu' method and GDM) and the ground truth manual segmentation, over 30 healthy OCT B-scans.}
\vspace{-10pt}
\centering \setlength{\tabcolsep}{1pt}
\resizebox{\columnwidth}{!}{
\begin{tabular}{lgkcgkcgkc} 
\toprule
&\multicolumn{3}{c}{SE ($\mu m$)}  &\multicolumn{3}{c}{AE ($\mu m$)}  &\multicolumn{3}{c}{HD ($\mu m$)}\\

Boundary          &PDS                &Chiu et al.        &GDM 
                  &PDS                &Chiu et al.        &GDM 
                  &PDS                &Chiu et al.        &GDM  \\
\midrule
ILM ($B_1$)       &-3.92$\pm$1.90     &-1.22$\pm$0.68    &0.273$\pm$0.33     
				  &4.615$\pm$2.03     &2.605$\pm$1.12    &0.924$\pm$0.26	 
				  &36.56$\pm$15.9     &22.12$\pm$9.23	 &3.702$\pm$1.62     \\

RNFL$_o$ ($B_2$)  &-2.57$\pm$1.38     &-1.67$\pm$1.34    &-0.53$\pm$0.37     
				  &3.864$\pm$1.49     &2.676$\pm$0.82    &1.262$\pm$0.34	 
				  &29.00$\pm$11.6     &21.25$\pm$5.98	 &7.340$\pm$2.16 \\

IPL-INL ($B_3$)   &-0.55$\pm$0.83     &-1.04$\pm$1.21    &-0.38$\pm$0.61     
				  &1.876$\pm$0.60     &2.020$\pm$0.79    &1.314$\pm$0.32	 
				  &8.619$\pm$3.77     &10.53$\pm$5.25    &7.258$\pm$1.92       \\

INL-OPL ($B_4$)   &0.012$\pm$0.58     &-0.90$\pm$0.61    &-0.71$\pm$0.71     
				  &1.708$\pm$0.39     &1.699$\pm$0.40    &1.807$\pm$0.51	 
				  &6.772$\pm$2.53     &7.036$\pm$2.84	 &7.505$\pm$2.96      \\
				  
OPL-ONL ($B_5$)   &-0.23$\pm$1.29     &-1.51$\pm$1.30    &-1.12$\pm$1.17     
				  &2.127$\pm$1.00     &2.133$\pm$1.05    &1.949$\pm$0.94	 
				  &10.22$\pm$3.70     &9.044$\pm$3.48    &7.463$\pm$3.24       \\

ONL-IS ($B_6$)    &6.010$\pm$0.83     &---    &-0.73$\pm$0.49     
				  &6.055$\pm$0.86     &---    &1.376$\pm$0.36	 
				  &9.969$\pm$1.58     &---	 &4.630$\pm$1.05       \\

IS-OS ($B_7$)     &-0.09$\pm$0.61     &0.194$\pm$0.49    &0.291$\pm$0.63     
				  &0.823$\pm$0.29     &0.720$\pm$0.25    &0.771$\pm$0.36	 
				  &3.676$\pm$1.63     &3.240$\pm$1.60    &2.611$\pm$0.74       \\

OS-RPE ($B_8$)    &5.202$\pm$2.25     &---    &-0.78$\pm$0.47     
				  &5.570$\pm$1.76     &---    &1.125$\pm$0.36	 
				  &8.913$\pm$2.28     &---    &3.601$\pm$0.96       \\
        
RPE-CH ($B_9$)    &-0.31$\pm$0.79     &-0.84$\pm$0.58    &-0.74$\pm$0.69     
				  &1.291$\pm$0.25     &1.228$\pm$0.47    &1.213$\pm$0.45	 
				  &4.237$\pm$1.47     &4.027$\pm$1.31	 &3.831$\pm$1.08    
\\

Overall                &0.394$\pm$0.39     &-1.00$\pm$0.54    &-0.49$\pm$0.23     
				       &3.103$\pm$0.74     &1.869$\pm$0.59    &1.305$\pm$0.32	 
				       &13.11$\pm$4.25     &11.04$\pm$3.75	  &5.327$\pm$1.11    \\        
\bottomrule
\end{tabular}}
\label{tb:healthyResults}
\end{table*}

In Table~\ref{tb:pathologicalResults} and Figure~\ref{fig:plotPathologicalResults}, the mean and standard deviation plots show that the GDM is more accurate and robust compared with the other two methods for pathological data. However, larger errors have been found at the last four boundaries $B_6$, $B_7$, $B_8$ and $B_9$ for all the segmentation methods. This is because the dry age-related macular degeneration has led irregularities to these retinal boundaries, making these methods less accurate and robust. The overall accuracy measured by the three quantities has decreased compared with the corresponding measurements listed in Table~\ref{tb:healthyResults}. Chiu's graph search method using Dijkstra's algorithm can be deemed as a discrete approximation of the proposed GDM. This makes its final results comparable to those of the GDM at some flat retinal boundaries and better than those of the PDS. However, the fast sweeping algorithm used to solve the Eikonal equation guarantees local resolution for the geodesic distance, which significantly reduces the grid bias and achieves sub-pixel accuracy for the geodesic path of the GDM. In addition to the novel weight function proposed in (\ref{eq:weights}), the GDM also resolves the metrisation problem caused by discrete graph method and thus can obtain more accurate results than Chiu's method for delineating cellular layers from both normal or pathological subjects. 

\newcolumntype{g}{>{\columncolor[RGB]{195, 195, 195}}c}
\newcolumntype{k}{>{\columncolor{Gray}}c}
\begin{table*}[htbp] 
\caption{Mean and standard deviation of SE ($\mu m$), AE ($\mu m$) and HD ($\mu m$) calculated using the results of different methods (the PDS, Chiu's method and GDM) and the ground truth manual segmentation, over 20 pathological OCT B-scans.}
\vspace{-10pt}
\centering \setlength{\tabcolsep}{1pt}
\resizebox{\columnwidth}{!}{
\begin{tabular}{lgkcgkcgkc} 
\toprule
&\multicolumn{3}{c}{SE ($\mu m$)}  &\multicolumn{3}{c}{AE ($\mu m$)}  &\multicolumn{3}{c}{HD ($\mu m$)}\\

Boundary          &PDS                &Chiu et al.        &GDM 
                  &PDS                &Chiu et al.        &GDM 
                  &PDS                &Chiu et al.        &GDM  \\
\midrule
ILM ($B_1$)       &-0.41$\pm$0.59     &-0.34$\pm$0.25    &-0.36$\pm$0.29     
				  &0.932$\pm$0.44     &0.796$\pm$0.17    &0.683$\pm$0.09	 
				  &6.461$\pm$4.86     &4.087$\pm$1.01	 &3.337$\pm$1.10     \\

RNFL$_o$ ($B_2$)  &-0.93$\pm$0.93     &-0.38$\pm$0.33    &-0.49$\pm$0.50     
				  &1.792$\pm$0.63     &1.717$\pm$0.53    &1.257$\pm$0.32	 
				  &6.145$\pm$1.84     &8.464$\pm$4.55	 &6.109$\pm$2.49 \\

IPL-INL ($B_3$)   &-0.23$\pm$0.62     &-0.22$\pm$0.27    &-0.32$\pm$0.32     
				  &1.228$\pm$0.21     &1.149$\pm$0.20    &0.926$\pm$0.16	 
				  &7.640$\pm$1.31     &5.857$\pm$0.98    &5.151$\pm$1.82       \\

INL-OPL ($B_4$)   &0.578$\pm$0.64     &0.555$\pm$0.39    &0.392$\pm$0.26     
				  &1.546$\pm$0.28     &1.563$\pm$0.30    &1.419$\pm$0.16	 
				  &7.165$\pm$1.07     &8.194$\pm$1.36	 &5.942$\pm$1.32      \\
				  
OPL-ONL ($B_5$)   &-0.04$\pm$1.08     &0.286$\pm$0.55    &-0.07$\pm$0.64     
				  &2.371$\pm$0.76     &2.255$\pm$0.60    &2.019$\pm$0.65	 
				  &11.28$\pm$1.95     &9.858$\pm$2.76    &9.281$\pm$2.25       \\

ONL-IS ($B_6$)    &3.339$\pm$1.22     &---    &-0.57$\pm$0.72     
				  &4.484$\pm$0.50     &---    &1.442$\pm$0.34	 
				  &15.23$\pm$4.03     &---	 &6.205$\pm$1.01       \\

IS-OS ($B_7$)     &-0.23$\pm$0.86     &1.030$\pm$1.06    &0.350$\pm$0.50     
				  &2.415$\pm$1.25     &2.399$\pm$1.05    &1.055$\pm$0.22	 
				  &15.95$\pm$10.2     &17.66$\pm$11.3    &6.795$\pm$4.65       \\

OS-RPE ($B_8$)    &2.371$\pm$4.17     &---    &0.028$\pm$0.41     
				  &5.927$\pm$2.34     &---    &1.821$\pm$0.47	 
				  &22.63$\pm$12.9     &---    &9.673$\pm$1.30       \\
        
RPE-CH ($B_9$)    &3.315$\pm$2.59     &3.011$\pm$2.98    &0.027$\pm$0.35     
				  &4.797$\pm$2.59     &5.146$\pm$2.70    &2.252$\pm$0.46	 
				  &31.23$\pm$12.9     &32.63$\pm$13.2	 &13.19$\pm$3.50    
\\

Overall                &0.863$\pm$0.59     &0.563$\pm$0.44    &-0.11$\pm$0.22     
				       &2.832$\pm$0.83     &2.146$\pm$0.70    &1.430$\pm$0.20	 
				       &13.75$\pm$4.72     &12.39$\pm$4.06	  &7.300$\pm$0.67    \\        
\bottomrule
\end{tabular}
}
\label{tb:pathologicalResults}
\end{table*}

\begin{figure}[h!] 
\centering  
{\includegraphics[width=0.72\textwidth]{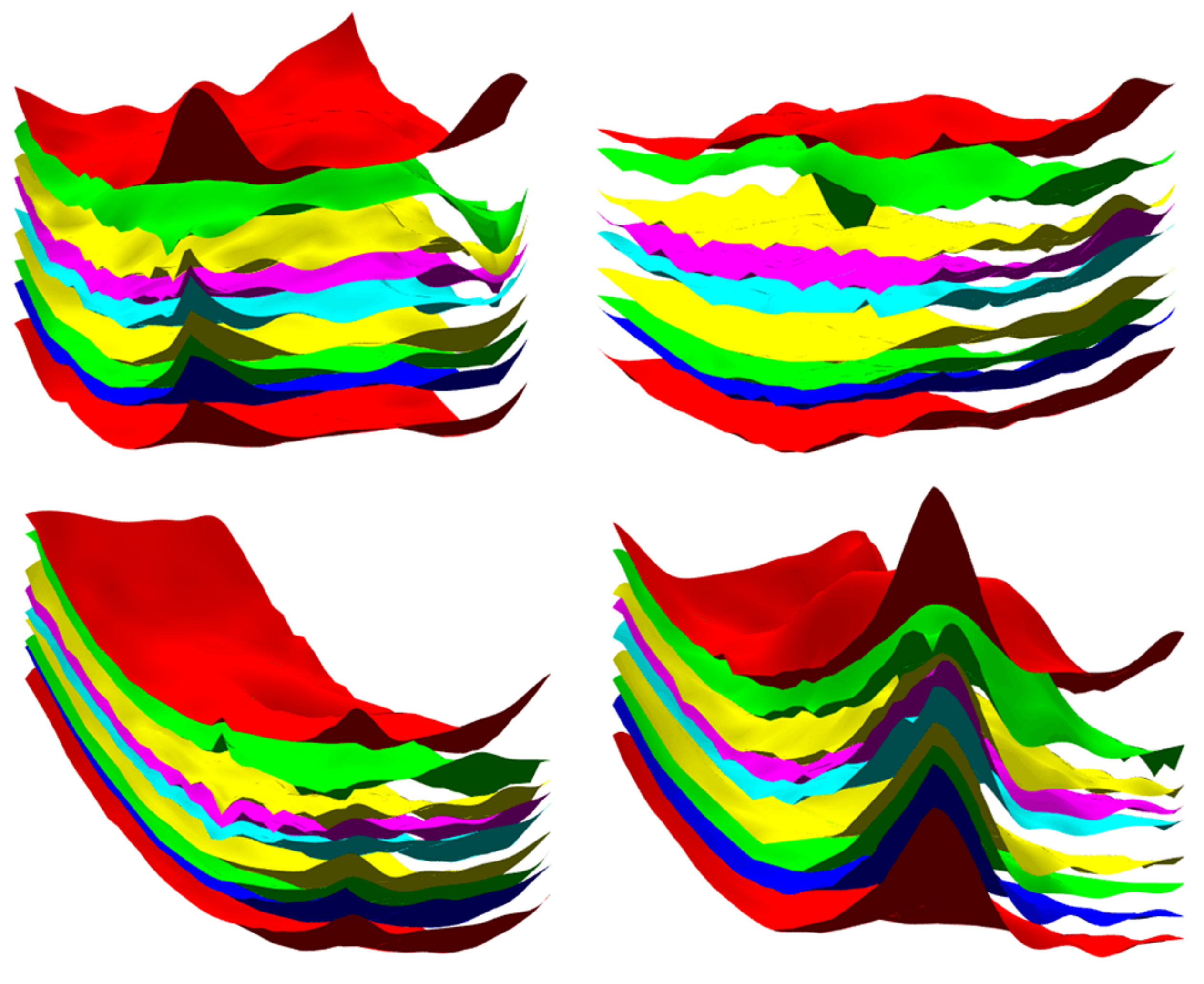}}
\vspace{-5pt}
\caption{3D rendered images of human in vivo intra-retinal layer surfaces obtained through segmenting Spectralis SD-OCT volumes with the proposed GDM method. Samples are named Volume 1, Volume 2, Volume 7 and Volume 9. The color used for each individual retinal surface is the same as in Figure~\ref{fig:OCTBoundary}.}
\label{fig:3D4Smaple}
\end{figure}

In the next section, the proposed GDM is used to segment the OCT volume dataset that includes samples from ten healthy adult subjects, named as Volume 1 to 10 respectively. Dufour's and OCTRIMA3D methods are also used to segment the same dataset for comparison purposes. In Figure~\ref{fig:3D4Smaple}, we demonstrate four representative segmentation results of GDM on Volume 1, 2, 7 and 9.

\begin{figure}[h!] 
\centering  
{\includegraphics[width=0.65\textwidth]{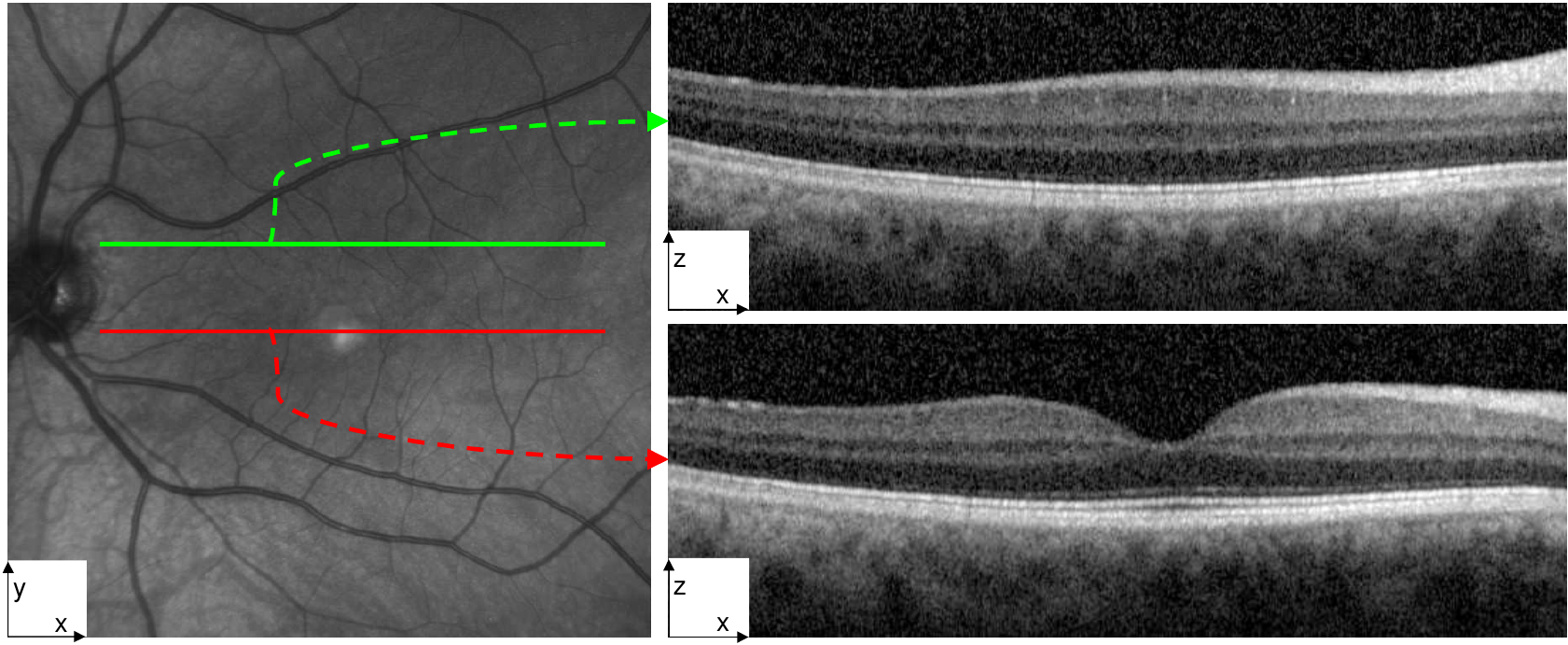}}\\
\vspace{-5pt}
\caption{Two B-scans extracted from the Volume 4 sample. The left shows the en-face representation of the OCT scan with two lines (green and red) overlaid representing the corresponding locations of two B-scans within a volume present in the right.}
\label{fig:enfaceBscans}
\end{figure}

\begin{figure}[h!] 
\centering  
{\includegraphics[width=0.32\textwidth, height=0.15\textwidth]{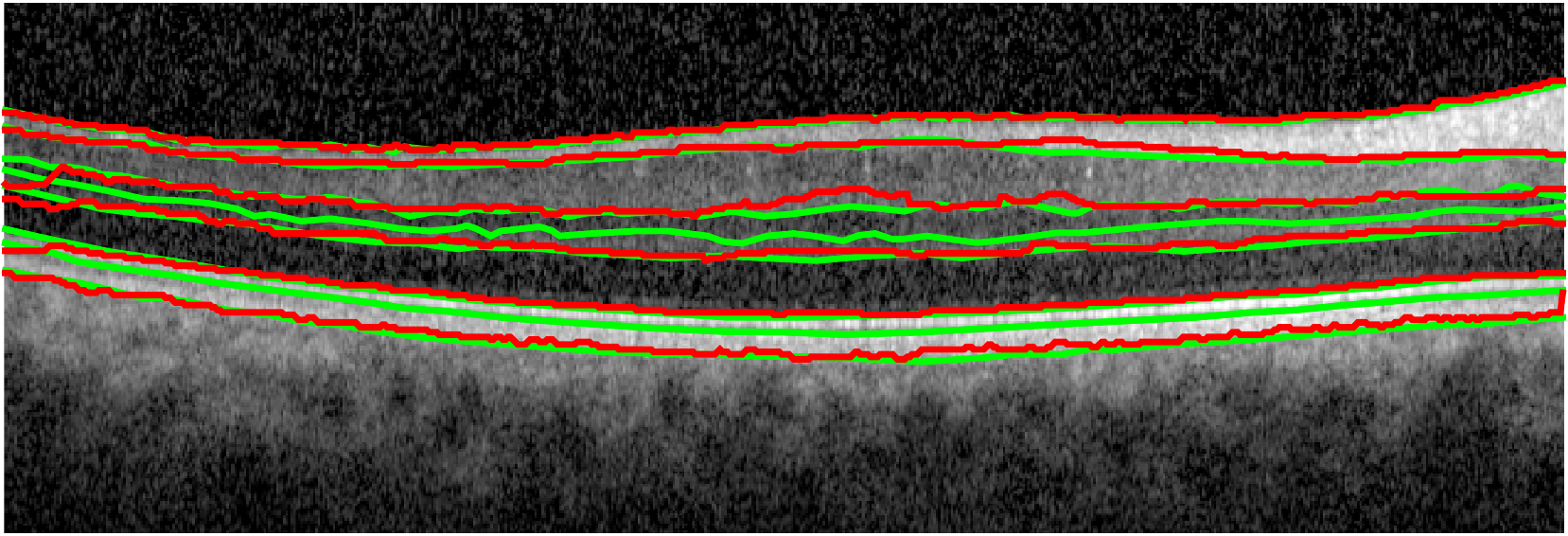}}
{\includegraphics[width=0.32\textwidth, height=0.15\textwidth]{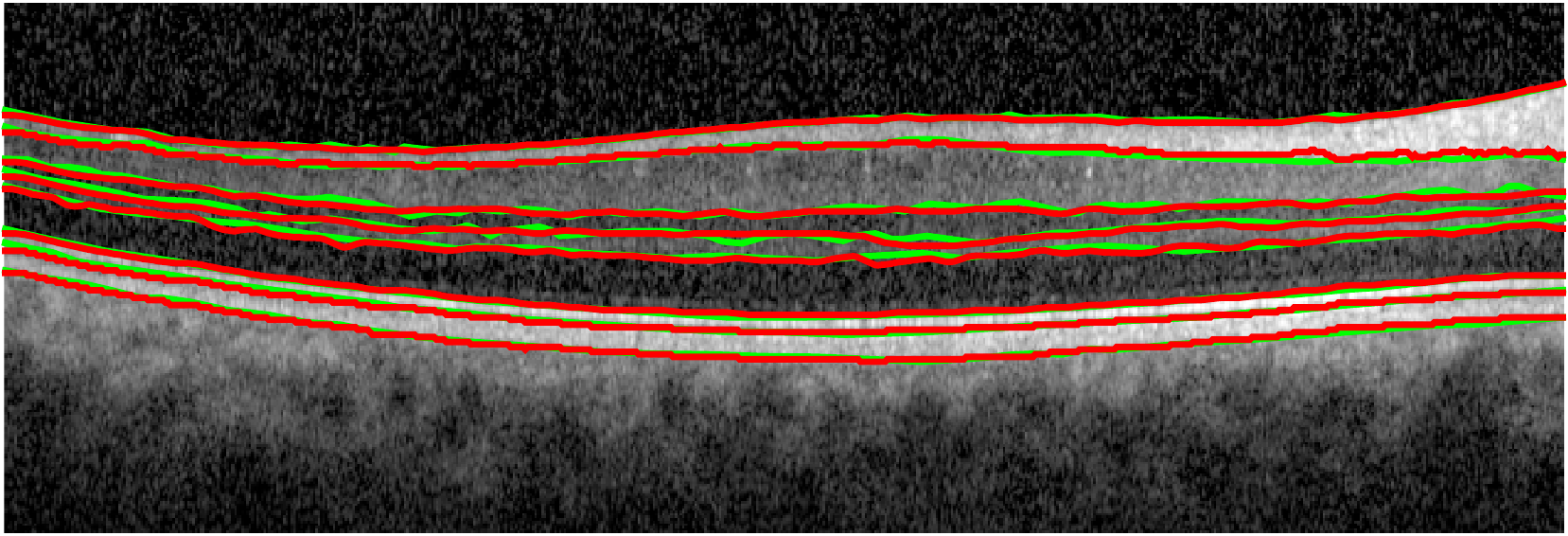}}
{\includegraphics[width=0.32\textwidth, height=0.15\textwidth]{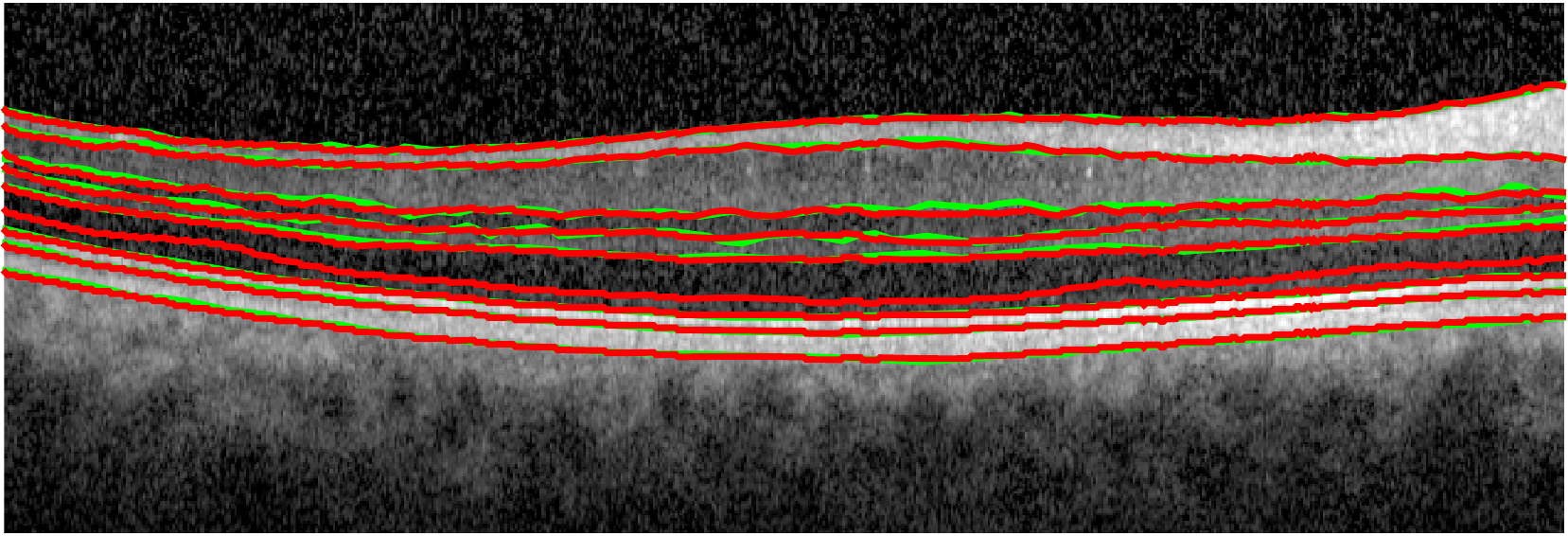}}\\
{\includegraphics[width=0.32\textwidth, height=0.15\textwidth]{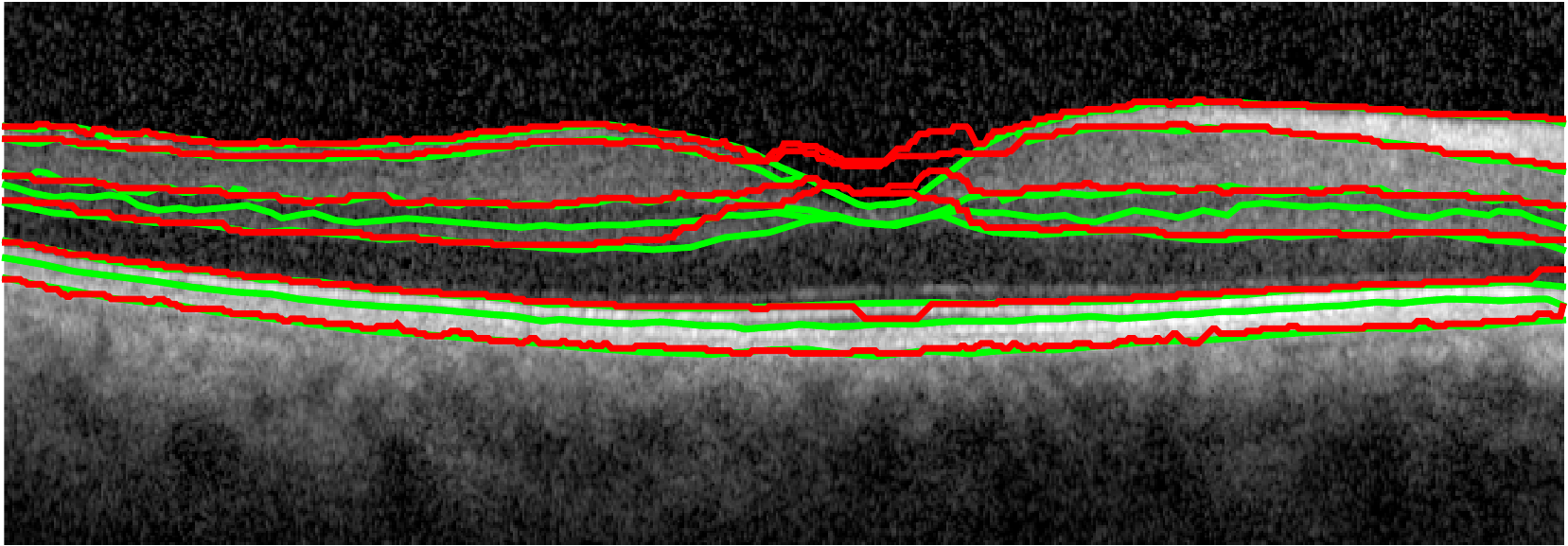}}
{\includegraphics[width=0.32\textwidth, height=0.15\textwidth]{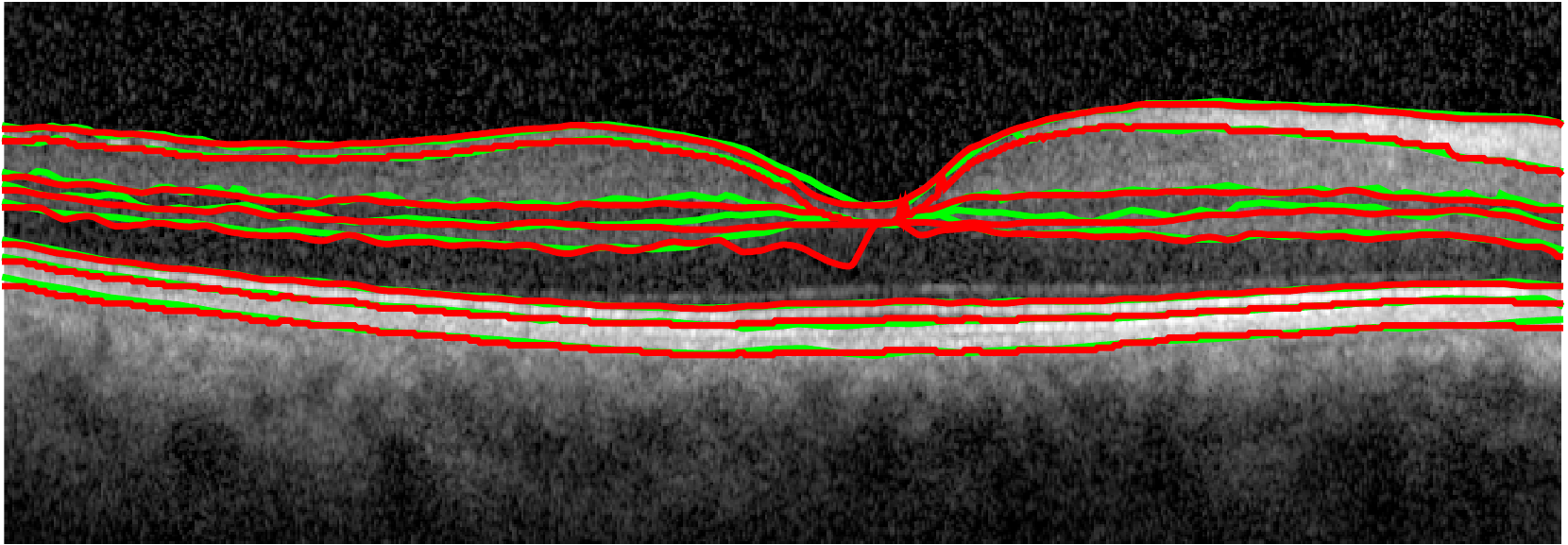}}
{\includegraphics[width=0.32\textwidth, height=0.15\textwidth]{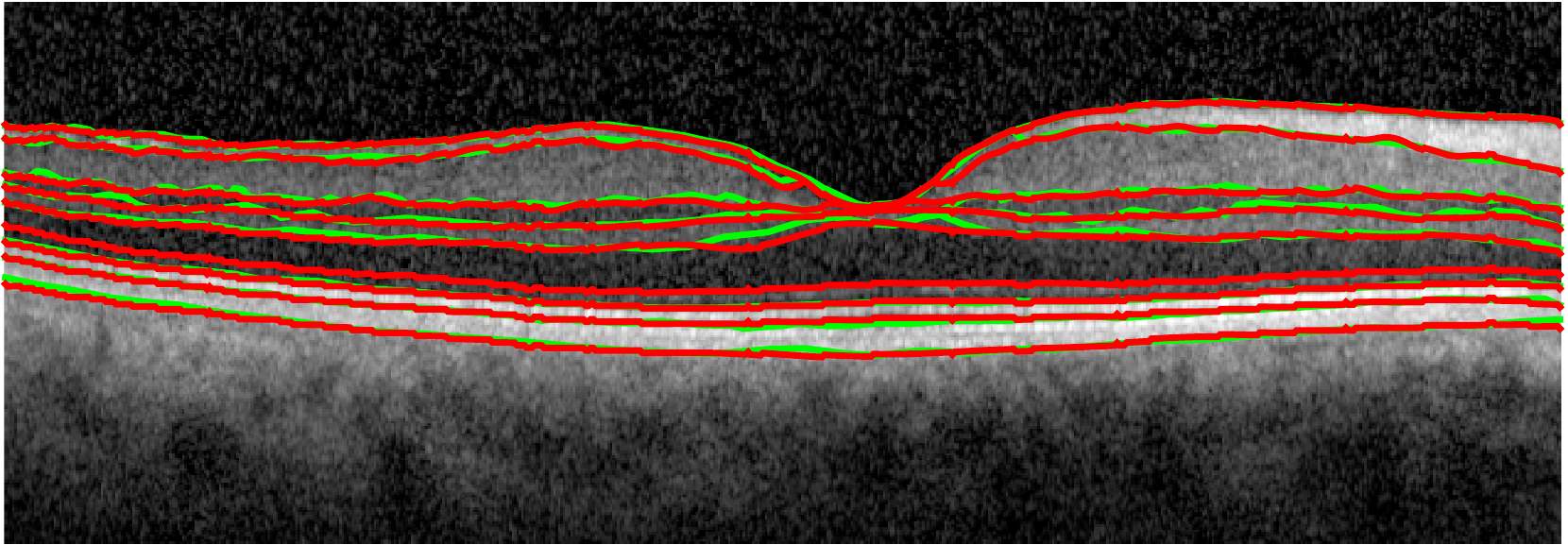}}\\
\vspace{-5pt}
\caption{The comparison between Dufour's method (left), OCTRIMA3D (middle) and GDM (right) on the two B-scans in Figure~\ref{fig:enfaceBscans}. The segmentation results by these methods are marked with red lines while the ground truth using manual labelling with green lines.}
\label{fig:2SlicesSeg}
\end{figure}

\begin{figure}[h!] 
\centering  
{\includegraphics[width=0.135\textwidth, height=0.085\textwidth]{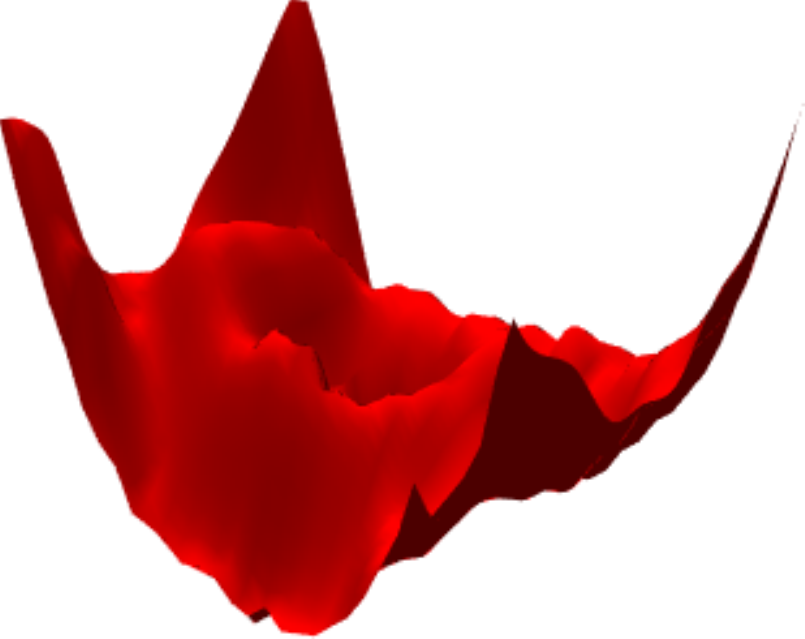}}
{\includegraphics[width=0.135\textwidth]{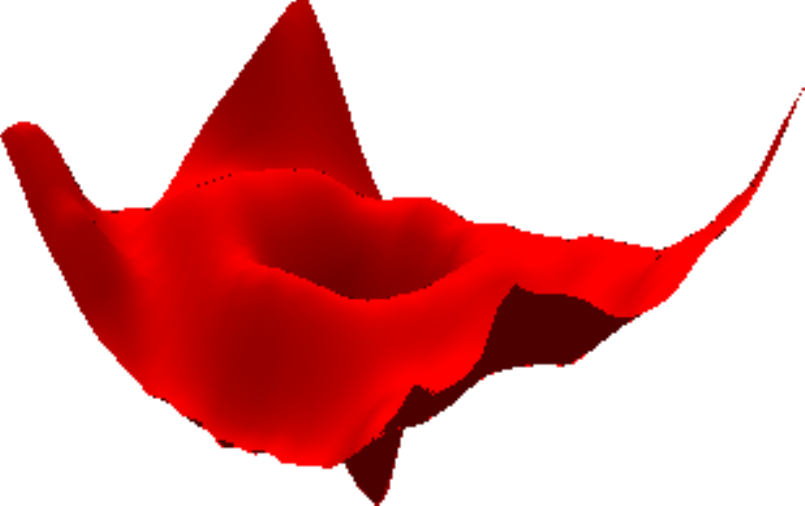}}
{\includegraphics[width=0.135\textwidth]{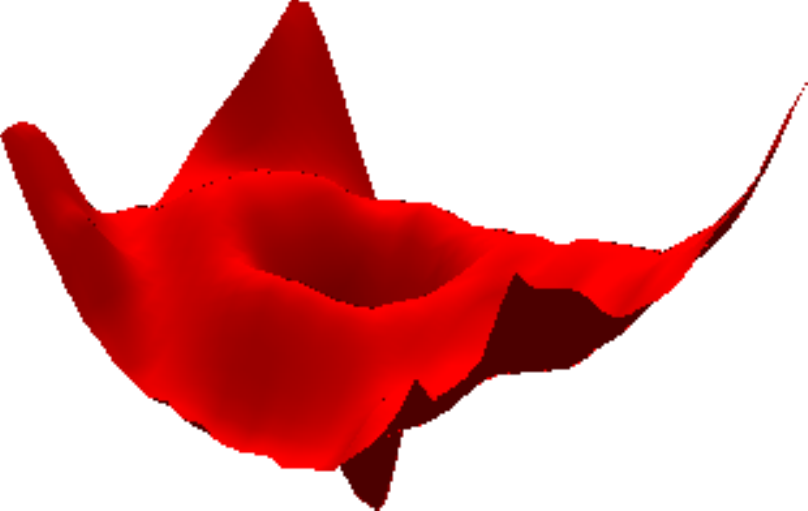}}
{\includegraphics[width=0.135\textwidth]{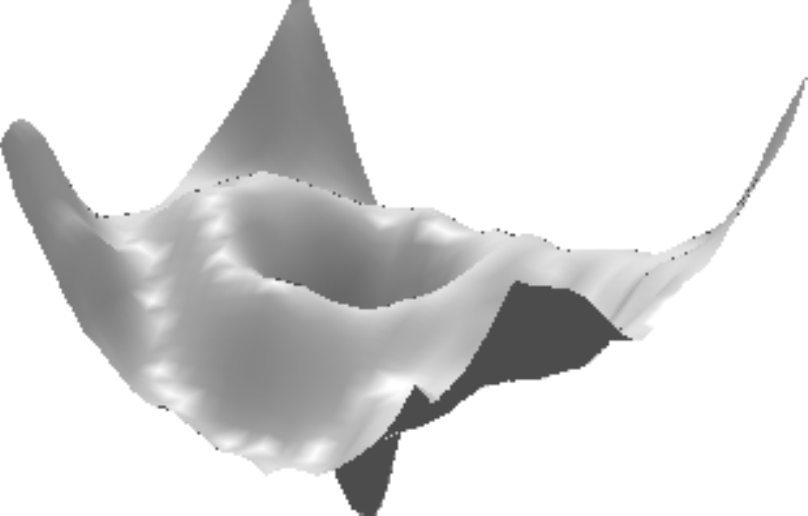}}
{\includegraphics[width=0.135\textwidth]{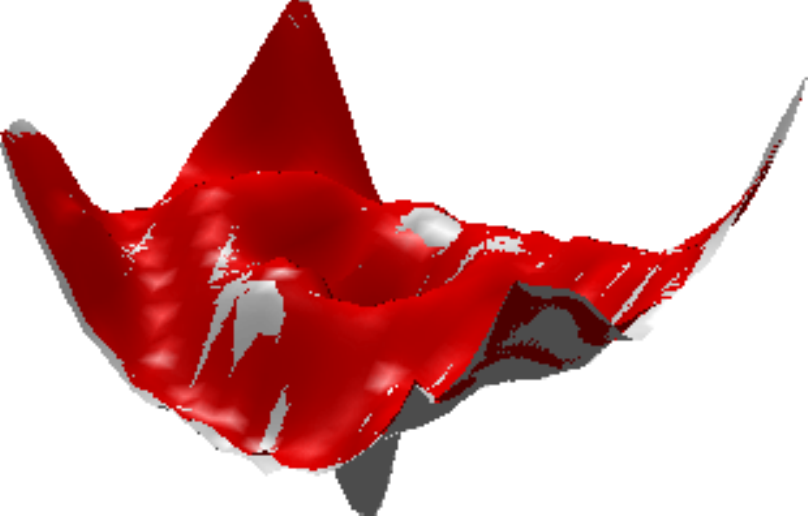}}
{\includegraphics[width=0.135\textwidth]{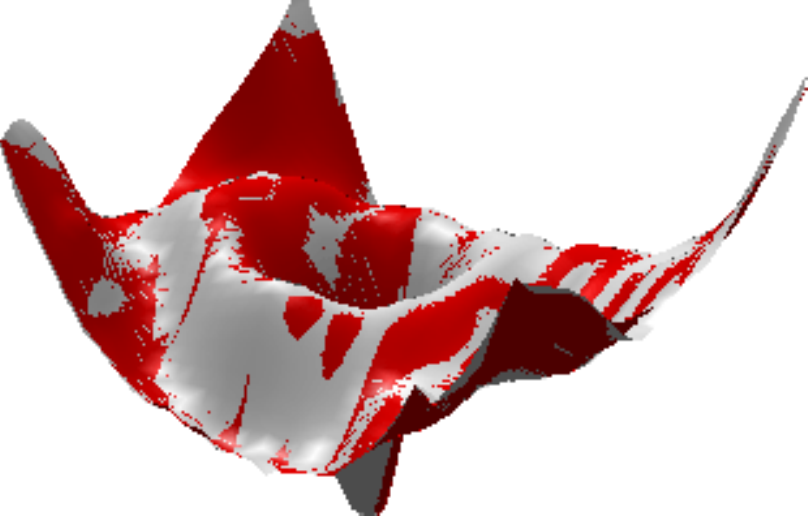}}
{\includegraphics[width=0.135\textwidth]{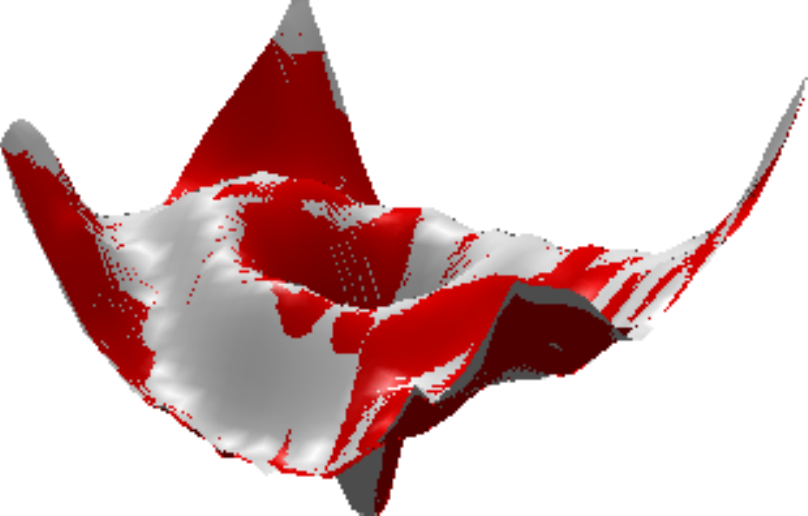}}
\\
{\includegraphics[width=0.135\textwidth, height=0.07\textwidth]{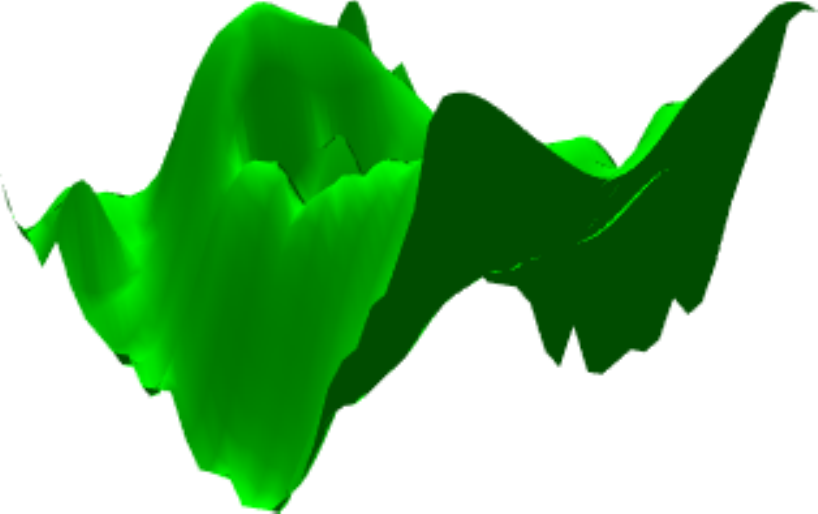}}
{\includegraphics[width=0.135\textwidth]{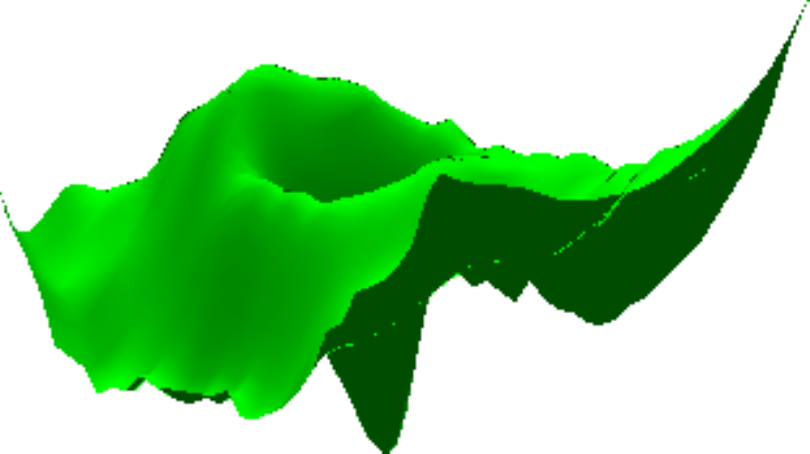}}
{\includegraphics[width=0.135\textwidth]{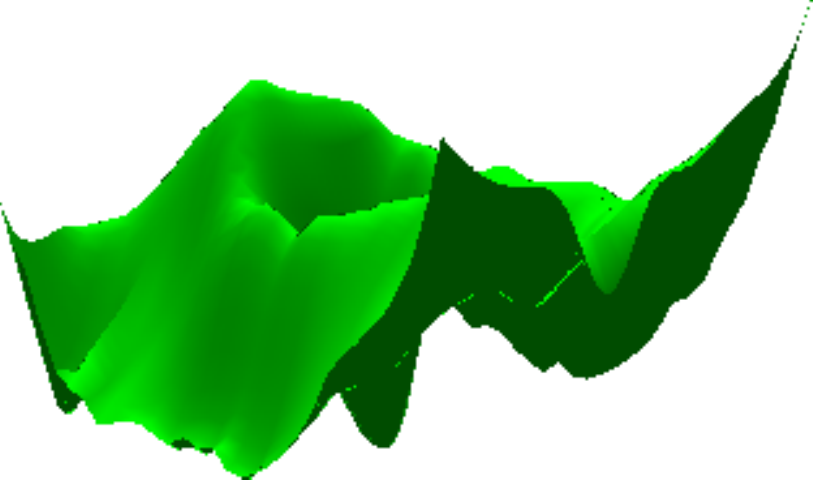}}
{\includegraphics[width=0.135\textwidth]{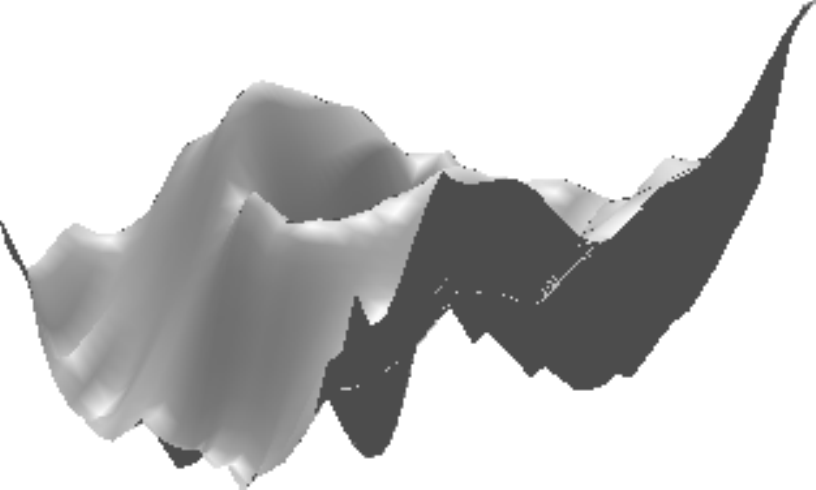}}
{\includegraphics[width=0.135\textwidth]{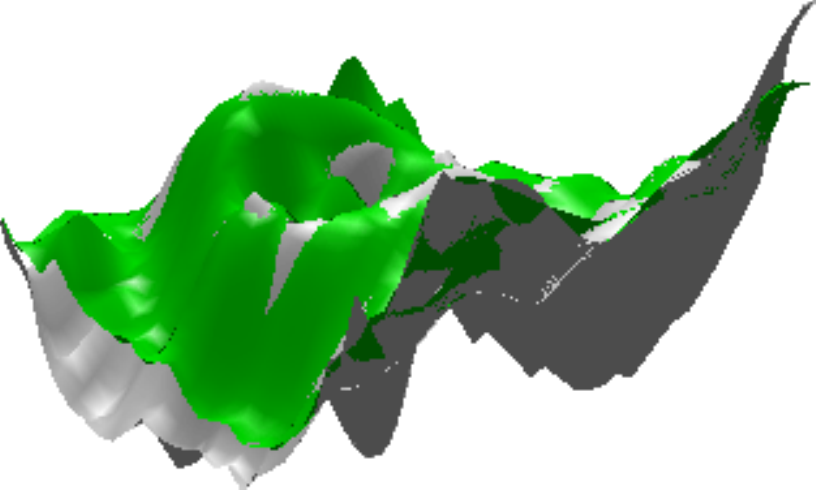}}
{\includegraphics[width=0.135\textwidth]{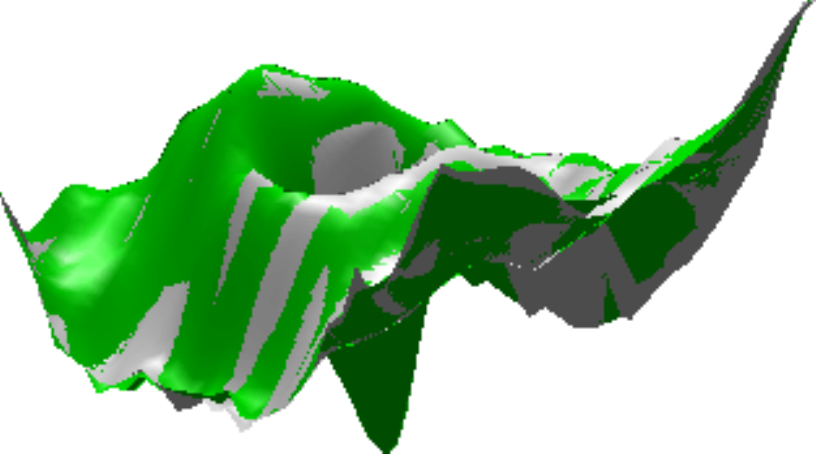}}
{\includegraphics[width=0.135\textwidth]{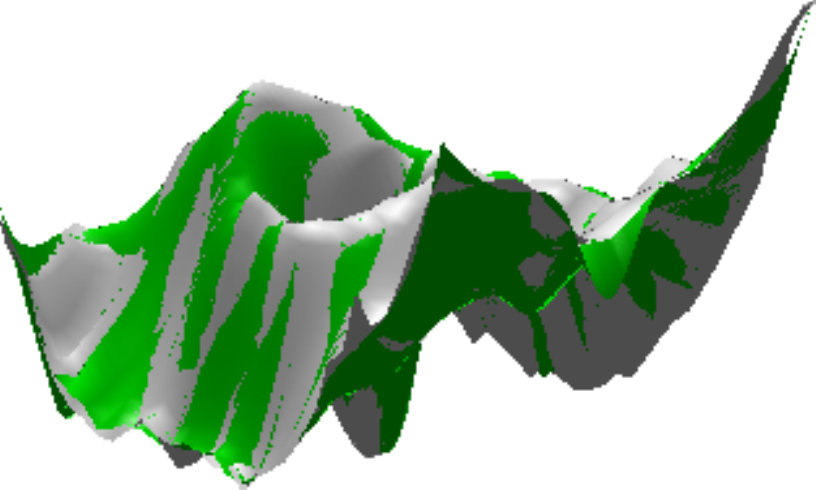}}
\\
{\includegraphics[width=0.135\textwidth]{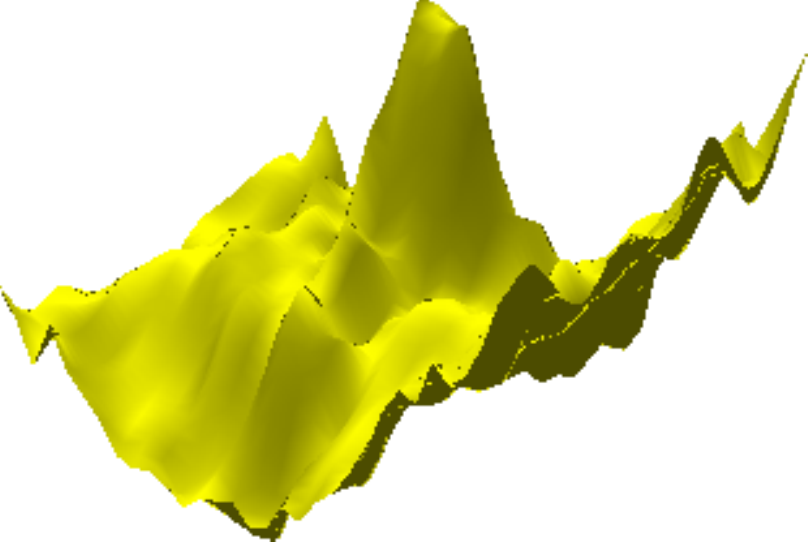}}
{\includegraphics[width=0.135\textwidth]{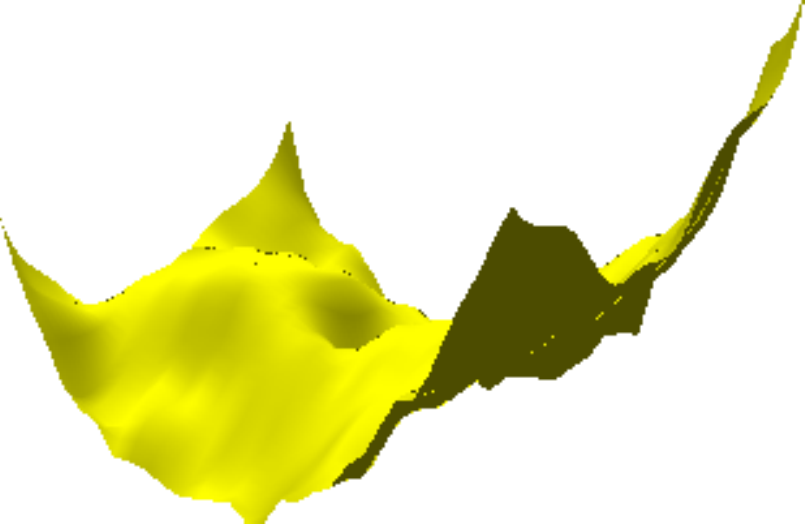}}
{\includegraphics[width=0.135\textwidth]{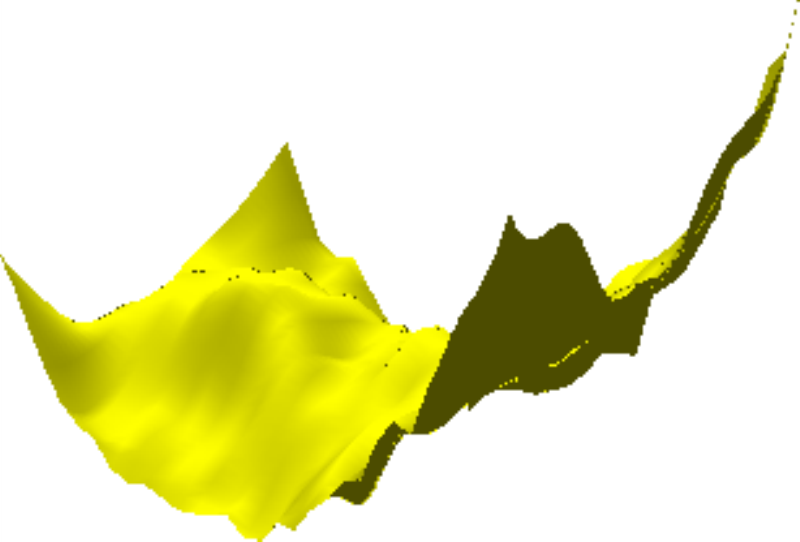}}
{\includegraphics[width=0.135\textwidth]{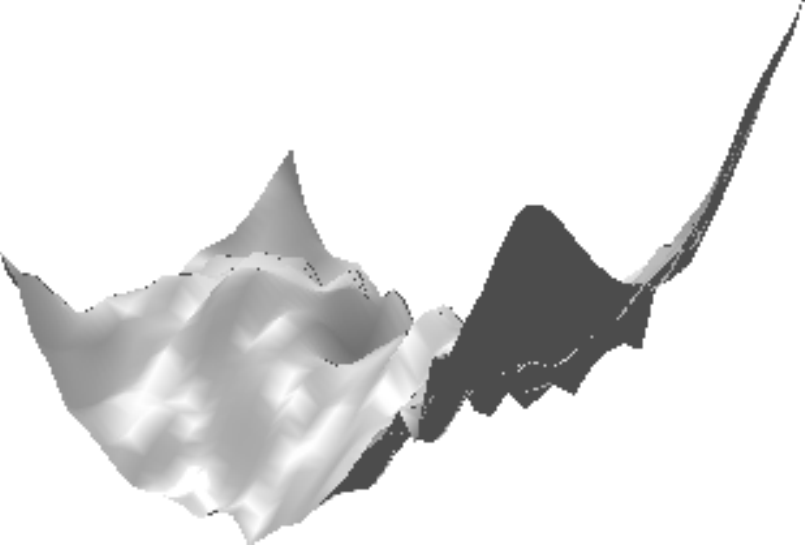}}
{\includegraphics[width=0.135\textwidth]{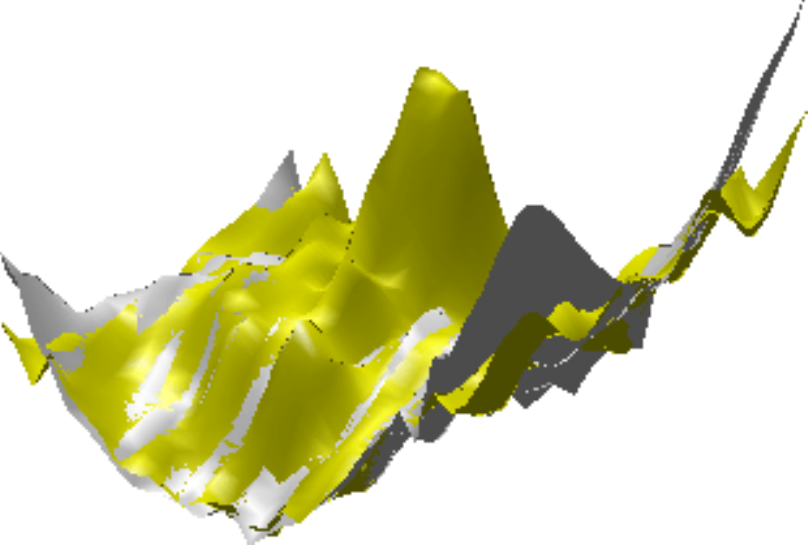}}
{\includegraphics[width=0.135\textwidth]{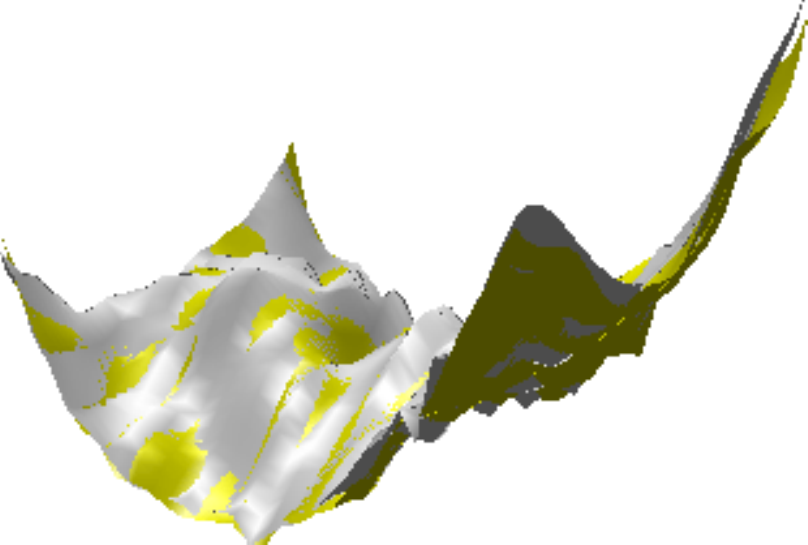}}
{\includegraphics[width=0.135\textwidth]{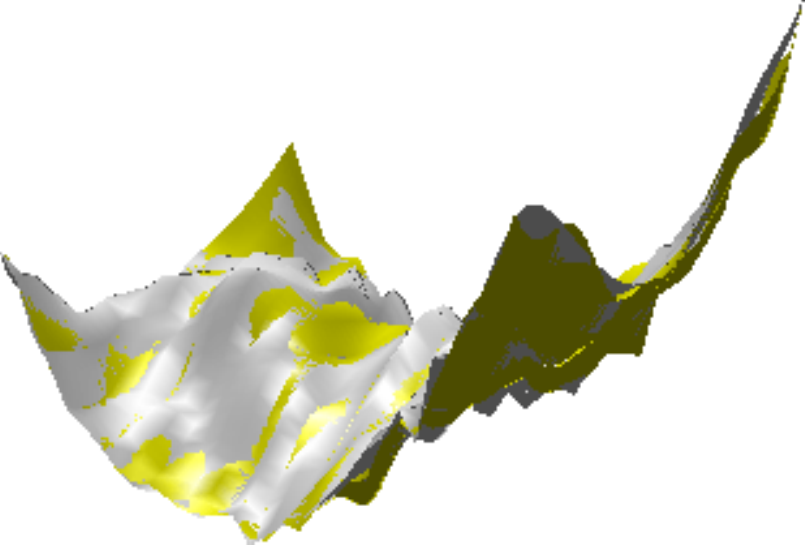}}
\\
{\includegraphics[width=0.135\textwidth]{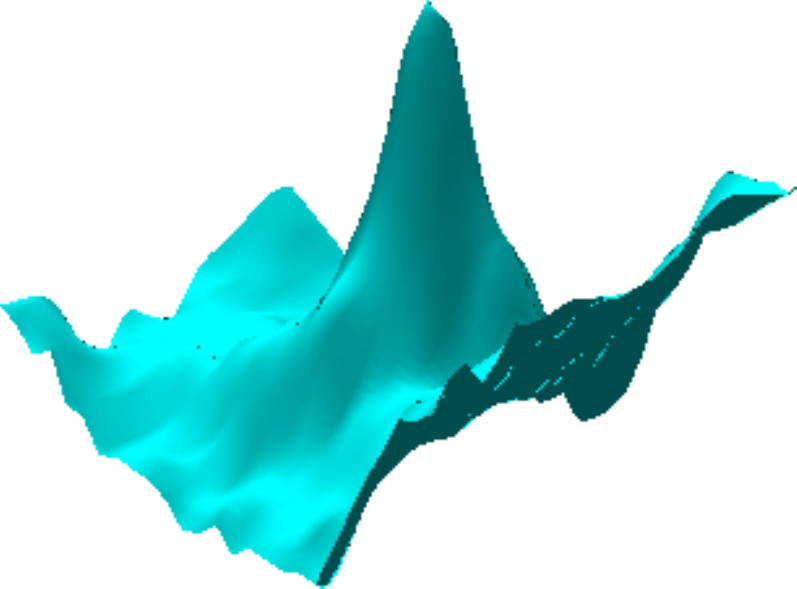}}
{\includegraphics[width=0.135\textwidth]{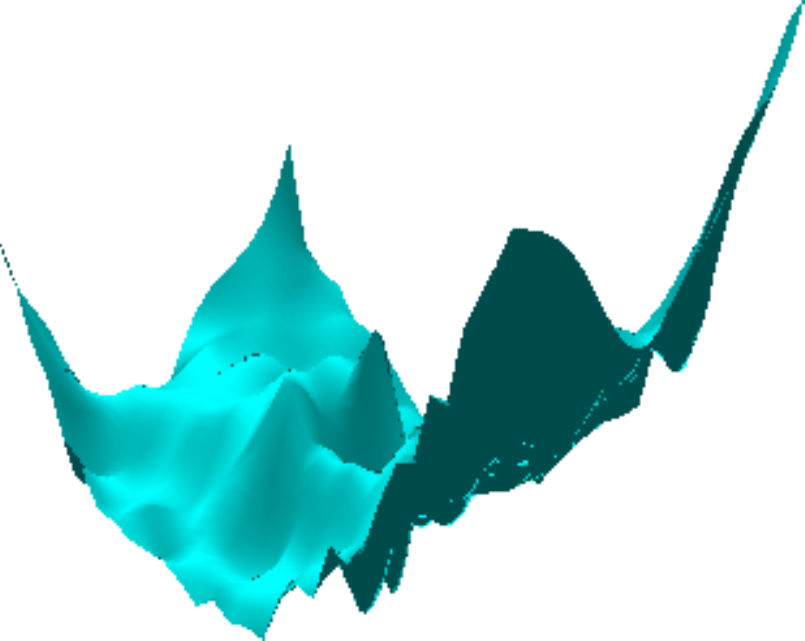}}
{\includegraphics[width=0.135\textwidth]{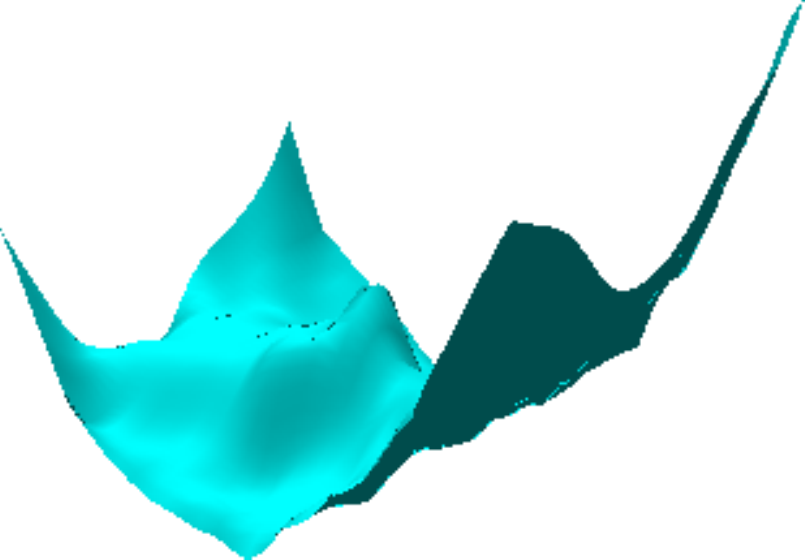}}
{\includegraphics[width=0.135\textwidth]{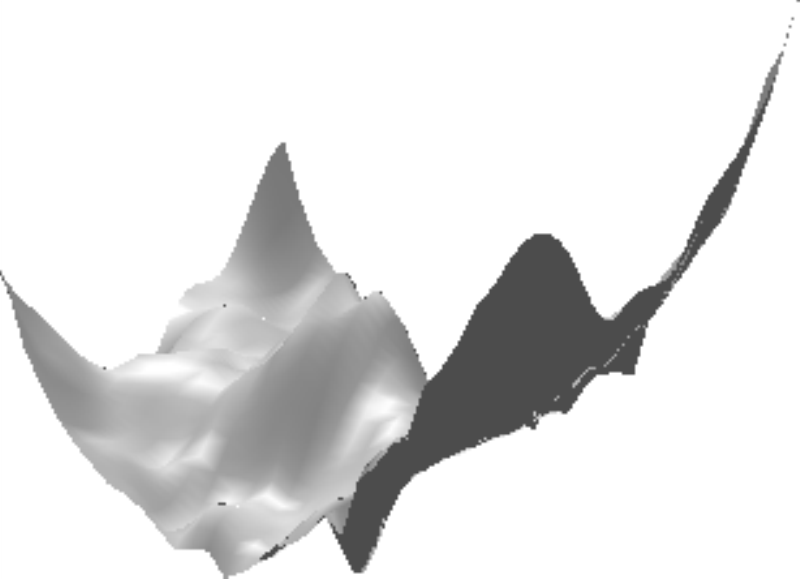}}
{\includegraphics[width=0.135\textwidth]{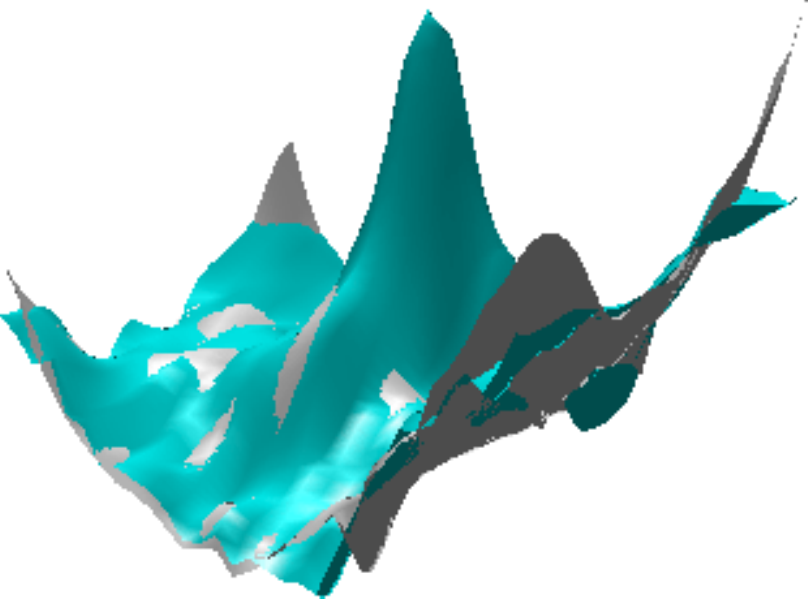}}
{\includegraphics[width=0.135\textwidth]{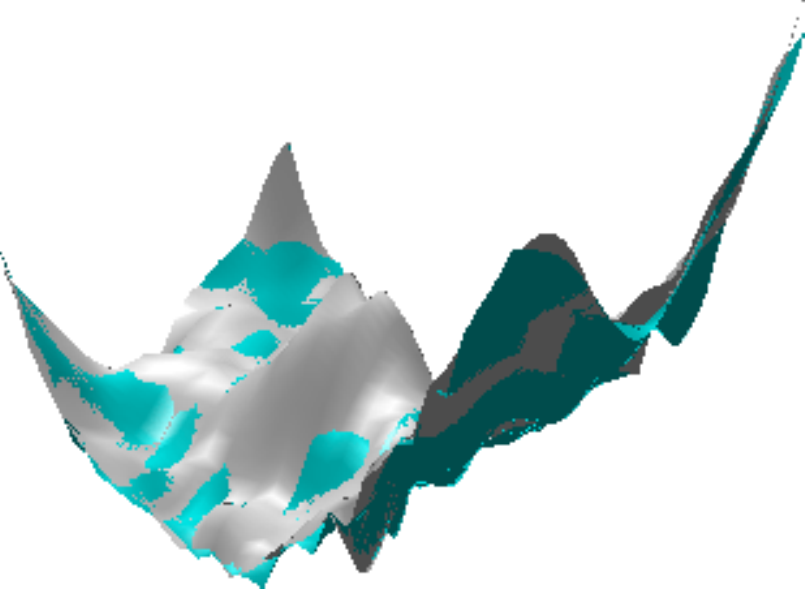}}
{\includegraphics[width=0.135\textwidth]{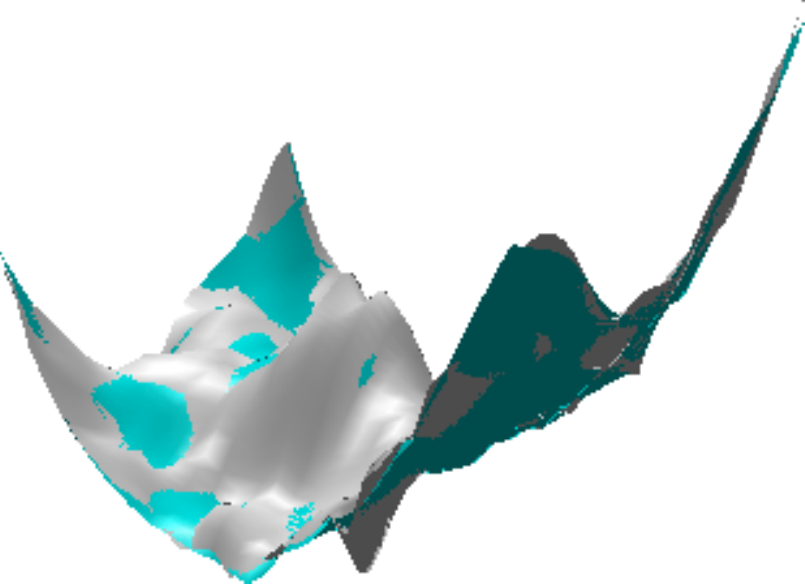}}
\\
{\includegraphics[width=0.135\textwidth]{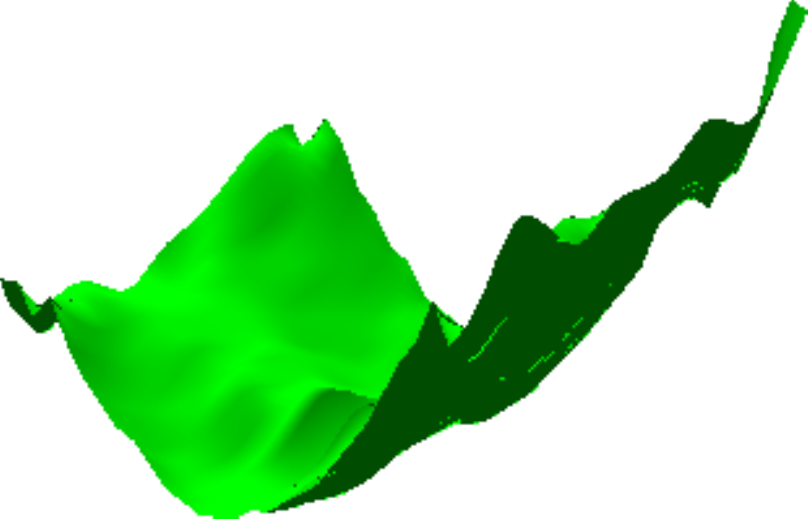}}
{\includegraphics[width=0.135\textwidth]{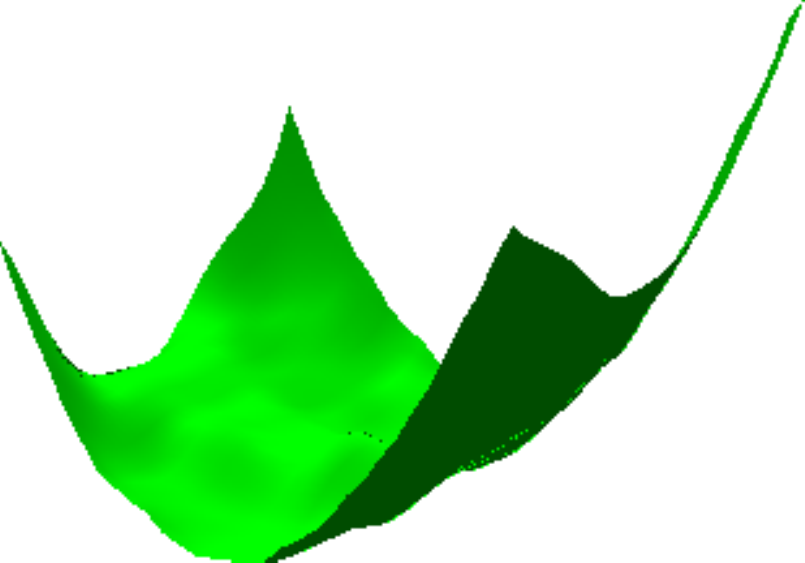}}
{\includegraphics[width=0.135\textwidth]{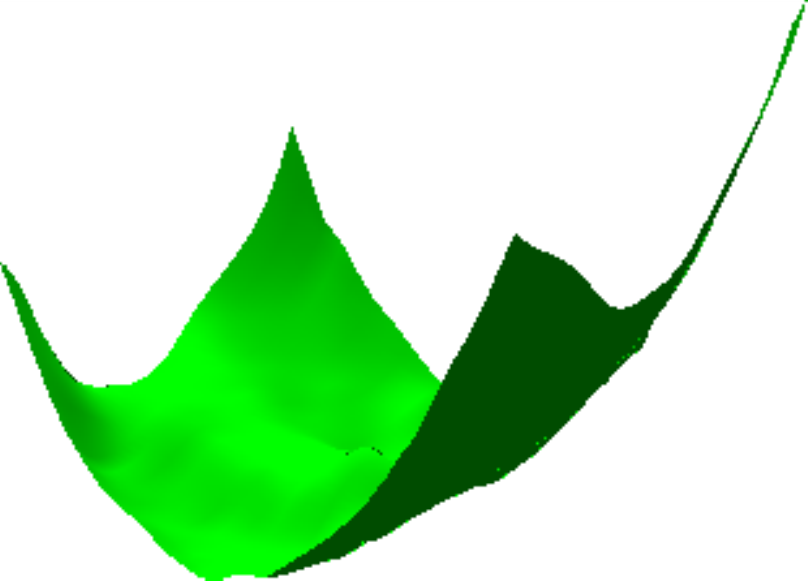}}
{\includegraphics[width=0.135\textwidth]{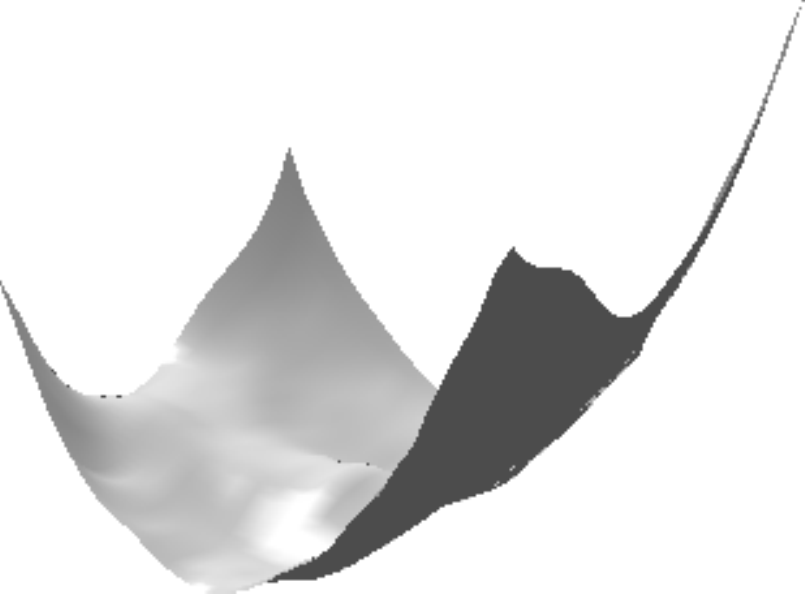}}
{\includegraphics[width=0.135\textwidth]{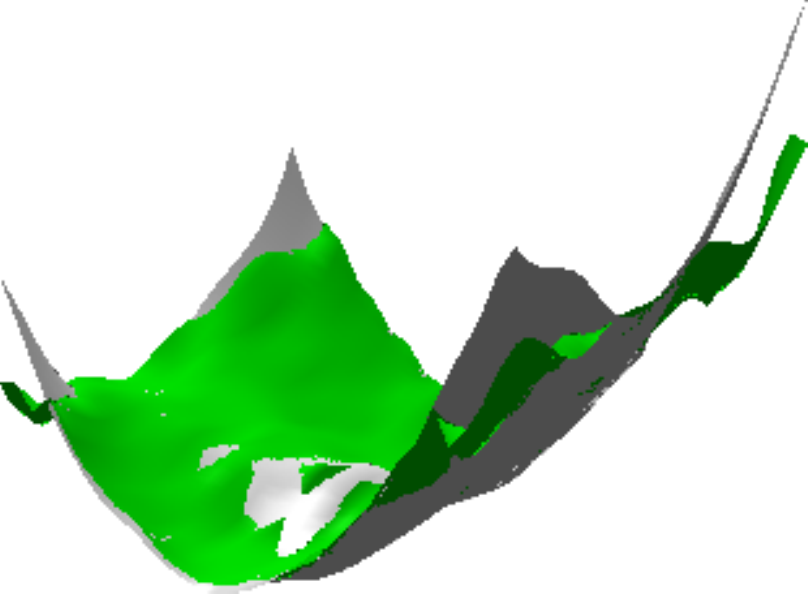}}
{\includegraphics[width=0.135\textwidth]{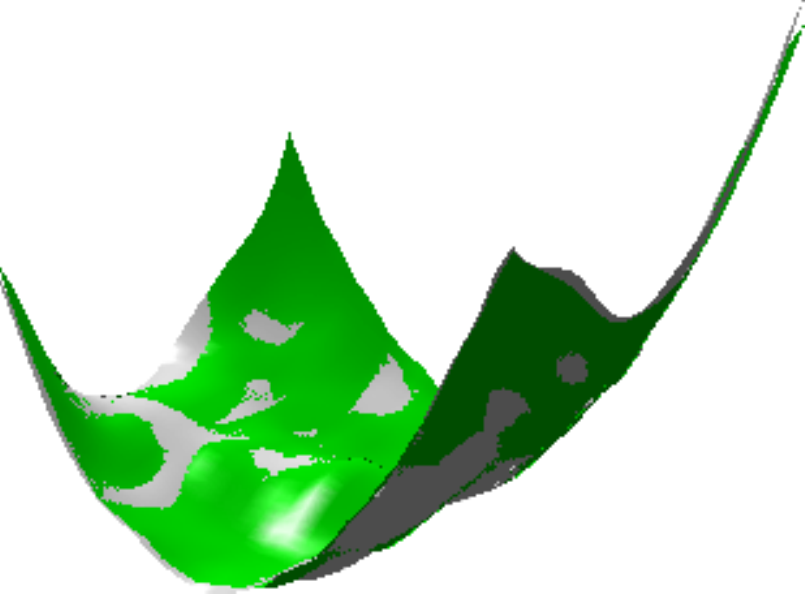}}
{\includegraphics[width=0.135\textwidth]{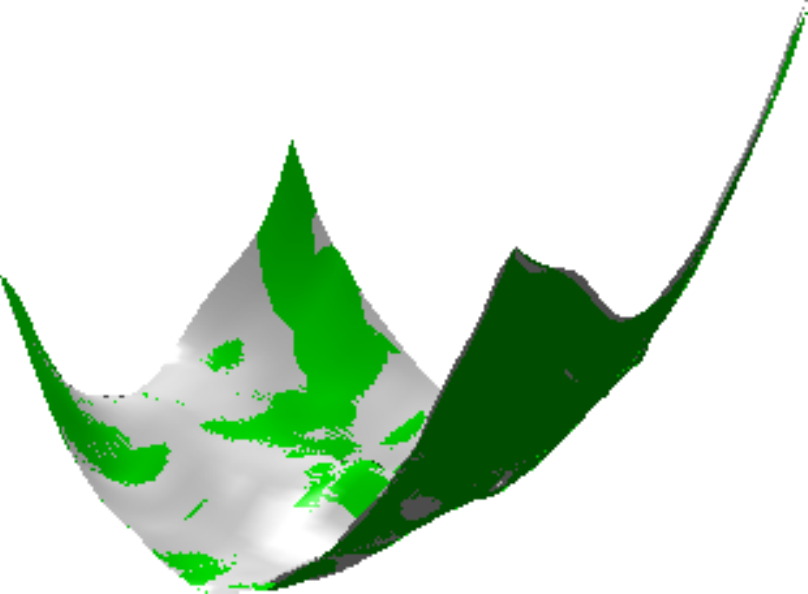}}
\\
\vspace{10pt}
\subfigure[]{\includegraphics[width=0.135\textwidth]{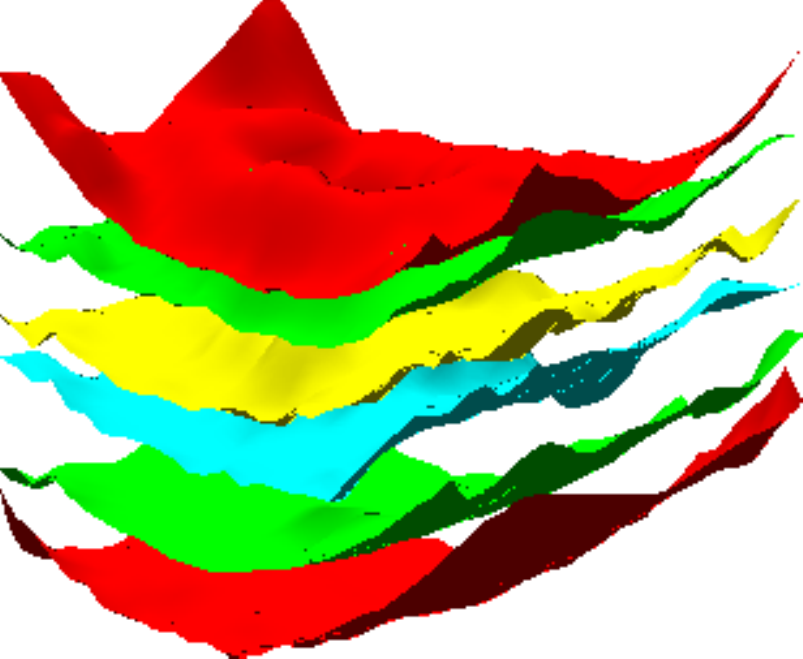}}
\subfigure[]{\includegraphics[width=0.135\textwidth]{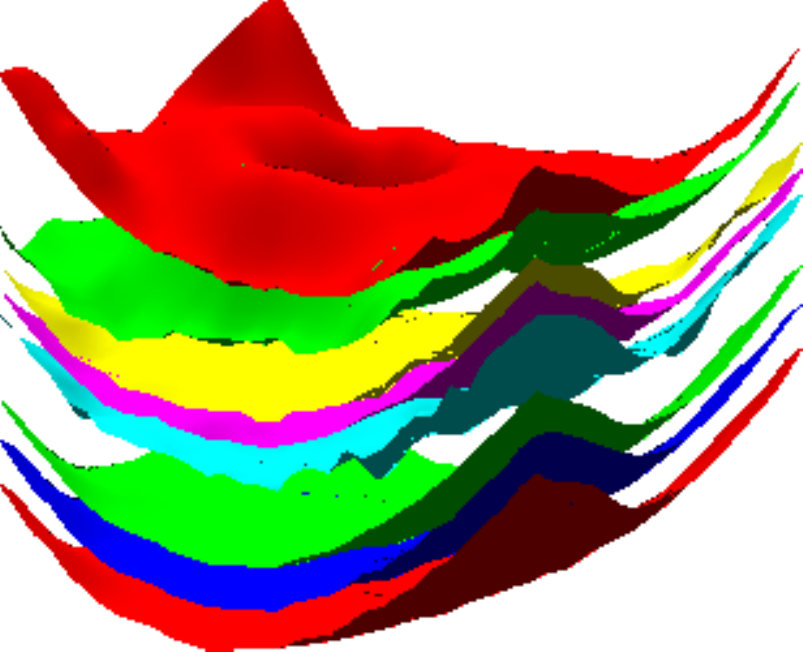}}
\subfigure[]{\includegraphics[width=0.135\textwidth]{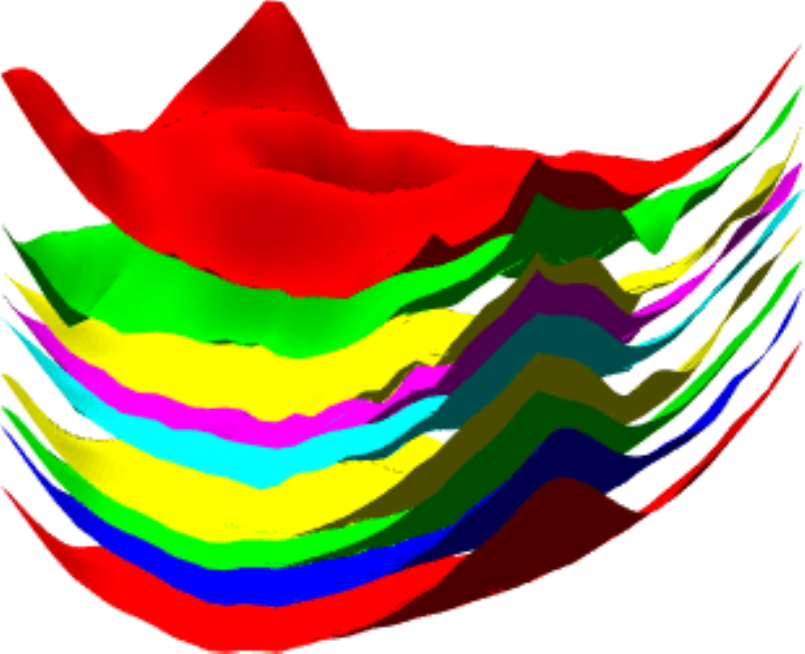}}
\subfigure[]{\includegraphics[width=0.135\textwidth]{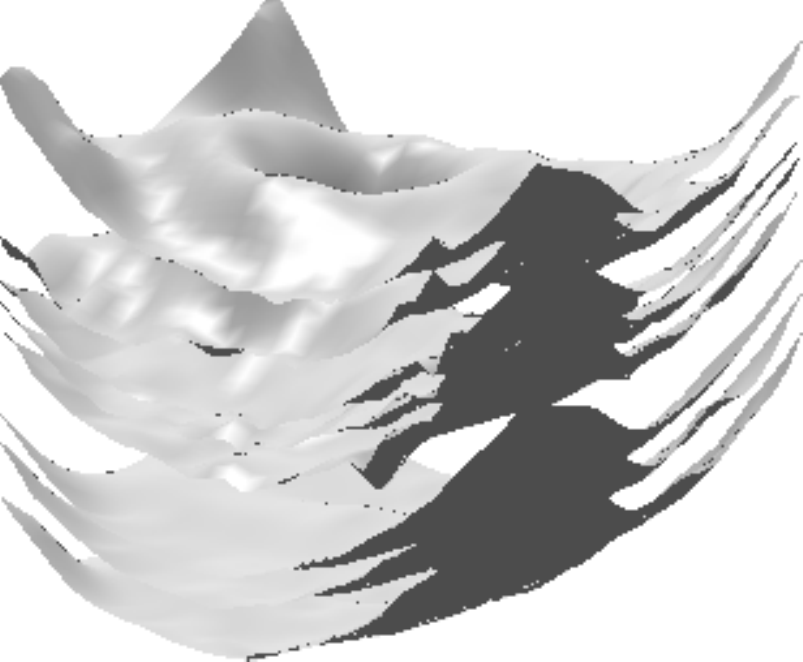}}
\subfigure[]{\includegraphics[width=0.135\textwidth]{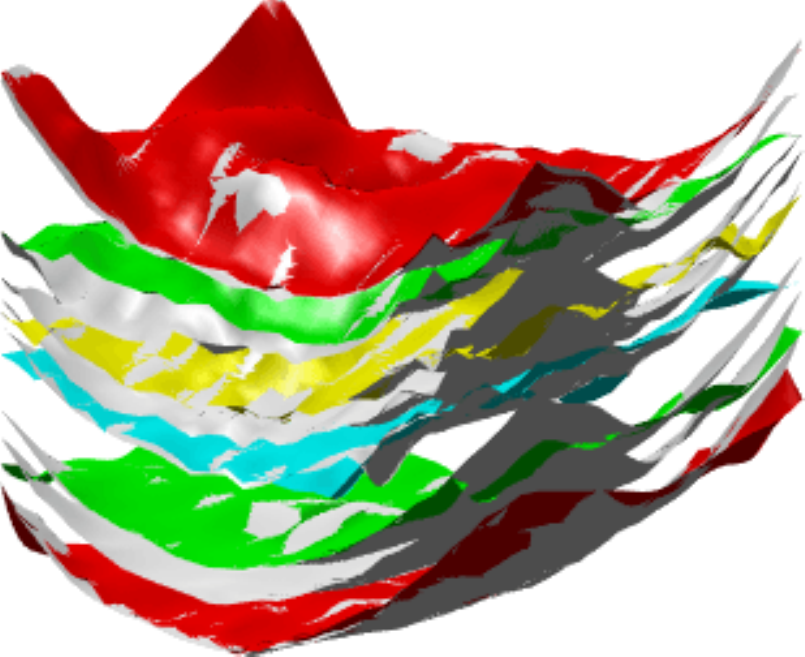}}
\subfigure[]{\includegraphics[width=0.135\textwidth]{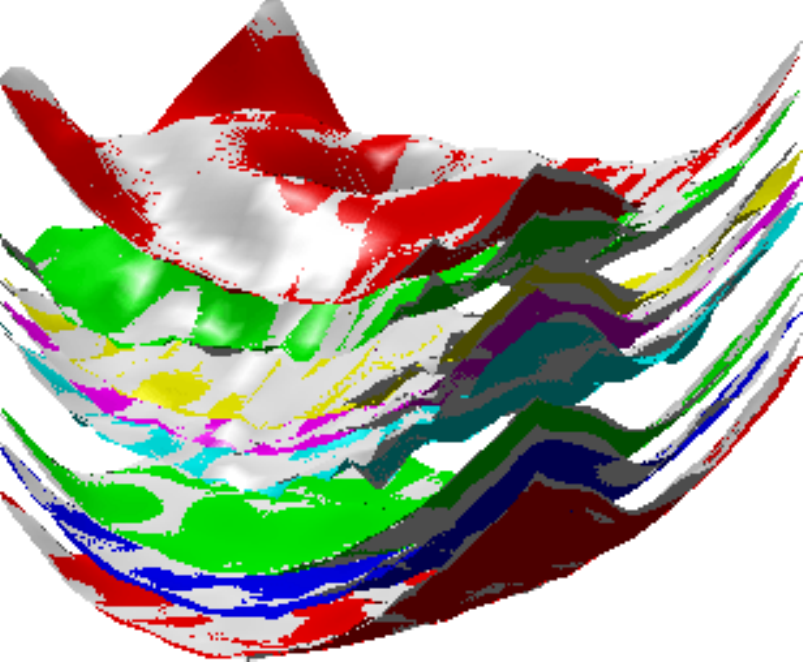}}
\subfigure[]{\includegraphics[width=0.135\textwidth]{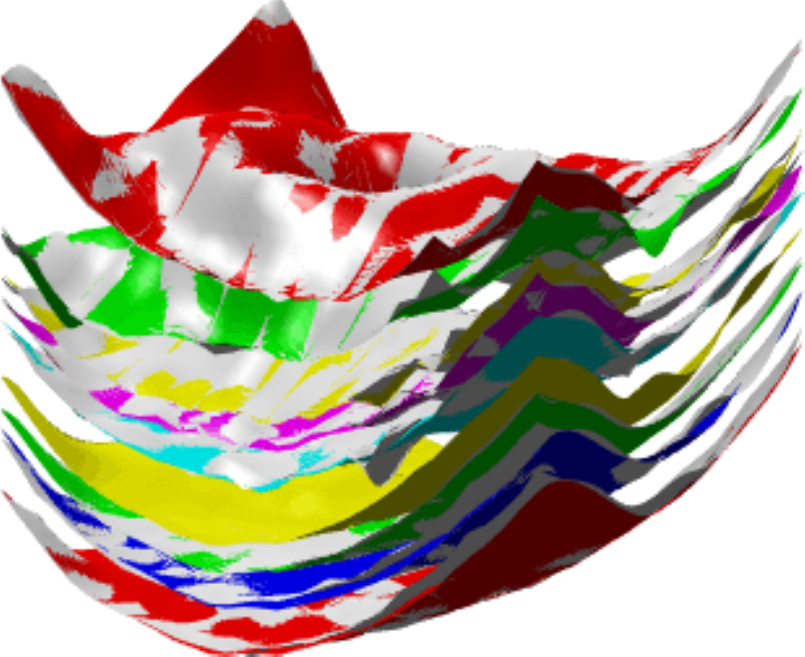}}
\\
\vspace{-5pt}
\caption{The 3D comparison between Dufour's method, OCTRIMA3D and GDM by segmenting the intra-retinal layer surfaces from the Volumes 4 sample. Column (a)-(d) are respectively Dufour's results, OCTRIMA3D results, GDM results and ground truth. Column (e)-(g) are respectively the segmentation results of the three compared methods, overlaid with ground truth. Row 1-6 are the results for the individual surface $B_1$, $B_2$, $B_3$, $B_5$, $B_7$ and total retina surfaces, respectively.}
\label{fig:compare3Dresults}
\end{figure}

The segmentation results of the three approaches on an exemplary sample (Volume 4) are shown in two distinctive B-scans in Figure~\ref{fig:enfaceBscans} and  \ref{fig:2SlicesSeg}, where one B-scan retinal structures are quite flat and the other contains the nonflat fovea region. Dufour's method has lower accuracy than the OCTIMA3D and GDM for both cases. OCTRIMA3D extends Chiu's method to 3D space and improves it by reducing the curvature in the fovea region using the inter-frame flattening technique, so the method performs very well for both flat and nonflat retinal structures. However, there still exist some obvious errors on the 5th boundary $B_5$. OCTRIMA3D is able to flatten the $B_1$ and in the meanwhile it also increases the curvature of its adjacent boundaries such as $B_5$, which might be the reason leading to the errors. Compared with the other two, the GDM's results show less green lines, verifying that the results are closer to ground truth and thus it is the most accurate among the three compared. In addition to the 2D visualisation, the 3D rendering of the results segmented by the three approaches is given in Figure~\ref{fig:compare3Dresults}. The experiment furthermore shows that Dufour's results deviate much from ground truth, while the OCTRIMA3D is better than Dufour's method and comparable to the GDM. The GDM results cover less grey ground truth and are thereby the best.

\newcolumntype{C}{>{\centering\arraybackslash}p{6.2em}}
\newcolumntype{g}{>{\columncolor[RGB]{195, 195, 195}}C}
\newcolumntype{k}{>{\columncolor{Gray}}C}
\begin{table*}[htbp] 
\caption{The SE ($\mu m$), AE ($\mu m$) and HD ($\mu m$) calculated using the results of different methods (Dufour's method, OCTRMIA3D and GDM) and the ground truth manual segmentation, for the OPL-ONL ($B_5$) surface in each of the 10 OCT volumes.}
\vspace{-10pt}
\centering \setlength{\tabcolsep}{1pt}
\resizebox{\columnwidth}{!}{
\begin{tabular}{cgkCgkCgkC} 
\toprule
&\multicolumn{3}{c}{SE ($\mu m$)}  &\multicolumn{3}{c}{AE ($\mu m$)}  &\multicolumn{3}{c}{HD ($\mu m$)}\\

Volume \#         &Dufour et al.      &OCTRIMA3D          &GDM 
                  &Dufour et al.      &OCTRIMA3D          &GDM 
                  &Dufour et al.      &OCTRIMA3D          &GDM  \\
\midrule
1                 &-1.194     &0.4559     &0.3782    
				  &2.3816     &1.3490     &1.0720	 
				  &25.688     &15.273	  &10.449    \\
				
2                 &-2.170     &-0.036     &-0.128     
				  &4.5250     &0.9089     &0.7814	 
				  &56.667     &11.570	  &7.0938    \\ 

3                 &-2.576     &0.4182     &0.5983     
				  &3.6129     &1.3237     &1.0989	 
				  &25.203     &16.719     &9.5326       \\

4                 &-2.296     &1.0987     &0.6774  
				  &3.8185     &1.5175     &1.0753	 
				  &51.522     &18.364	  &9.6151      \\
				  
5                 &-1.680     &1.3288     &0.5909     
				  &4.3327     &1.5012     &0.9005	 
				  &56.223     &11.889     &8.8419      \\
			
6                 &-2.623     &1.0732     &0.2974     
				  &4.0682     &1.4838     &0.9493	 
				  &43.070     &19.201	  &9.5281       \\
		
7                 &-2.326     &0.5294     &0.4529    
				  &3.1506     &0.9378     &0.7433	 
				  &31.782     &8.6701     &6.4803       \\

8                 &-0.636     &1.1355     &0.6833     
				  &2.3955     &1.4455     &1.0069	 
				  &25.481     &17.930     &11.685      \\
        
9                 &-4.206     &0.3077     &0.0859     
				  &4.5813     &1.0780     &0.7678	 
				  &43.223     &8.9694	  &5.7191    \\

10                &-2.648     &0.6701     &0.2606     
				  &4.4903     &1.0627     &0.7877	 
				  &41.017     &11.666	  &10.961    \\                  
\bottomrule
\end{tabular}}
\label{tb:volumeB5}
\end{table*}

\newcolumntype{C}{>{\centering\arraybackslash}p{6.2em}}
\newcolumntype{g}{>{\columncolor[RGB]{195, 195, 195}}C}
\newcolumntype{k}{>{\columncolor{Gray}}C}
\begin{table*}[htbp] 
\caption{The SE ($\mu m$), AE ($\mu m$) and HD ($\mu m$) calculated using the results of different methods (Dufour's method, OCTRMIA3D and GDM) and the ground truth manual segmentation, for the IS-OS ($B_7$) surface in each of the 10 OCT volumes}
\vspace{-10pt}
\centering \setlength{\tabcolsep}{1pt}
\resizebox{\columnwidth}{!}{
\begin{tabular}{cgkCgkCgkC} 
\toprule
&\multicolumn{3}{c}{SE ($\mu m$)}  &\multicolumn{3}{c}{AE ($\mu m$)}  &\multicolumn{3}{c}{HD ($\mu m$)}\\

Volume \#         &Dufour et al.      &OCTRIMA3D          &GDM 
                  &Dufour et al.      &OCTRIMA3D          &GDM 
                  &Dufour et al.      &OCTRIMA3D          &GDM  \\
\midrule
1                 &-0.432     &-0.148      &-0.019    
				  &1.1013     &0.5391      &0.4437	 
				  &16.559     &4.7616	   &4.5805     \\

2                 &0.7476     &-0.276      &-0.079     
				  &2.0329     &0.5539      &0.3971	 
				  &20.309     &5.2093	   &3.7743 \\
		
3                 &-0.311     &-0.291      &-0.106     
				  &1.4347     &0.5406      &0.4629	 
				  &18.432     &2.9790      &4.0176       \\

4                 &0.3652     &-0.116      &0.3363     
				  &1.6954     &0.5271      &0.4601	 
				  &27.853     &5.3672	   &2.7882      \\

5                 &0.6057     &-0.098      &0.0994     
				  &1.7567     &0.4756      &0.3500	 
				  &26.556     &3.7573      &3.4150       \\


6                 &0.9825     &-0.592      &-0.139     
				  &2.4970     &0.7247      &0.4066	 
				  &23.487     &5.9301	   &3.9297       \\
				  
7                 &-1.247     &-0.536      &0.0237     
				  &1.3895     &0.7501      &0.3716	 
				  &10.016     &3.1398      &3.6980      \\

8                 &-0.311     &-0.069      &0.1740     
				  &1.0438     &0.4053      &0.3466	 
				  &15.044     &4.2301      &4.3940       \\

9                 &-0.755     &-0.111     &0.1407     
				  &0.8068     &0.5422     &0.3939	 
				  &3.5210     &3.4263	  &3.3868    \\

10                &-0.099     &-0.220     &0.1028    
				  &1.2941     &0.5609     &0.4246	 
				  &13.313     &3.1210	  &3.5361    \\        
\bottomrule
\end{tabular}
}
\label{tb:volumeB7}
\end{table*}

\newcolumntype{C}{>{\centering\arraybackslash}p{6.2em}}
\newcolumntype{g}{>{\columncolor[RGB]{195, 195, 195}}C}
\newcolumntype{k}{>{\columncolor{Gray}}C}
\begin{table*}[htbp] 
\caption{The OSE ($\mu m$), OAE ($\mu m$) and OHD ($\mu m$) calculated using the results of different methods (Dufour's method, OCTRMIA3D and GDM) and the ground truth manual segmentation, for overall retina surfaces in each of the 10 OCT volumes}
\vspace{-10pt}
\centering \setlength{\tabcolsep}{1pt}
\resizebox{\columnwidth}{!}{
\begin{tabular}{cgkCgkCgkC} 
\toprule
&\multicolumn{3}{c}{OSE ($\mu m$)}  &\multicolumn{3}{c}{OAE ($\mu m$)}  &\multicolumn{3}{c}{OHD ($\mu m$)}\\

Volume \#         &Dufour et al.      &OCTRIMA3D          &GDM 
                  &Dufour et al.      &OCTRIMA3D          &GDM 
                  &Dufour et al.      &OCTRIMA3D          &GDM  \\
\midrule
1                 &-1.271     &0.3607      &0.4338     
				  &1.8358     &1.1204      &0.9538
				  &17.486     &9.3358	   &7.9163  \\
				
2                 &-1.161     &0.0246      &0.0640     
				  &2.5380     &0.9652      &0.7238	 
				  &29.682     &7.7987	   &6.1267 \\

3                 &-1.513     &-0.052      &0.3456	     
				  &2.1470     &0.9343      &0.7838	 
				  &19.985     &8.3491      &6.9920       \\

4                 &-1.431     &0.4272      &0.3560     
				  &2.5278     &1.0374      &0.8667	 
				  &31.346     &9.4042	   &7.3130      \\
				  
5                 &-1.020     &0.6369      &0.5021    
				  &2.4119     &1.0794      &0.8289	 
				  &32.607     &8.6822      &7.1379       \\
			
6                 &-1.434     &0.4216     &0.3969     
				  &2.6754     &1.1371     &0.8606	 
				  &28.629     &9.5267	  &7.2548       \\
		
7                 &-2.010     &0.0059     &0.3283     
				  &2.2458     &0.9682     &0.7407	 
				  &21.788     &7.0644     &6.8279       \\

8                 &-1.031     &0.5815     &0.5785     
				  &1.7462     &1.1063     &0.9067	 
				  &17.610     &10.100     &8.5112     \\

9                 &-1.951     &0.0542     &0.2014    
				  &2.1368     &0.8771     &0.6922	 
				  &21.344     &5.7482	  &5.4794    \\

10                     &-1.513     &0.1022     &0.2109     
				       &2.3315     &0.8397     &0.6596	 
				       &24.841     &6.3250	   &6.7132    \\        
\bottomrule
\end{tabular}
}
\label{tb:volumeOveall}
\end{table*}

\begin{figure}[h!] 
\centering  
{\includegraphics[width=0.28\textwidth]{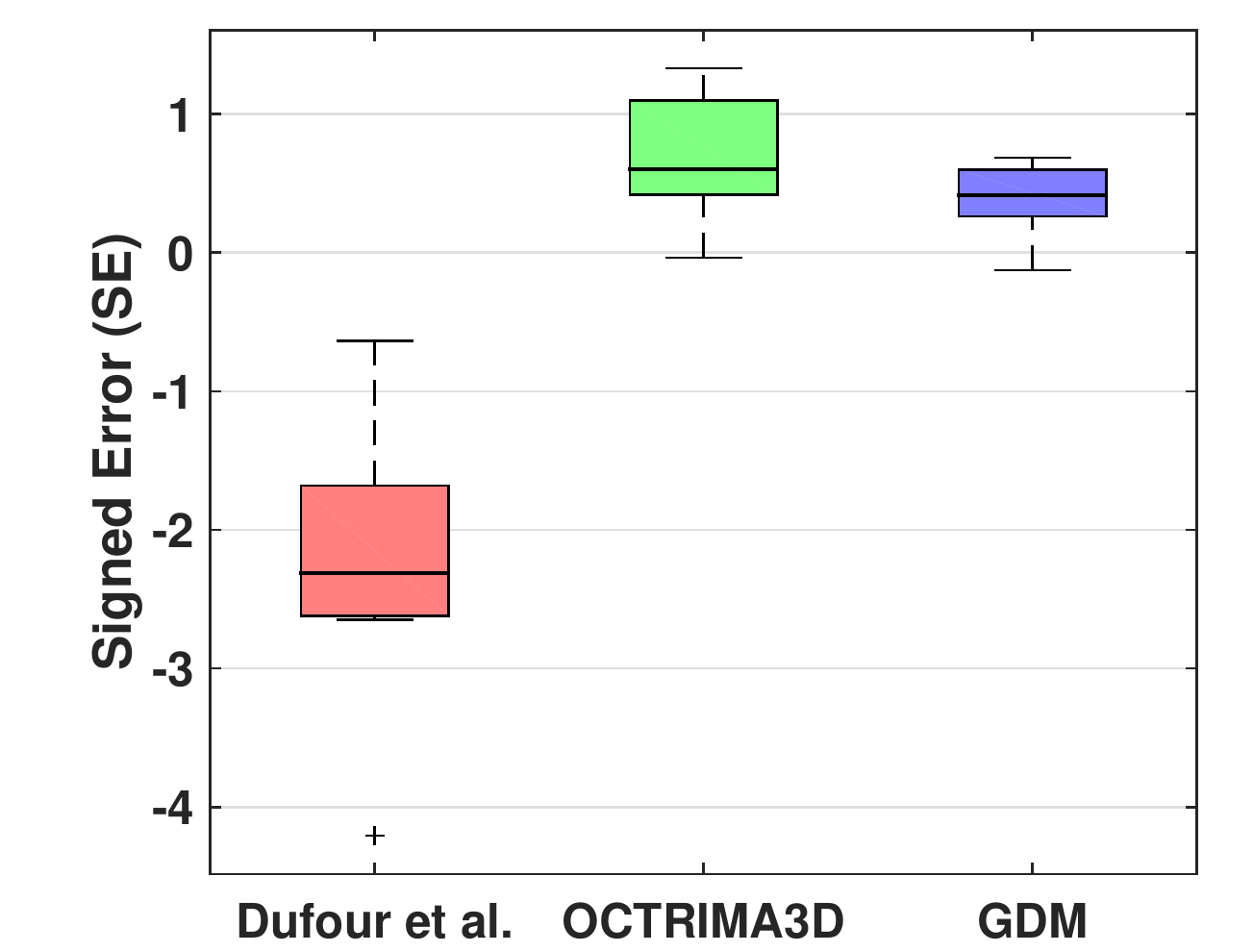}}
{\includegraphics[width=0.28\textwidth]{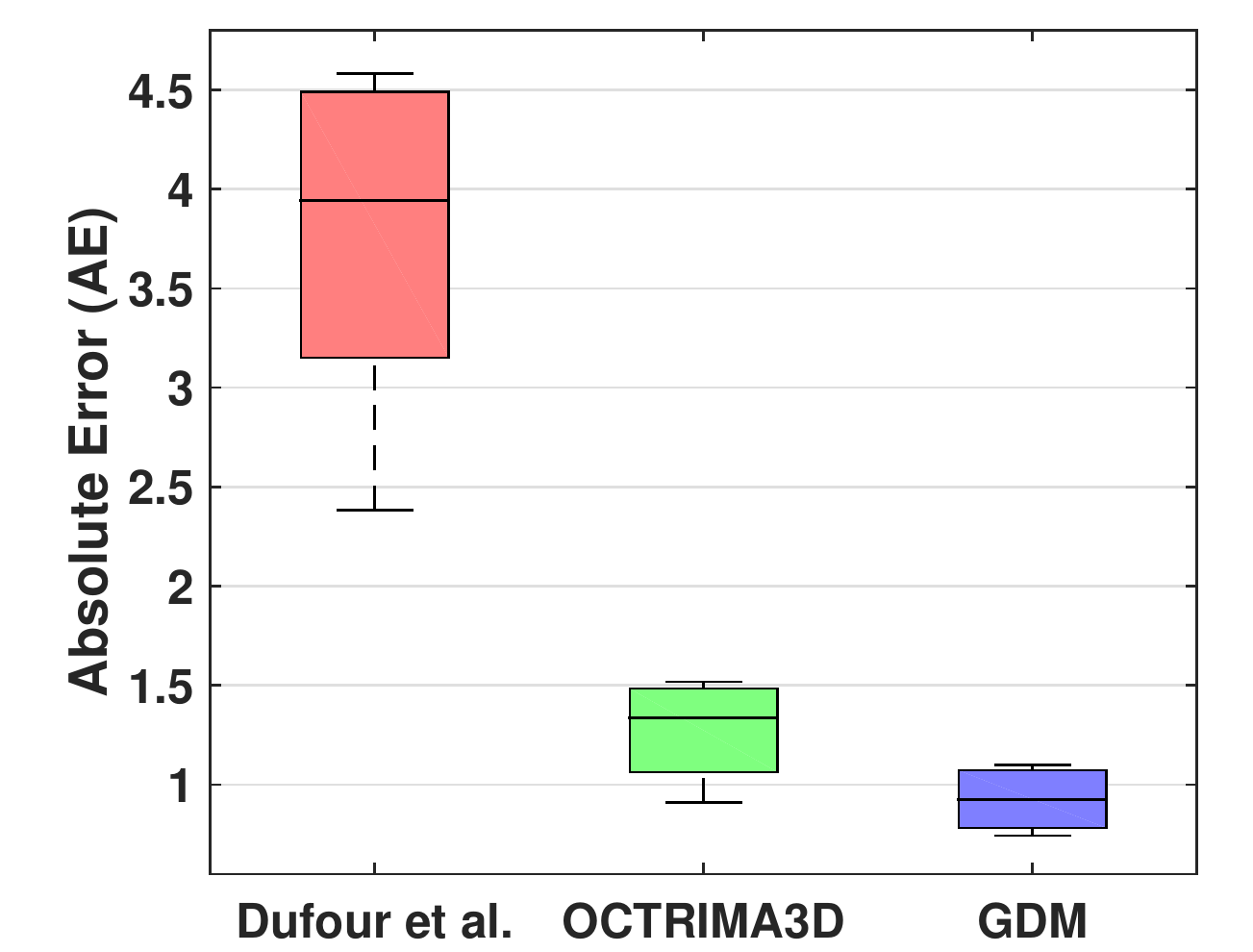}}
{\includegraphics[width=0.28\textwidth]{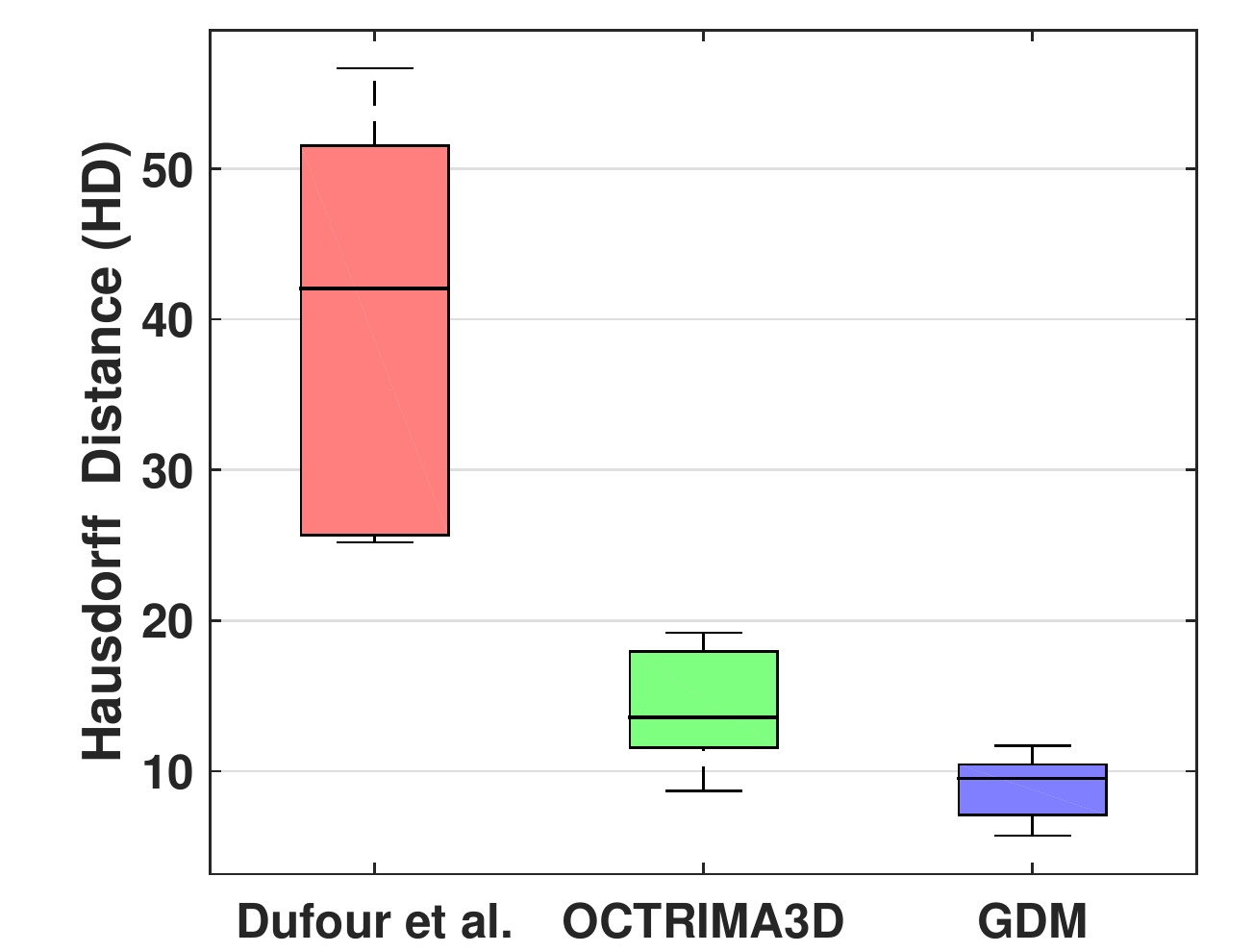}}\\
{\includegraphics[width=0.28\textwidth]{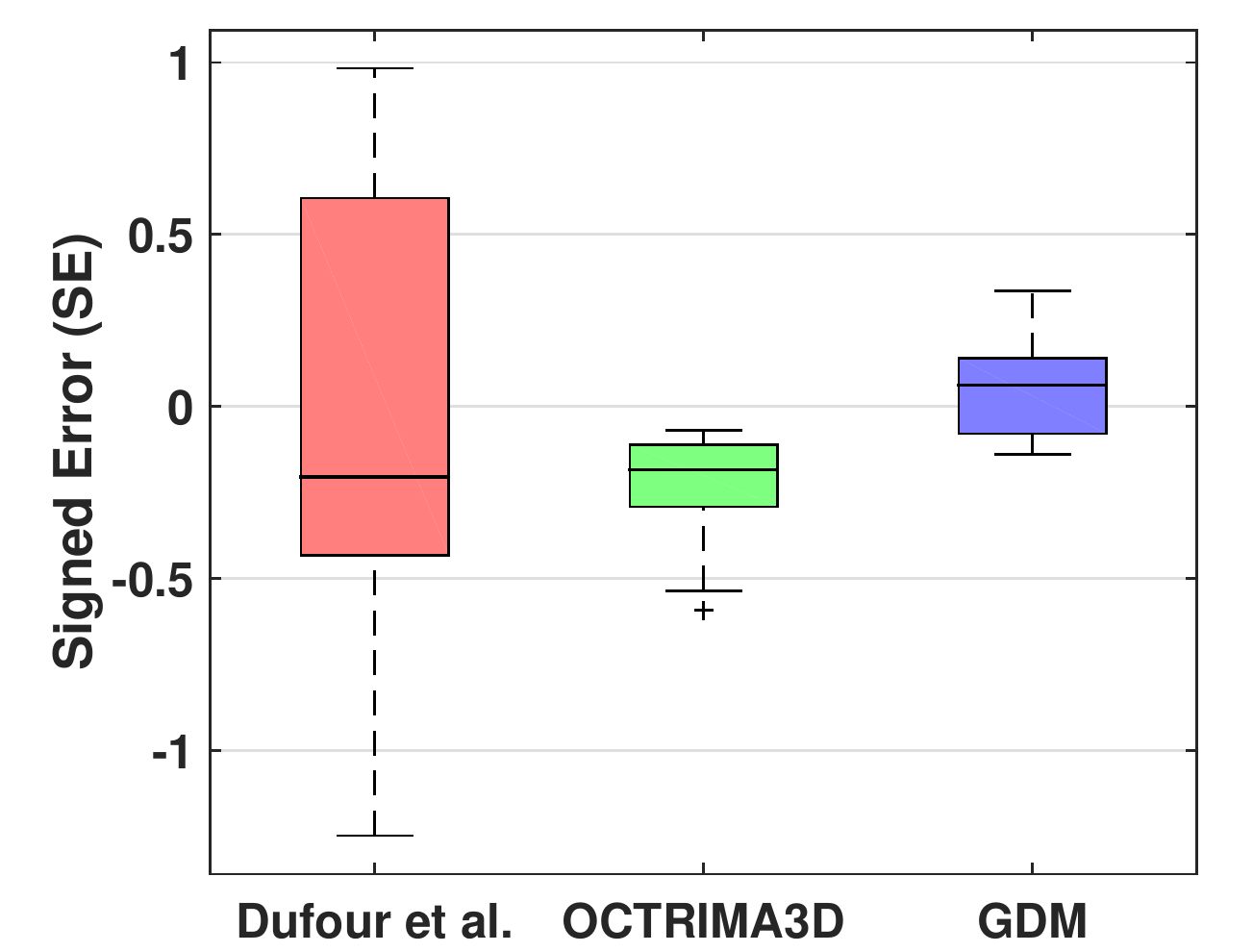}}
{\includegraphics[width=0.28\textwidth]{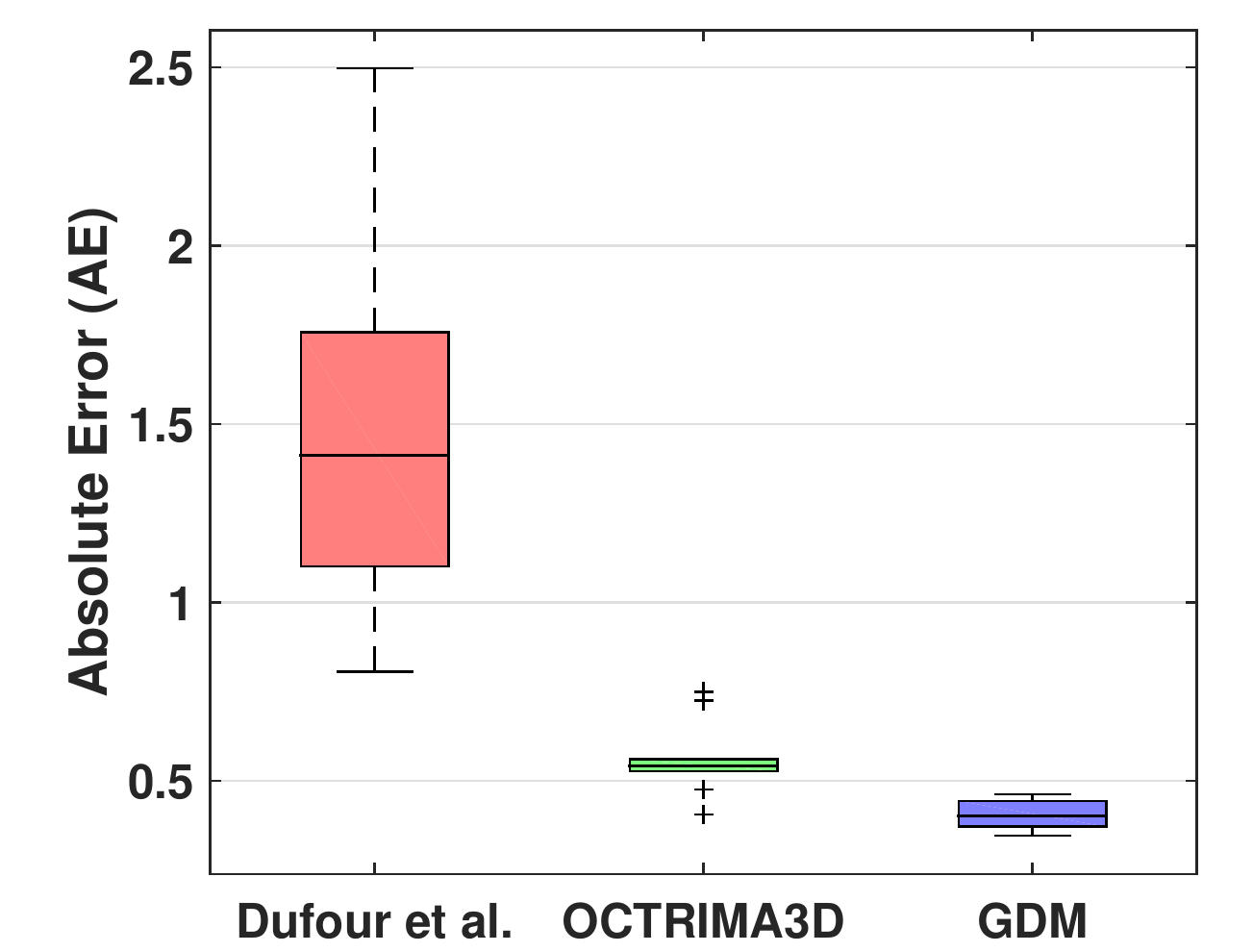}}
{\includegraphics[width=0.28\textwidth]{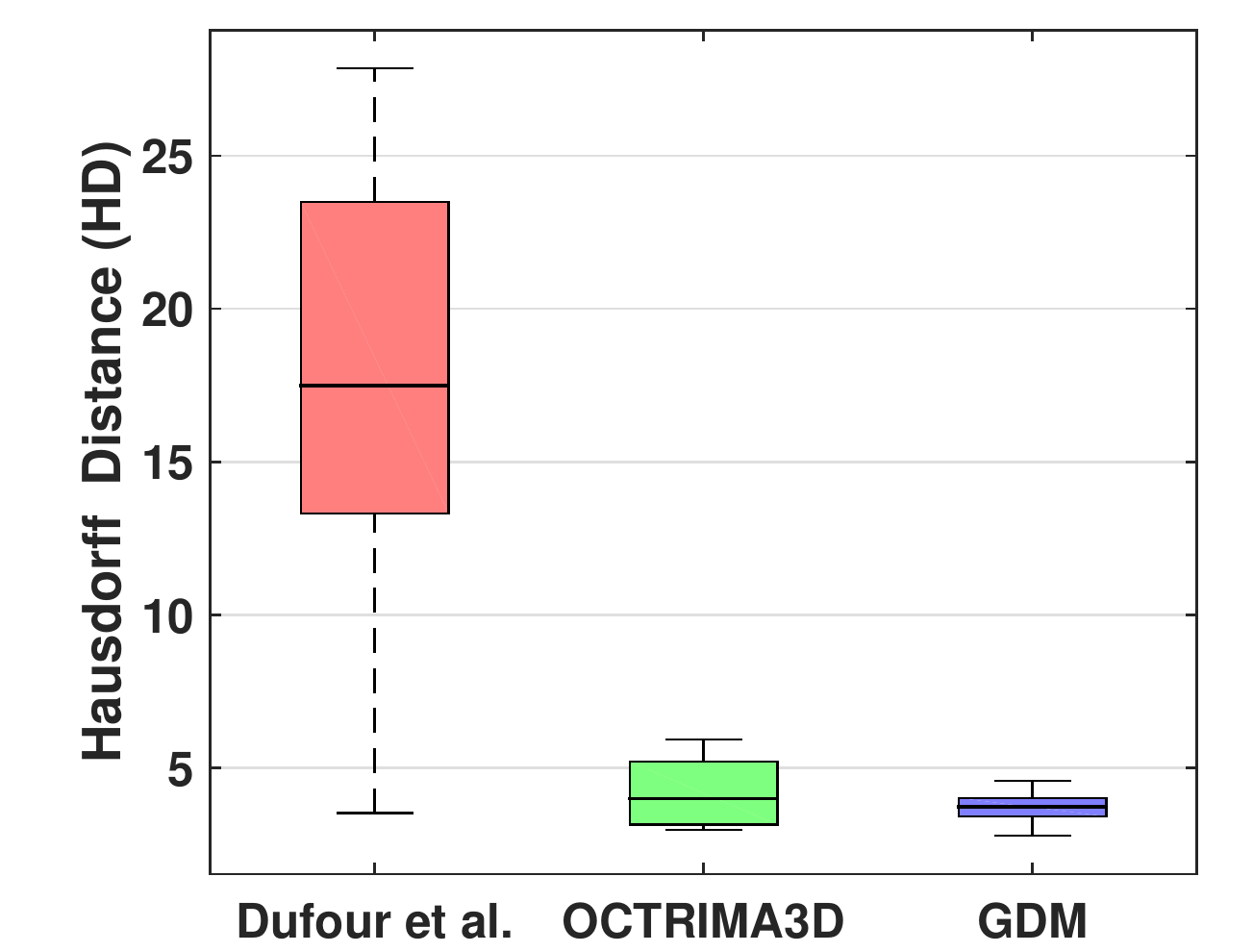}}\\
{\includegraphics[width=0.28\textwidth]{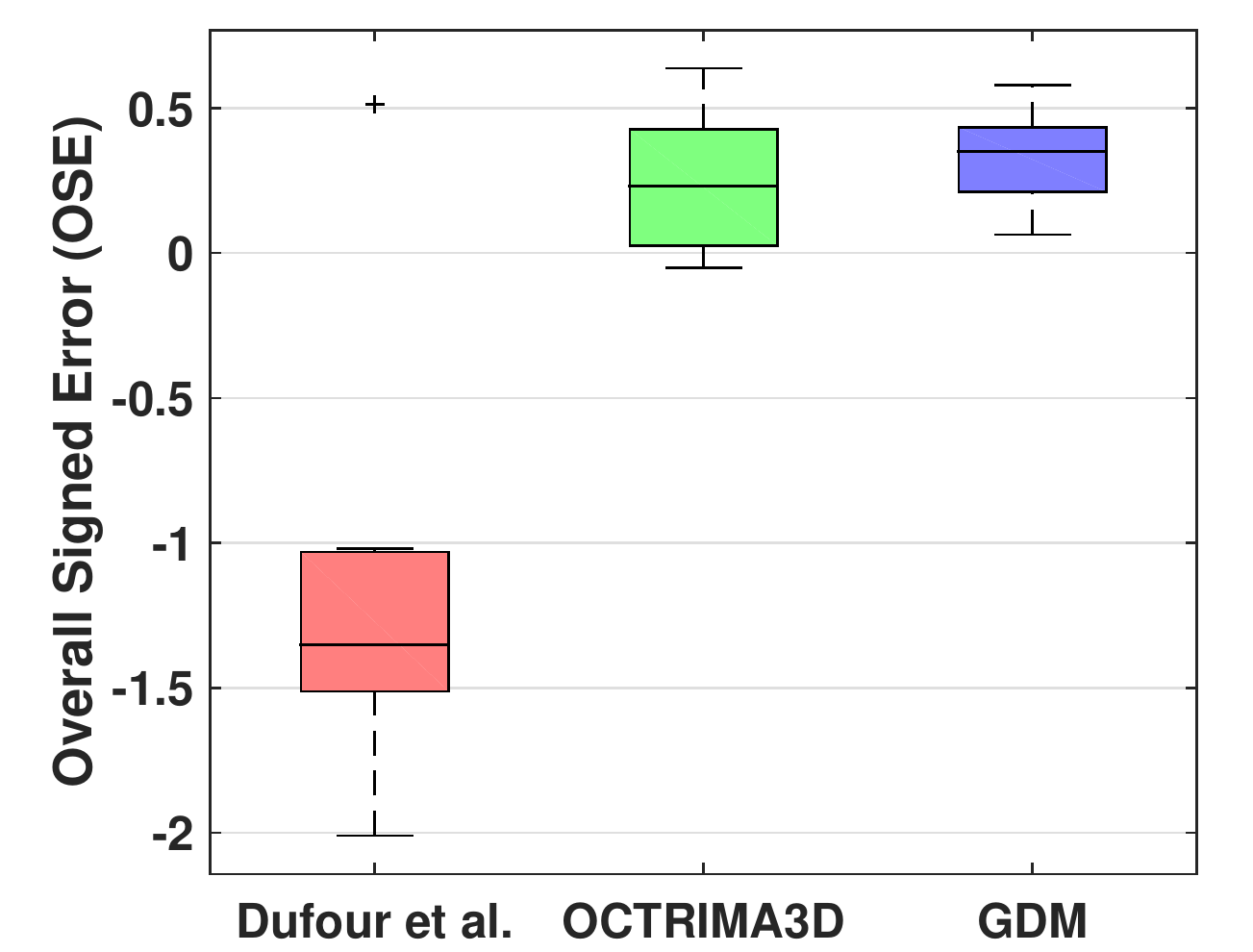}}
{\includegraphics[width=0.28\textwidth]{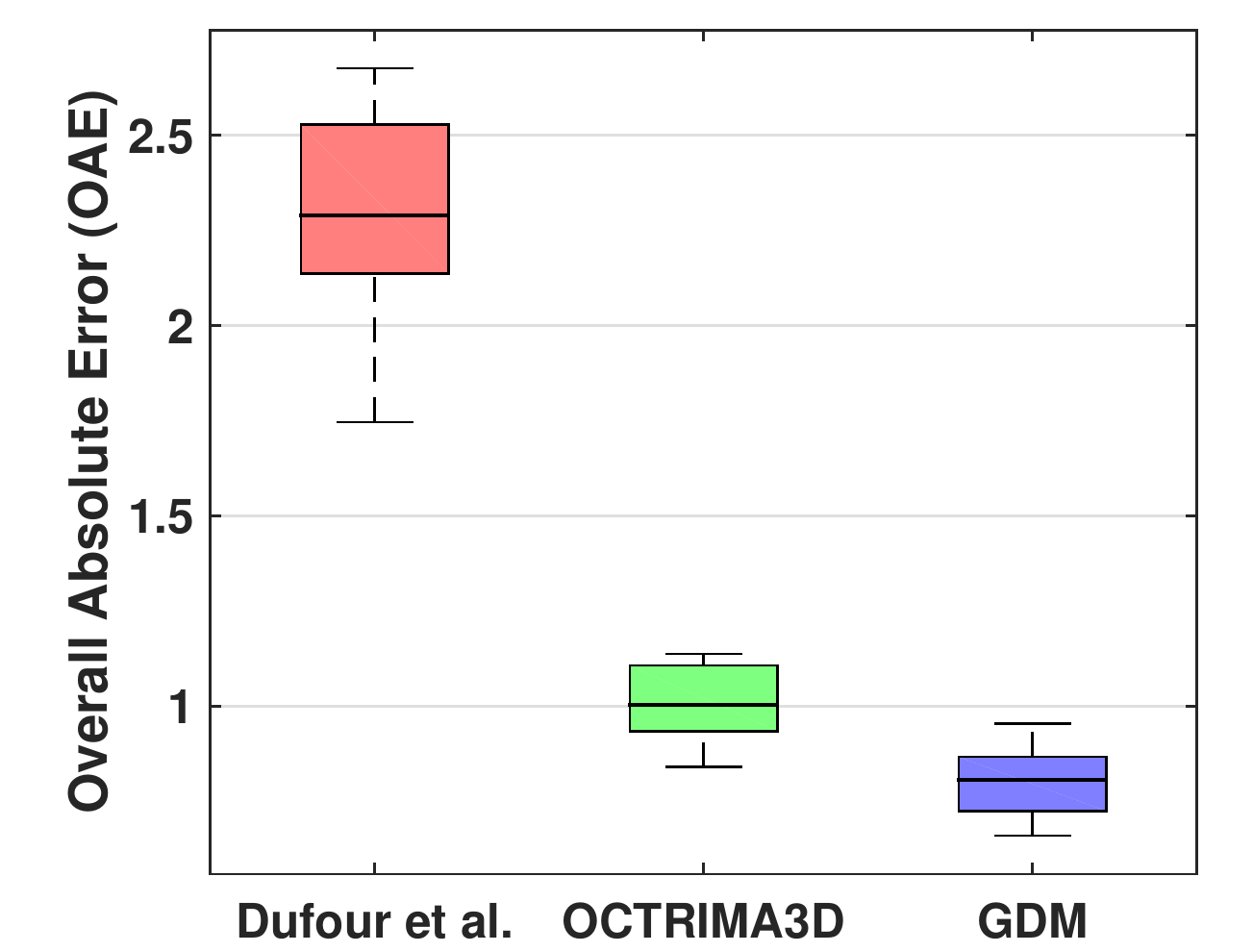}}
{\includegraphics[width=0.28\textwidth]{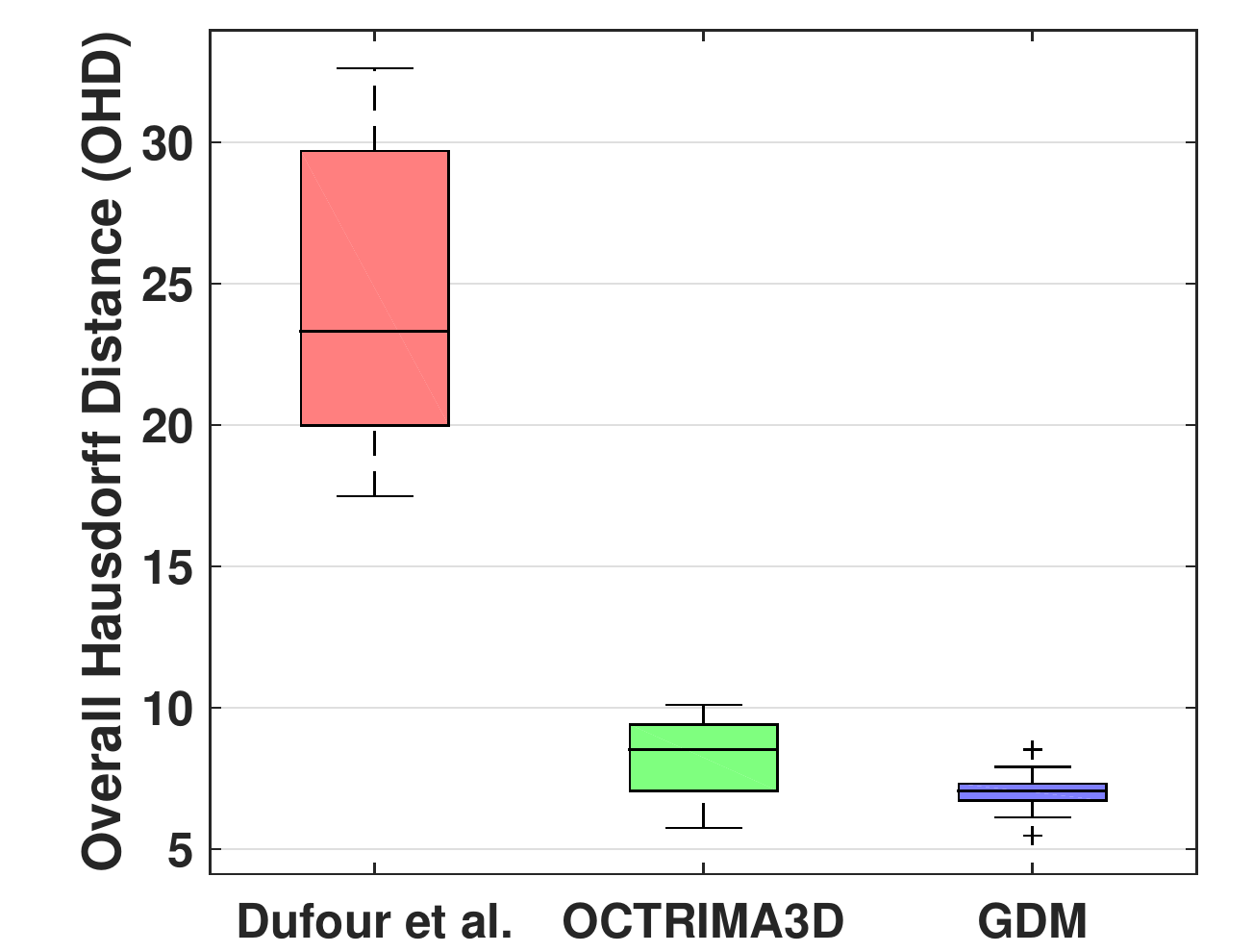}}\\
\vspace{-5pt}
\caption{Boxplots for the SE ($\mu m$), AE ($\mu m$), HD ($\mu m$), OSE ($\mu m$), OAE ($\mu m$) and OHD ($\mu m$) obtained by different methods in Table~\ref{tb:volumeB5}-\ref{tb:volumeOveall} for the 10 OCT volumes. 1st row: boxplots of Table~\ref{tb:volumeB5}; 2nd row: boxplots of Table~\ref{tb:volumeB7}; 3rd row: boxplots of Table~\ref{tb:volumeOveall}.}
\label{fig:boxPlotsVolume}
\end{figure}

Table~\ref{tb:volumeB5}-\ref{tb:volumeOveall} contain quantitative information for comparing the accuracy of the three methods on the 10 OCT volumes. Table~\ref{tb:volumeB5} lists the quantities for the surface $B_5$ around the fovea region, and Table~\ref{tb:volumeB7} presents the numerical results for the surface $B_7$ that is flatter and smoother. In Table~\ref{tb:volumeB5}, the SE quantity indicates that Dufour's method produces larger segmentation bias than the OCTRIMA3D and GDM. The SE values by the GDM are in the range of [-0.128$\mu m$ 0.6833$\mu m$], showing less variability than those by the other two methods. Moreover, the GDM leads to the smallest AE and HD quantities in all 10 cases, indicating that the GDM is the best among all the methods. Compared with those in Table~\ref{tb:volumeB5}, the quantities in Table~\ref{tb:volumeB7} show a significant improvement of all the methods. For example, the range of the HD quantity by Dufour's method has dropped from [25.688$\mu m$ 56.667$\mu m$] to [3.521$\mu m$ 27.853$\mu m$]. In addition, the accuracy gap between the OCTRIMA3D and GDM has been reduced. The HD values of Volume 3, 7 and 10 by the OCTRIMA has even become smaller than the corresponding values by the GDM. These improvements are the fact that the retinal surface $B_5$ is flat and the voxel values remain fairly constant. From the OAE and OHD in Table~\ref{tb:volumeOveall} we can observe that the accuracy of the GDM is the highest for the total retina surfaces among the existing approaches.

\begin{figure}[h!] 
\centering  
{\includegraphics[width=0.7\textwidth]{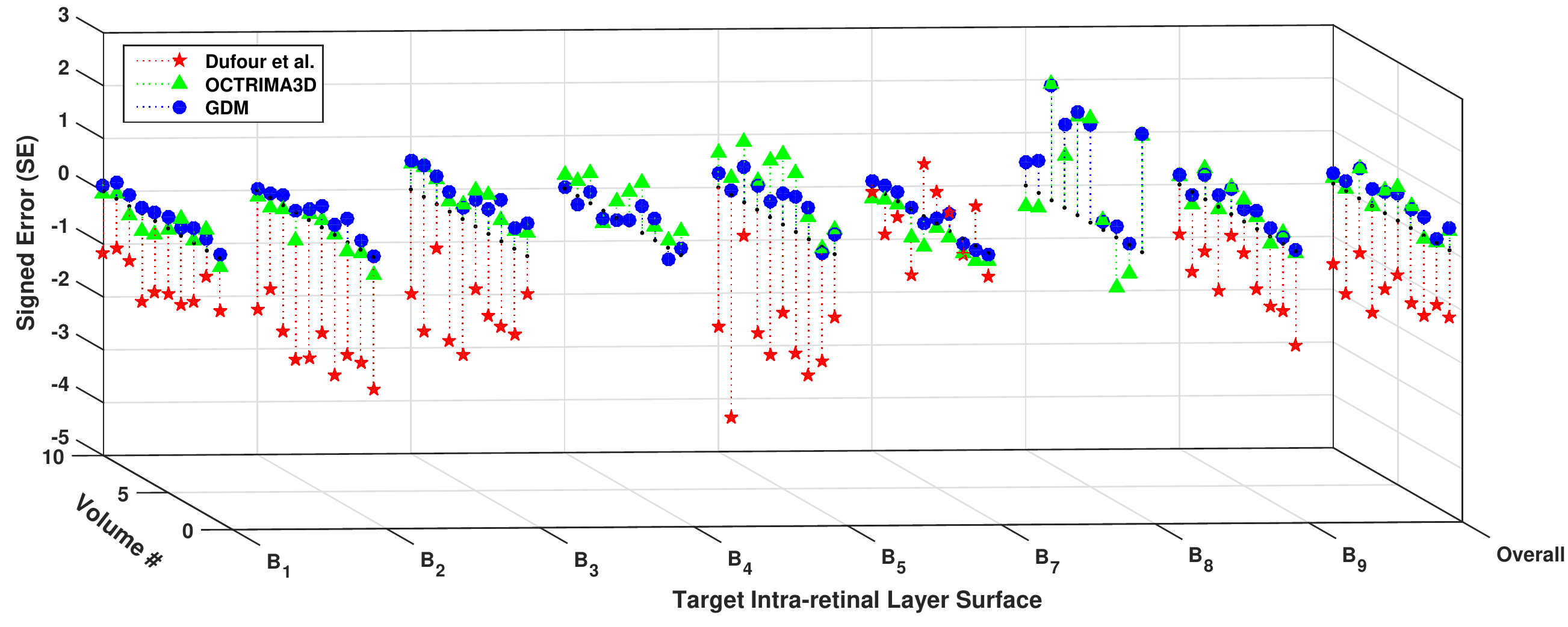}}\\
\vspace{-2pt}
{\includegraphics[width=0.7\textwidth]{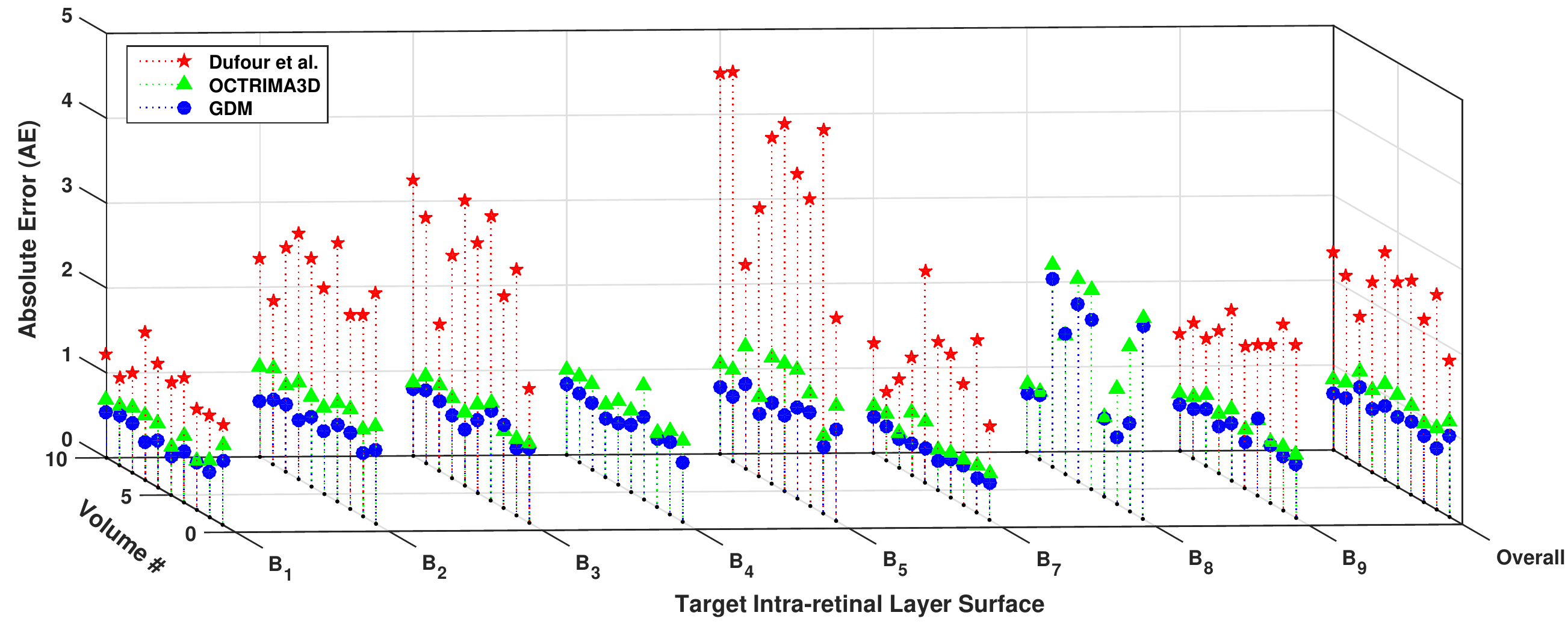}}\\
\vspace{-2pt}
{\includegraphics[width=0.7\textwidth]{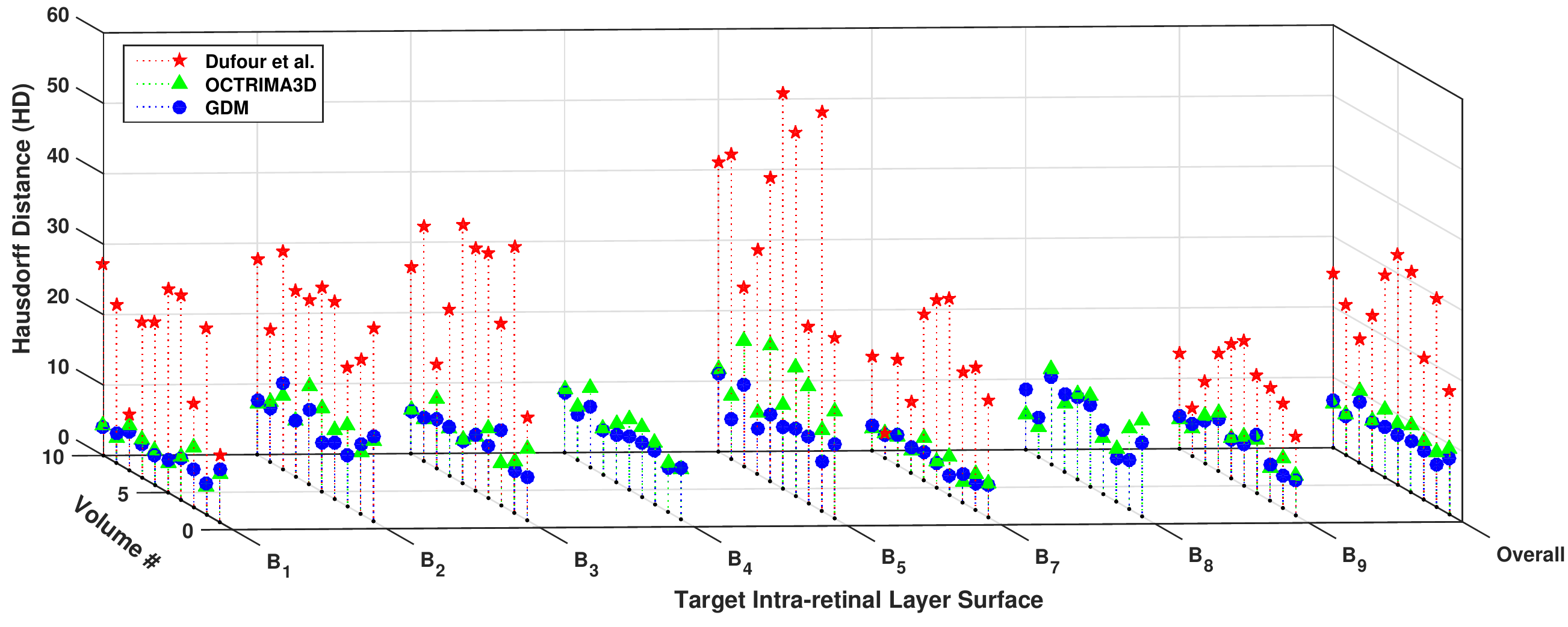}}\\
\vspace{-5pt}
\caption{3D plots of the SE ($\mu m$), AE ($\mu m$) and HD ($\mu m$) obtained using 10 OCT volumes using Dufour' method, OCTRMIA3D and GDM.}
\label{fig:3Dplot}
\end{figure}

The corresponding boxplots of Table~\ref{tb:volumeB5}-\ref{tb:volumeOveall} are shown in Figure~\ref{fig:boxPlotsVolume}. It is clear that the proposed GDM method performs consistently better, with higher accuracy and lower error rates for both flat and nonflat retina layers. The boxplots show that there is little variation in performance across the modelled structures and that even in the worst case scenario the proposed method yields lower error rate than the average performance of other methods. Furthermore, in Figure~\ref{fig:3Dplot} we present the 3D plots of the SE, AE and HD quantities computed by the three methods on the 10 OCT volumes. The SE values by the GDM are closer to zero and the AE and HD values by it remain smaller. The overall distribution of these discrete data points also indicates that the GDM results are less oscillating. We can thus conclude from Figure~\ref{fig:3Dplot} that the GDM is the best among all the methods compared for extracting intra-reintal layer surfaces from 3D OCT volume data in terms of both accuracy and robustness. 

\subsection{Computational Complexity Analysis}
The experimental results in Section \ref{NumericalRes} have shown that the performance of our algorithm is superior over others in terms of accuracy. In this section the performance of the different approaches in terms of the computational time is demonstrated. We implemented PDS, Chiu's method and GDM using Matlab 2014b on a Windows 7 platform with an Intel Xeon CPU E5-1620 at 3.70GHz and 32GB memory. For a $633 \times 496$ sized B-scan, with initialisation close to the true retinal boundaries, it takes 3.625s (500 iterations) for PDS to delineate two parallel boundaries. Chiu's method needs 1.962s to detect one boundary, while the GDM only takes 0.415s. Note that the time complexity of Chiu's graph search method is $O(|E|log(|V|))$, where $|V|$ and $|E|$ are the number of nodes and edges. In the context of boundary detection, $|V|= MN$ and $|E|=8MN$. Hence the time complexity of the method is $O(MNlog(MN))$. In contrast, our GDM solved using fast sweeping only has linear complexity of $O(MN)$, which is more efficient than Chiu's method. Instead of directly doing segmentation in 3D, the OCTRMIA3D explores spatial dependency between two adjacent B-scans and applies Chiu's method to each 2D frame independently. The OCTRMIA3D is thus able to track retinal boundaries in 3D volume efficiently. It has been reported in \cite{tian2015real} that the processing time of the OCTRMIA3D for the whole OCT volume of $496\times644\times51$ voxels was 26.15s, which is faster than our GDM (40.25s is used to segment a $496\times633\times10$ sized volume). However, such procedure in the OCTRMIA3D complicates the whole 3D segmentation process and might make the algorithm less general. Finally, Dufour's graph method needs 14.68s to detect the six intra-retinal layers surfaces on a $496\times633\times10$ sized volume. Dufour's method was implemented using a different programming language (C) and delineated different number of retinal surfaces from those of GDM so comparison can not be made between the two methods.  

\section{Conclusion}
We have presented a new automated segmentation framework based on the geodesic distance for delineating retinal layer boundaries in 2D/3D OCT images. The framework integrates horizontal and vertical gradient information and can thus account for changes in the both directions. Further, the exponential weight function employed within the framework enhances the foveal depression regions and highlights the weak and low contrast boundaries. As a result, the proposed method is able to segment complex retinal structures with large curvatures and other irregularities caused by pathologies. Extensive numerical results, validated with ground truth, demonstrate the effectiveness of proposed framework for segmenting both normal and pathological OCT images. The proposed method has achieved higher segmentation accuracy than existing methods, such as the parameter active contour model and the graph theoretic based approaches. Ongoing research includes integrating the segmentation framework into a system for detection and quantification of retinal fractures and other diseases of the retina.

\section{Appendix}
We present the 3D fast sweeping algorithm to solve the Eikonal equation (\ref{eq:eikonalEq}). Given a seed point $s_1$, its distance function $d(x)$ satisfies the following Eikonal equation
\begin{equation} 
\left|\nabla d(x)\right|=f(x), x\notin s_1 \label{eq:Eiknoal}
\end{equation}
with $d(s_1)=0$ and $f(x)=W^{-1}(x)$ where $W$ is defined in (\ref{eq:weights}). (\ref{eq:Eiknoal}) is a typical partial differential equation and it can be solved efficiently by using the fast sweeping algorithm proposed by Zhao \cite{zhao2005fast}. To do so, the Godunov upwind difference scheme is used to discretise (\ref{eq:Eiknoal}) as follows
\begin{equation}
\left[(d_{i,j,k}^n-d_{xmin}^n)^+\right]^2+\left[(d_{i,j,k}^n-d_{ymin}^n)^+\right]^2+\left[(d_{i,j,k}^n-d_{zmin}^n)^+\right]^2=f_{i,j,k}^2
\label{eq:dicreteEiknoal}
\end{equation} 
In equation (\ref{eq:dicreteEiknoal}), $d_{xmin}^n=min(d_{i,j+1,k}^n, d_{i,j-1,k}^n)$, $d_{ymin}^n=min(d_{i+1,j,k}^n, d_{i-1,j,k}^n)$, $d_{zmin}^n=min(d_{i,j,k+1}^n,d_{i,j,k-1}^n)$ and
$x^+=\left\{
\begin{array}{cc}
      x & x>0 \\
      0 & x\leq 0
\end{array} 
\right.$.
Boundary conditions need to be handled in the computational grid space. One-sided upwind difference is used for each of the 6 boundary faces of the grid space. For example, at the left boundary face, a one-sided difference along the $x$ direction is computed as
\begin{equation} \nonumber
\left[(d_{i,1,k}^n-d_{i,2,k}^n)^+\right]^2+\left[(d_{i,1,k}^n-d_{ymin}^n)^+\right]^2+\left[(d_{i,1,k}^n-d_{zmin}^n)^+\right]^2=f_{i,1,k}^2 
\end{equation}
$d_{xmin}^n$, $d_{ymin}^n$ and $d_{zmin}^n$ are then sorted in increasing order and the sorted version is recorded as $a_1$, $a_2$ and $a_3$. So, the unique solution to (\ref{eq:dicreteEiknoal}) is given as follows:
\begin{equation} 
d_{i,j,k}^{n+1}=min(d_{i,j,k}^n,\widetilde{d_{i,j,k}}) \label{eq:uniSolu}
\end{equation}
where $\widetilde{d}_{i,j,k}$ is a piecewise function containing three parts
\begin{equation} \nonumber
\widetilde{d_{i,j,k}} = \left\{ \begin{array}{l}
\frac{1}{3}\left( {{a_1} + {a_2} + {a_3} + \sqrt {3f_{i,j,k}^2 - {{({a_1} - {a_2})}^2} - {{({a_1} - {a_3})}^2} - {{({a_2} - {a_3})}^2}} } \right)\\
\frac{1}{2}\left( {{a_1} + {a_2} + \sqrt {2f_{i,j,k}^2 - {{({a_1} - {a_2})}^2}} } \right)\\
{a_1} + {f_{i,j,k}}
\end{array} \right.
\end{equation}
The three parts correspond to the following intervals, respectively
\begin{align*}
f_{i,j,k}^2 \ge  ({a_1} &- {a_3})^2 + {({a_2} - {a_3})^2}\\
{({a_1} - {a_2})^2} \le f_{i,j,k}^2 &< {({a_1} - {a_3})^2} + {({a_2} - {a_3})^2}\\
f_{i,j,k}^2 &< {({a_1} - {a_2})^2}
\end{align*}

To solve (\ref{eq:uniSolu}), which is not in analytical form, the fast Gauss-Seidel iteration with alternating sweeping orderings is used. For initialization, the value of the seed point $s_1$ is set to zero, and this value is fixed in later calculations. The rest of the points are set to large values, and these values will be update later. The whole 3D grid is traversed in the following orders for the Gauss-Seidel iteration
\begin{align*}
	(1)\; i=1:M, j=1:N, k=1:H;\; &(2)\; i=M:1, j=N:1, k=H:1\;\\
	(3)\; i=M:1, j=1:N, k=1:H;\; &(4)\; i=1:M, j=N:1, k=H:1\;\\
    (5)\; i=M:1, j=N:1, k=1:H;\; &(6)\; i=1:M, j=1:N, k=H:1\;\\
    (7)\; i=1:M, j=N:1, k=1:H;\; &(8)\; i=M:1, j=1:N, k=H:1\;\\
\end{align*}

\bibliographystyle{unsrt}
\bibliography{FinalVersion}

\end{document}